\documentclass{article}

\PassOptionsToPackage{numbers, compress}{natbib}

 \usepackage[main, final]{neurips_2026}

\usepackage[utf8]{inputenc} 
\usepackage[T1]{fontenc}    
\usepackage{hyperref}       
\usepackage{url}            
\usepackage{booktabs}       
\usepackage{amsmath}        
\usepackage{amsfonts}       
\usepackage{nicefrac}       
\usepackage{microtype}      
\usepackage{xcolor}         
\usepackage{graphicx}       
\usepackage{subcaption}
\usepackage{caption} 
\usepackage{calc} 
\usepackage{adjustbox}
\usepackage{enumitem}
\usepackage{array}

\title{FactorizedHMR: A Hybrid Framework for Video Human Mesh Recovery}

\author{%
  Patrick Kwon \\
  Institute of Artificial Intelligence \\
  University of Central Florida\\
  Orlando, FL 32817 \\
  \texttt{yo564250@ucf.edu} \\
  \And
  Chen Chen \\
  Institute of Artificial Intelligence \\
  University of Central Florida\\
  Orlando, FL 32817 \\
  \texttt{chen.chen@ucf.edu} \\
}

\begin{document}

\maketitle

\begin{abstract}
Human Mesh Recovery (HMR) is fundamentally ambiguous: under occlusion or weak depth cues, multiple 3D bodies can explain the same image evidence. This ambiguity is not uniform across the body, as distal articulations such as the arms and legs are more uncertain compared to torso pose and root structure. Building on this observation, we propose \textbf{FactorizedHMR}, a two-stage framework that treats these two regimes differently. A deterministic regression module first recovers a stable torso-root anchor, and a probabilistic flow-matching module then completes the remaining non-torso articulation. To make this completion reliable, we combine a composite target representation with geometry-aware supervision and feature-aware classifier-free guidance, preserving the torso-root anchor while improving single-reference recovery of ambiguity-prone articulation. We also introduce a synthetic data pipeline that provides the paired image-camera-motion supervision under diverse viewpoints. Across camera-space and world-space benchmarks, FactorizedHMR remains competitive with strong baselines, with the clearest gains in occlusion-heavy recovery and drift-sensitive world-space metrics. Our project page is available at \url{https://yj7082126.github.io/factorizedhmr/}

\end{abstract}

\section{Introduction}

Human Mesh Recovery (HMR) reconstructs 3D human pose, shape, and motion from visual observations, enabling applications in animation, virtual reality, and 3D content creation. Yet accurate and temporally consistent video HMR remains difficult under partial observation, especially with occlusion, truncation, and depth ambiguity. Such cases impose uncertainty that admits multiple plausible 3D explanations, making it challenging for deep-learning models to infer a concrete solution. 

\begin{figure}[t]
    \centering
    \hspace*{-0.005\textwidth}%
    \begin{minipage}{1.01\textwidth}
        \centering

        \begin{minipage}[t]{0.62\linewidth}
            \vspace{0pt}
            \centering
            \begin{tabular}{@{}ccc@{}}
                \scalebox{1}[1.05]{\includegraphics[width=0.31\linewidth]{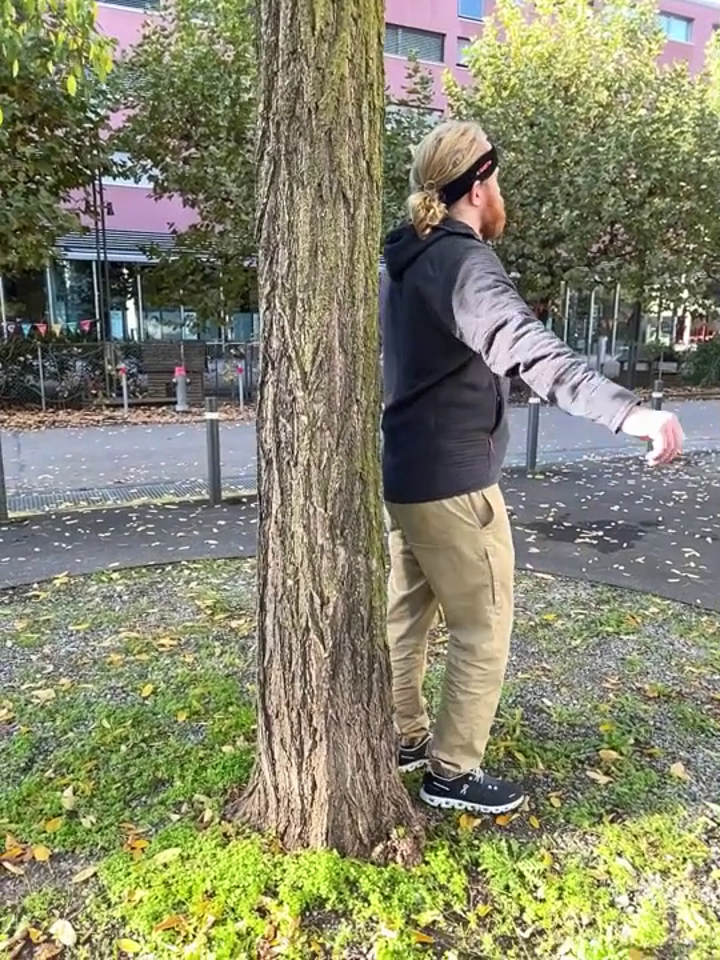}} &
                \scalebox{1}[1.05]{\includegraphics[width=0.31\linewidth]{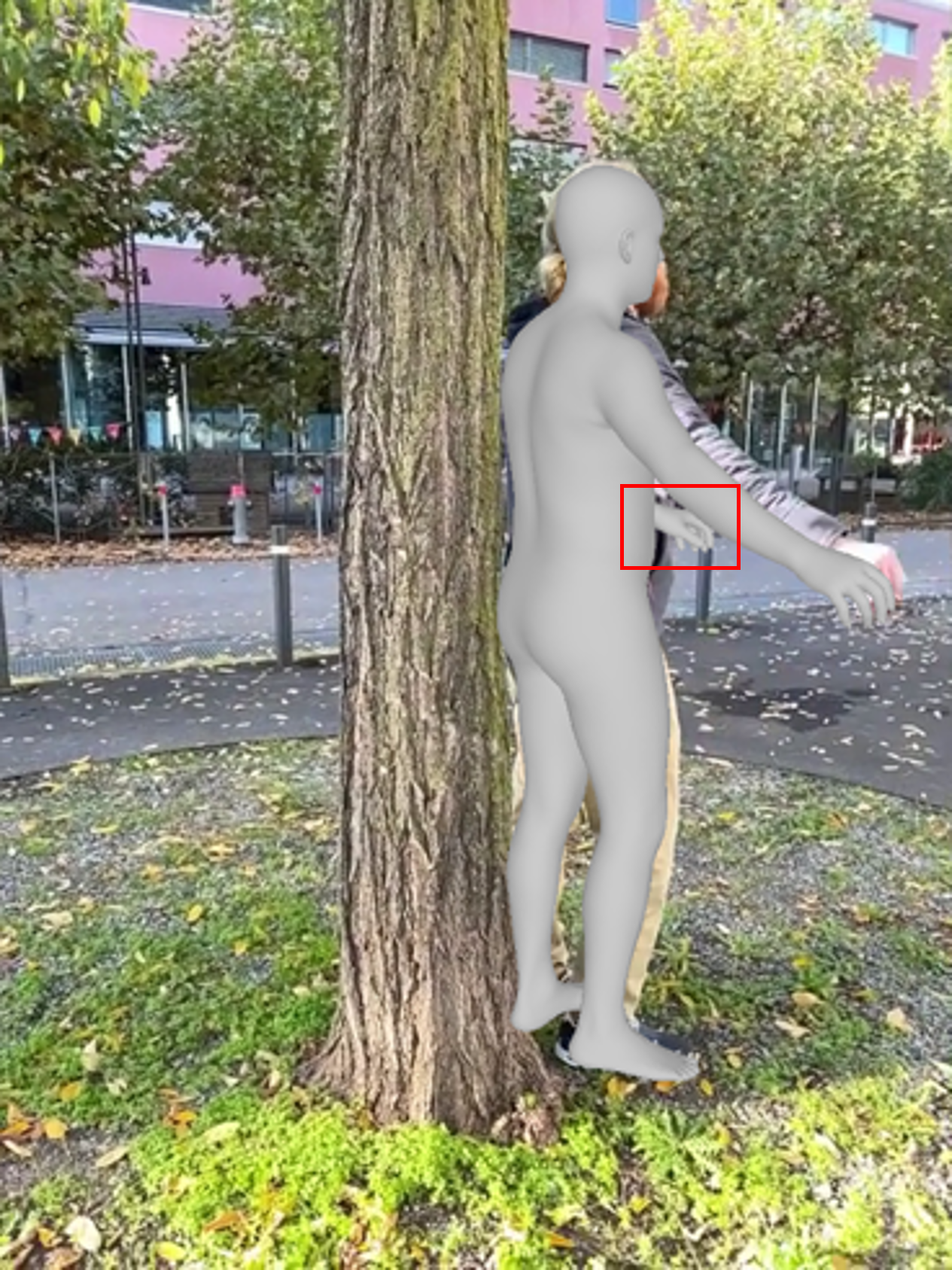}} &
                \scalebox{1}[1.05]{\includegraphics[width=0.31\linewidth]{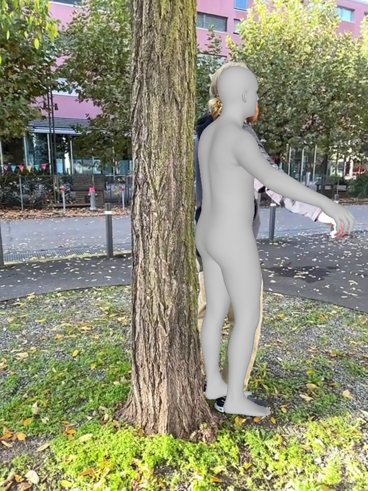}} \\
                \shortstack{(a) GT} & {(b) GVHMR \cite{shen2024gvhmr}} & {(c) Ours}
            \end{tabular}
        \end{minipage}
        \hfill
        \begin{minipage}[t]{0.36\linewidth}
            \vspace{0pt}
            \centering
            \includegraphics[width=0.9\linewidth]{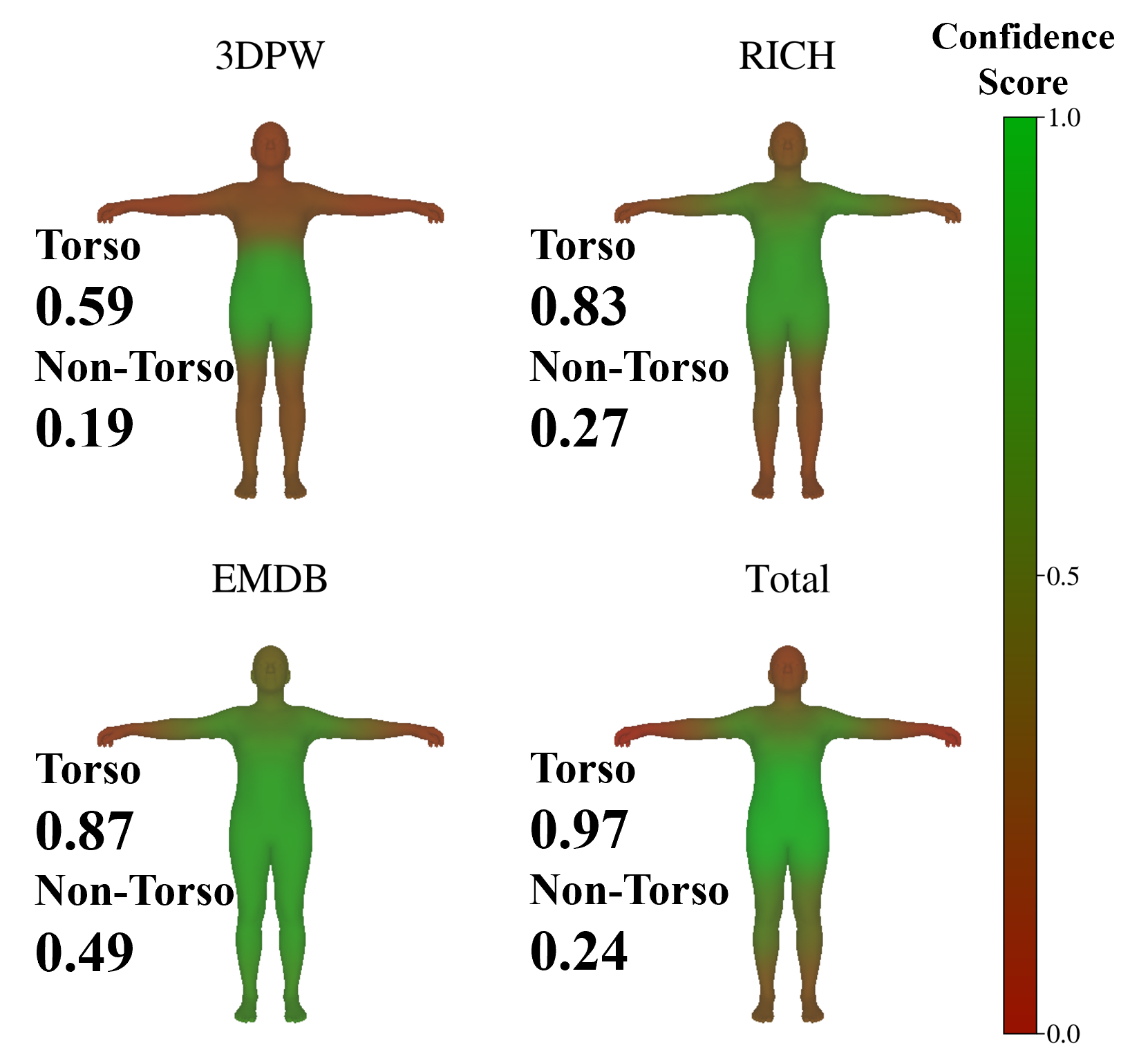}
        \end{minipage}

        \vspace{0.5em}

        \begin{minipage}[t]{0.62\linewidth}
            \vspace{-0.5em}
            \captionof{figure}{Qualitative comparison on EMDB \cite{kaufmann2023emdb}. The left arm and hand are heavily occluded by the tree, so multiple poses are plausible. GVHMR \cite{shen2024gvhmr} predicts a left hand that is not visible in the image (red square), illustrating how deterministic pipelines can commit to an implausible average solution under ambiguity. Our method instead preserves the visible pose while producing a more plausible completion for the occluded limb.}
            \label{fig:qualitative_occlusion}
        \end{minipage}
        \hfill
        \begin{minipage}[t]{0.36\linewidth}
            \vspace{-0.5em}
            \captionof{figure}{Visualization of per-joint confidence consistency across the test datasets, where higher values correspond to more reliable observation. Visual evidence is concentrated in the torso, while distal limbs are less consistently detected.}
            \label{fig:confidence_coverage}
        \end{minipage}
    \end{minipage}
    \vspace{-24pt}
\end{figure}

Most prior work addresses this with deterministic formulations, such as optimization-based fitting of parametric body models \cite{loper2015smpl, pavlakos2019smplx,bogo2016keep,huang2017towards,lassner2017unite,kanazawa2018end,kolotouros2019spin} and feed-forward regression models \cite{kanazawa2018hmr,kolotouros2019cmr,kocabas2020vibe,kocabas2021pare,goel2023hmr2}. These methods are efficient, but under ambiguity they tend to regress toward an average solution, leading to biased or over-smoothed reconstructions \cite{kocabas2020vibe,choi2021tcmr,goel2023hmr2}. Fig.~\ref{fig:qualitative_occlusion} shows a representative failure under severe occlusion, where a deterministic prediction averages into a visually unsupported hallucination. Probabilistic approaches improve expressiveness by modeling multiple plausible outcomes through generative modeling \cite{kingma2013vae,sohn2015cvae,ho2020ddpm,bishop1994mdn,pavlakos2018cvpr,wehrbein2021flows,kolotouros2021prohmr,stathopoulos2024scorehmr,saleem2025genhmr}, but applying it uniformly to the full state can be inefficient and unstable, as it allocates generative capacity and injects unnecessary variation to well-constrained variables such as torso, root, and camera quantities that are often better handled deterministically. Recent world-grounded systems expose the same trade-off: while these deterministic methods provide stable structural estimation, their limitations in ambiguous trajectory recovery expose an accuracy gap \cite{wang2024tram,shin2023wham,shen2024gvhmr}. This exposes a central limitation of existing formulations: they usually choose between deterministic stability and probabilistic expressiveness at the level of the whole body, rather than asking which parts are actually ambiguous.

This ambiguity is visible in Fig.~\ref{fig:confidence_coverage}: torso and proximal joints are better supported by visual evidence, whereas distal limbs are more variable. Appendix Tab.~\ref{tab:gvhmr_per_joint_mpJPE} shows this pattern under a deterministic baseline \cite{shen2024gvhmr}, where distal joints incur substantially larger errors. This motivates our central claim: \textit{HMR benefits from uncertainty-aware completion, in which stable variables are estimated deterministically while ambiguity-prone variables are refined probabilistically.}

We propose an uncertainty-aware factorization that separates video HMR into stable structural estimation and ambiguity-prone motion completion. Stage~1 deterministically estimates a \textit{structural anchor}, constituted of variables such as torso pose, body shape, and coarse camera-space motion, while Stage~2 uses conditional flow matching to complete the remaining non-torso articulation and world-space motion. By using probabilistic modeling as a targeted completion mechanism for
ambiguity-prone variables, rather than fully recovering the calibrated posterior over all plausible human motions, we concentrate generative capacity where uncertainty is most severe.

To make this effective, we equip Stage~2 with a composite motion representation that encodes both body joint rotations and positions, allowing Stage~2 to complete body articulation while remaining consistent with the structural anchor. We further pair it with representation-aware noising and geometry-aware supervision, including direct projection and joint-bone consistency losses, so that the completed motion remains faithful to the input rather than drifting away from the structural anchor.

Owing to the probabilistic nature of Stage~2, we use feature-aware classifier-free guidance to strengthen observation conditioning and improve fidelity under ambiguity. Finally, we introduce a camera-aware synthetic data pipeline that supplies the paired visual-motion supervision required by the factorized design: it supports stable camera-aware anchor estimation in Stage~1 and diverse ambiguity-aware completion learning in Stage~2. Together, these components turn the proposed factorization into a coherent and trainable pipeline for uncertainty-aware completion in video HMR. Since HMR benchmarks provide only a single target motion even for ambiguous frames, we evaluate whether our pipeline improves recovery in visually degraded settings, rather than fully characterizing all plausible reconstructions. In this setting, FactorizedHMR is designed to combine the stability of deterministic regression for well-constrained structure with the flexibility of probabilistic refinement for ambiguity-prone motion.

In summary, our contributions are as follows:
\begin{itemize}[leftmargin=*]
    \item We propose FactorizedHMR, a hybrid video HMR framework that decouples stable torso-root estimation from ambiguity-prone motion completion.
    
    \item We formulate selective probabilistic HMR as masked conditional flow matching, keeping anchor variables fixed while generating only non-torso and world-motion variables.

    \item We introduce geometry-aware completion objectives that combine composite rotation--joint targets, representation-aware noising, joint-bone consistency, projection loss, and feature-aware CFG.

    \item We develop a camera-aware synthetic training pipeline and demonstrate competitive overall performance, with the strongest gains under severe occlusion.
\end{itemize}

\section{Related Work}
\label{sec:related}

\begin{figure}[t]
    \centering
    \setlength{\tabcolsep}{0pt}
    \begin{tabular}{@{}m{0.56\linewidth}@{\hspace{0.03\linewidth}}m{0.4\linewidth}@{}}
        \centering\includegraphics[width=\linewidth]{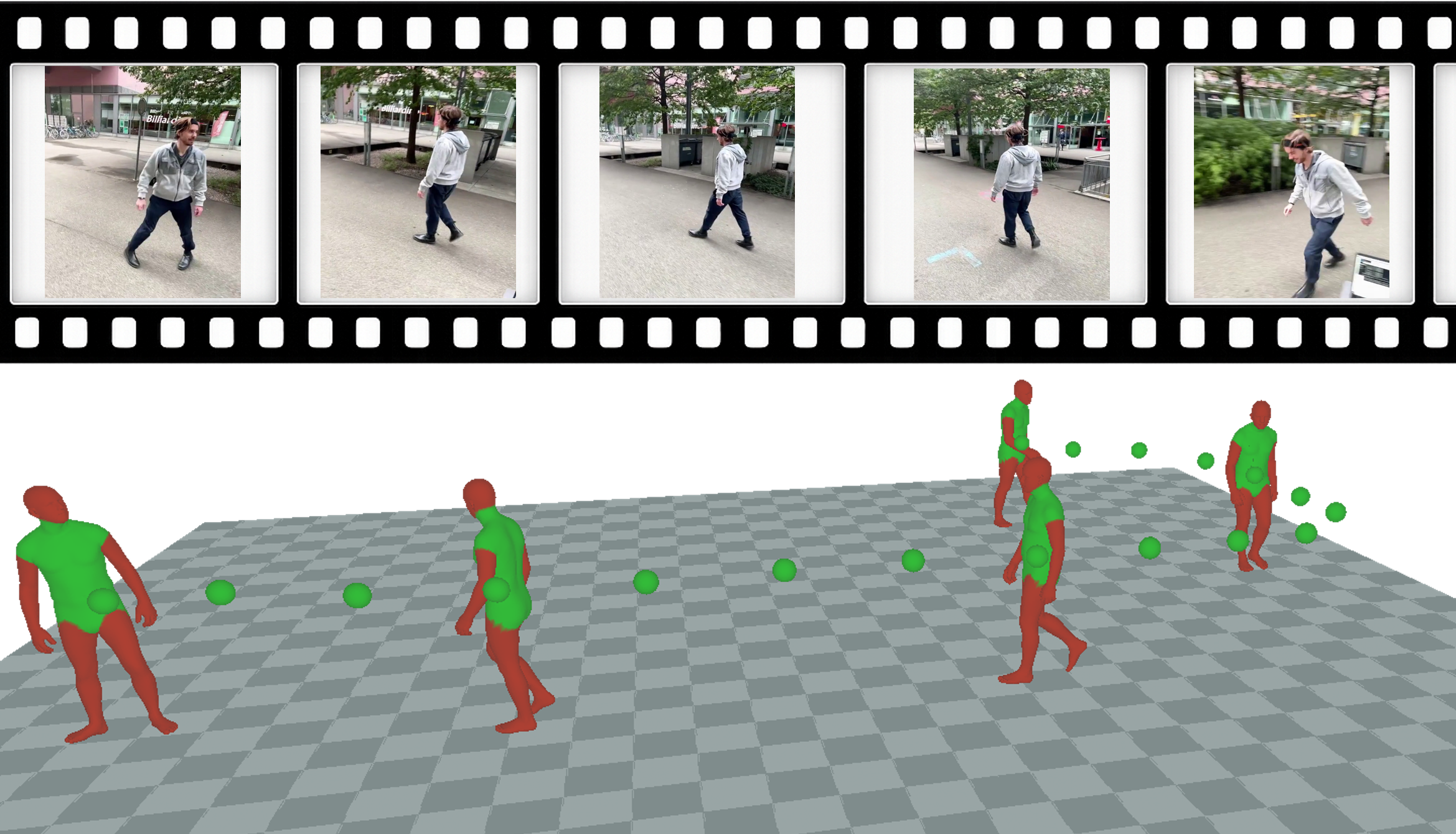} &
        \centering\includegraphics[width=\linewidth]{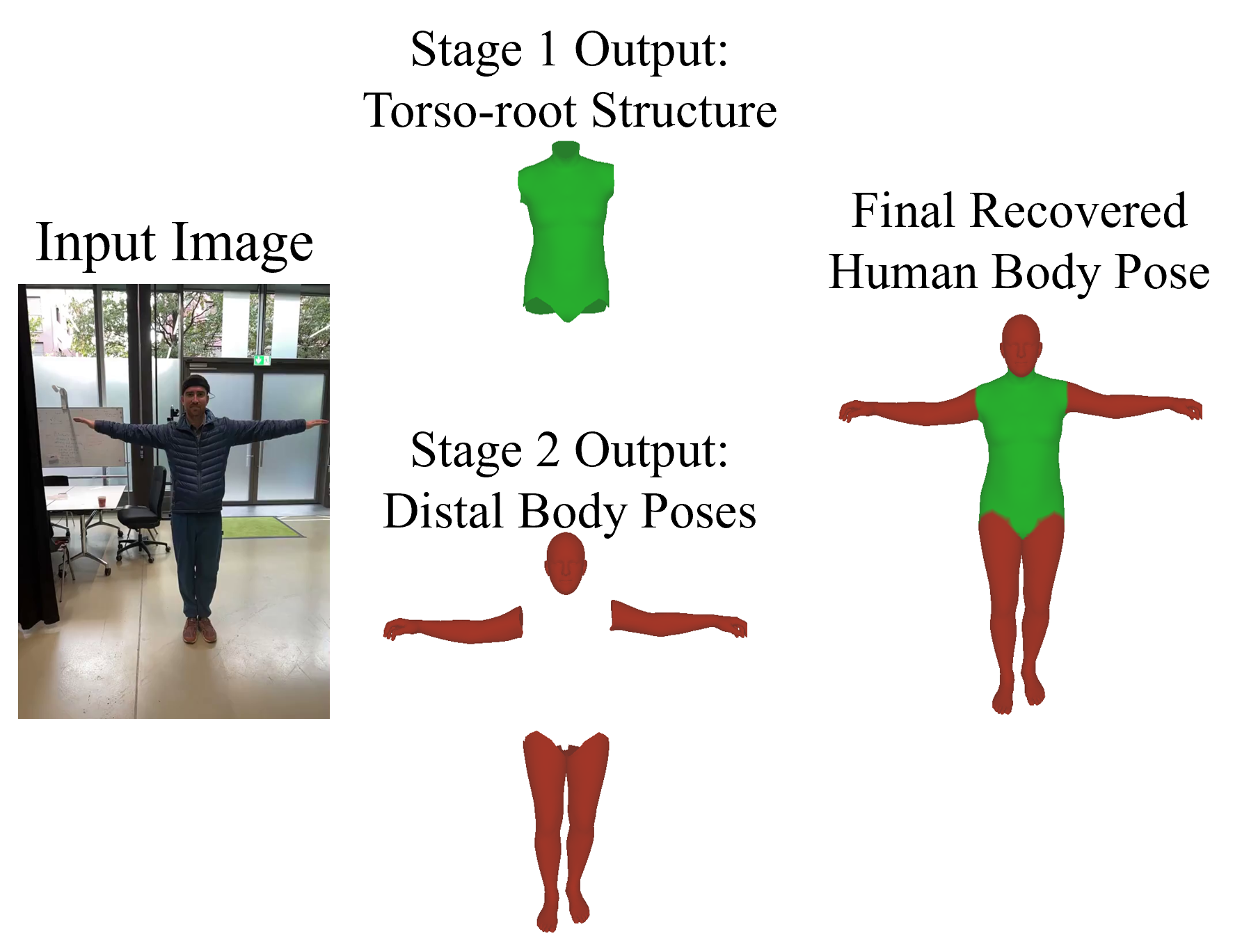}
    \end{tabular}
    \vspace{-2pt}
    \caption{Example results for FactorizedHMR. Stage~1 \textcolor{green}{(green)} deterministically estimates a torso-root anchor for the less ambiguous body parts and Stage~2 \textcolor{red}{(red)} uses conditional flow matching to complete the non-torso body poses, both for camera and world space.}
    \label{fig:placeholder}
    \vspace{-16pt}
\end{figure}

\textbf{Deterministic Human Mesh Recovery.} Early HMR work formulates reconstruction as deterministic regression from images to parametric body models such as SMPL \cite{loper2015smpl} and SMPL-X \cite{pavlakos2019smplx}. Optimization-based methods such as \cite{bogo2016keep} fit body models to 2D observations, whereas learning-based approaches \cite{kanazawa2018hmr,kolotouros2019cmr,kolotouros2019spin} regress pose and shape parameters directly from visual features. Subsequent work improves robustness through part-aware modeling \cite{kocabas2021pare}, hybrid kinematic constraints \cite{li2021hybrik}, and transformer-based architectures \cite{lin2021metro,goel2023hmr2,dwivedi2024tokenhmr}. For videos, temporal models such as \cite{kocabas2020vibe,choi2021tcmr} exploit motion consistency, while other approaches incorporate full-frame context \cite{li2022cliff} or iterative refinement \cite{song2020lgd,choutas2022ltfmm}. These methods are strong when the solution is well constrained, but they remain fundamentally deterministic and therefore often drift toward mean predictions under heavy ambiguity.

\textbf{Probabilistic and Generative Human Motion Modeling.} To better handle ambiguity, recent work introduces probabilistic and generative formulations for human pose and motion. Earlier approaches such as ProHMR \cite{kolotouros2021prohmr} model uncertainty directly in HMR, while motion priors such as HUMOR \cite{rempe2021humor} capture plausible human dynamics. More recently, diffusion-based methods \cite{ho2020ddpm,song2020ddim,tevet2023mdm,yuan2023physdiff,chen2023motiondiffuse} have enabled multi-modal motion generation and reconstruction. Several works extend these ideas to HMR, including ScoreHMR and GenHMR \cite{stathopoulos2024scorehmr,saleem2025genhmr}, as well as broader motion models such as GENMO \cite{li2025genmo}. Other directions explore diffusion priors for body fitting \cite{ho2025phd}, sim-to-real transfer \cite{gao2025disrtinbed}, and masked generative modeling \cite{fiche2025mega,lu2025dposerx}. While these methods capture uncertainty effectively, they typically model all joints and motion variables stochastically. Our method instead argues for selective probabilistic modeling: we preserve deterministic estimation where the evidence is reliable and reserve generation for the genuinely ambiguous subspace.

\textbf{World-Grounded Human Motion Recovery.} Recent work extends HMR beyond camera-centric pose estimation to world-grounded recovery, where global trajectory, scale, and camera motion are modeled jointly. Early efforts explore temporal and scene-aware reasoning \cite{arnab2019temporal3dpw,liu20214dhumanscene,guzov2021hps,hassan2021ssm}, while datasets such as \cite{vonmarcard20183dpw,mahmood2019amass,huang2022rich,black2023bedlam,kaufmann2023emdb} provide supervision for global motion and human-scene interaction. More recent methods explicitly address dynamic cameras and world-space consistency. For example, \cite{li2022dd,sun2023trace} jointly reason about camera motion and human dynamics, while \cite{shin2023wham,wang2024tram} recover globally consistent trajectories from in-the-wild videos. GVHMR \cite{shen2024gvhmr} introduces gravity-aligned coordinates for more stable world reasoning, DuoMo \cite{wang2026duomo} uses diffusion for global motion reconstruction, and PromptHMR \cite{wang2025prompthmr} studies prompt-conditioned recovery in a world-consistent setting. Our work complements this line by introducing uncertainty-aware factorization into world-grounded HMR, refining the ambiguous components while keeping the stable structure fixed.

\section{Methodology}
\label{sec:methods}

\subsection{Problem formulation}

Given an input video $\mathbf{V}=\{I^t\}_{t=0}^{T}$, our goal is to recover temporally consistent human motion in both camera space and world space.
We adopt SMPL-X \cite{pavlakos2019smplx} to represent the body. For each frame $t$, the final recovered motion state is
\begin{equation}
\mathbf{s}^{0:T}
=
\left\{
\theta^{0:T},\beta,\Gamma_c^{0:T},\tau_c^{0:T},\Gamma_w^{0:T},\tau_w^{0:T}
\right\}
\end{equation}
where $\theta \in \mathbb{R}^{21\times 3}$ denotes the local body pose, represented by the relative rotations of the 21 body joints,
$\beta \in \mathbb{R}^{10}$ denotes body shape,
$\Gamma_c$ and $\tau_c$ denote the camera-space trajectory (global orientation and translation), and
$\Gamma_w$ and $\tau_w$ denote the world-space trajectory.\footnote{For notational simplicity and to conserve space, we omit the frame index $t$ when the meaning is clear, and use explicit time indices only when needed.}

Our key observation is that uncertainty in human body motion is not uniform. Torso pose, body shape, and camera-space root motion are usually well constrained by visual evidence, and can therefore be treated as a \textit{structural anchor}. In contrast, non-torso articulation---including the limbs and head---is less certain because these joints are defined relative to the torso and are more sensitive to occlusion, truncation, and depth ambiguity. World-motion variables are likewise more ambiguous, as they require disentangling body motion from camera motion.

\begin{figure}[t]
    \centering
    \includegraphics[width=\linewidth]{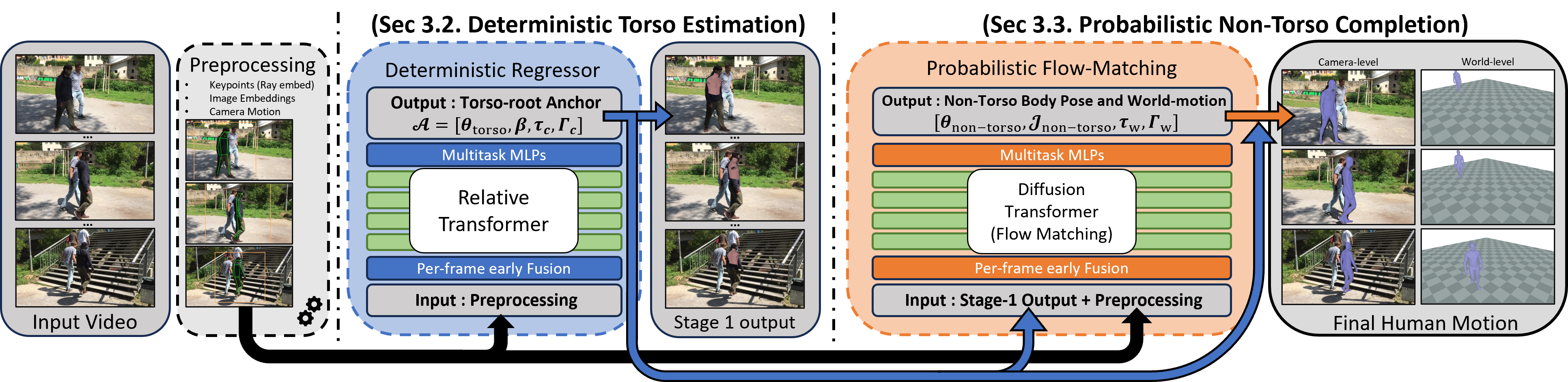}
    \caption{Overview of FactorizedHMR. Input video frames are preprocessed into ray-embedded keypoints, image embeddings, and camera-motion features. Stage~1 uses a deterministic regressor to predict a torso-root structural anchor, $\mathcal{A}=[\theta_\text{torso},\beta,\Gamma_c,\tau_c]$. Conditioned on the shared video features and this anchor, Stage~2 uses masked flow matching to complete the non-torso articulation and world-motion variables, yielding the final human motion in camera and world space. The two stages are trained sequentially, and the training losses are described in Sec.~3.4.}
    \label{fig:overview}
    \vspace{-12pt}
\end{figure}

Accordingly, we decompose the body pose as $\theta=[\theta_\text{torso}, \theta_\text{non-torso}]$.
Here, $\theta_\text{torso} \in \mathbb{R}^{8\times 3}$ refers to the 8 body poses constituting the torso region, which we define as the hips, spine, and shoulders.
$\theta_\text{non-torso} \in \mathbb{R}^{13\times 3}$ refers to the 13 remaining body poses outside the torso, namely the legs, arms, and head.
We then divide the pipeline into two stages. Stage~1 is a deterministic regressor that estimates the structural anchor $\mathcal{A}_t$:
\begin{equation}
\mathcal{A}
=
\left[
\theta_\text{torso},\beta,\Gamma_c,\tau_c
\right]
\end{equation}
Conditioned on $\mathcal{A}$, Stage~2 is formulated as a flow-matching model trained with a probabilistic objective, and its role is to complete the ambiguous non-torso articulation $\theta_\text{non-torso}$ together with the world-motion variables. The final motion state $\mathbf{s}^{0:T}$ is obtained by combining $\mathcal{A}$ with the Stage~2 completed variables. Figure~\ref{fig:overview} summarizes the pipeline.

\subsection{Stage 1: Deterministic structural estimation}

Stage~1 uses a deterministic transformer~\cite {Vaswani2017AttentionIA} with rotary position embeddings~\cite{Su2021RoFormerET} to estimate the structural anchor $\mathcal{A}$ from video observations, motivated by GVHMR~\cite{shen2024gvhmr}. To support reliable reasoning about both camera-space and world-space motion, GVHMR defines a gravity-view coordinate system in addition to the camera-view system, where the world-space trajectory is not predicted directly; instead, GVHMR predicts the global orientation with respect to the gravity direction $g$ $(\Gamma_\text{GV})$ and local root velocities $v_\text{root}$, from which the final $\Gamma_w$ and $\tau_w$ are recovered. We adopt a similar representation, but we do not predict $\Gamma_\text{GV}$ or $v_\text{root}$ in Stage~1. Instead, Stage~1 is restricted to the lower-uncertainty subset and predicts only torso pose, body shape, and camera-space trajectory variables. The more ambiguous world-motion components are deferred to Stage~2, which keeps the first stage focused on stable structural estimation.

Inputs include bounding-box features~\cite{li2022cliff}, 2D keypoint observations~\cite{shin2023wham}, image features~\cite{goel2023hmr2}, and relative camera-motion features~\cite{shin2023wham,shen2024gvhmr}.
Rather than using raw keypoint coordinates, we convert them to camera-ray directions using the camera intrinsics and encode them with sinusoidal embeddings~\cite{mildenhall2020nerf}.
Each modality is projected to a shared feature dimension with a dedicated MLP, and the resulting embeddings are summed into per-frame tokens
processed by the relative transformer.


\subsection{Stage 2: Non-torso articulation and world completion via masked flow matching}

The rotations of each body pose in $\theta$ are defined relative to its parent joint; for example, the elbow joint is defined relative to the shoulder joint.
Relative joint rotations alone are not always sufficient for completing ambiguous non-torso articulation, because they specify pose only with respect to parent joints and do not directly expose the resulting limb geometry. This is especially important in Stage~2, where the model must predict uncertain articulation while conditioning on the fixed torso anchor from Stage~1. We therefore augment the rotation representation with 3D camera-space joint positions $\mathcal{J}$, allowing the model to use both rotational and positional cues~\cite{Zhang2024RoHMRH}.
Following GVHMR's notation, the Stage~2 latent is \begin{equation}
z =
\left[
(\theta_\text{torso}, \theta_\text{non-torso}), (\mathcal{J}_\text{torso}, \mathcal{J}_\text{non-torso}), \beta, \Gamma_c, \tau_c, \Gamma_\text{GV}, v_\text{root}
\right]
\end{equation}
with $\mathcal{J}_\text{torso}$ and $\mathcal{J}_\text{non-torso}$ corresponding to $\theta_\text{torso}$ and $\theta_\text{non-torso}$. The variables copied from Stage~1 are fixed by the mask $\mathcal{M}$, while Stage~2 generates the non-torso and world-motion subset. The rotation branch defines the final body, and the joint-position branch serves as auxiliary geometric supervision during training.

\textbf{Representation-Aware Noising and Inference.}
We partition the Stage~2 latent into a known subset $\mathbf{z}_K$ and an unknown subset $\mathbf{z}_U$, where $K$ denotes the variables fixed by the Stage~1 anchor $\mathcal{A}$ and $U$ denotes the variables that Stage~2 must generate:
\begin{equation}
\mathbf{z} = \bigl(\mathbf{z}_K,\mathbf{z}_U\bigr)
\end{equation}
with the default partition
\begin{equation}
\mathbf{z}_K=
\left[
\theta_\text{torso},\mathcal{J}_\text{torso},\beta,\Gamma_c,\tau_c
\right],
\qquad
\mathbf{z}_U=
\left[
\theta_\text{non-torso},\mathcal{J}_\text{non-torso},\Gamma_{\text{GV}},v_{\text{root}}
\right]
\end{equation}
Here, $\mathbf{z}_K$ contains the coordinates copied from $\mathcal{A}$, namely the low-uncertainty torso, shape, and camera-space variables, whereas $\mathbf{z}_U$ contains the ambiguous non-torso and world-motion variables that are completed probabilistically. We use a binary mask $\mathbf{M}$ that is zero on $\mathbf{z}_K$ and one on $\mathbf{z}_U$.

Given effective source noise $\tilde{\boldsymbol{\epsilon}}$, we define the masked probability path
\begin{equation}
\mathbf{z}_t =
(1-\mathbf{M})\odot\mathbf{z}_K
+
\mathbf{M}\odot\left((1-t)\tilde{\boldsymbol{\epsilon}}+t\mathbf{z}\right)
\end{equation}
so that the known coordinates remain fixed while the unknown subspace is transported from noise to data.
In practice, for the joint-position coordinates we scale the source noise standard deviation by $0.5$, since isotropic Gaussian perturbations are more destructive in Cartesian joint space.


\textbf{Masked Flow Matching.} We instantiate the generative completion model with flow matching and train a velocity field over this masked path:
\begin{equation}
\mathcal{L}_{\mathrm{FM}}
=
\mathbb{E}_{t,\tilde{\boldsymbol{\epsilon}},\mathbf{z}}
\left[
\left\|
\mathbf{M}\odot
\left(
v_\phi(\mathbf{z}_t,t,\mathbf{c})
-
(\mathbf{z}-\tilde{\boldsymbol{\epsilon}})
\right)
\right\|_2^2
\right]
\end{equation}
where $v_\phi$ predicts the velocity field from the source distribution to the target latent.
At inference time, only the generated coordinates are initialized from Gaussian noise, while the known structural coordinates remain fixed.
Following flow matching, we integrate the learned velocity field with an ODE sampler while re-imposing $\mathbf{z}_K$ after each update,
yielding conditional completion around a fixed structural anchor. \textit{To the best of our knowledge, this is the first attempt to use flow matching for human mesh recovery tasks.}

\textbf{Anchor Conditioning and CFG.} Stage~2 shares the visual and camera conditions used by Stage~1, and additionally receives a compact torso-pose condition formed from the Stage~1 anchor $\mathcal{A}$: camera-frame root orientation and selected torso poses. This exposes the structural anchor both as fixed coordinates in $\mathbf{z}_K$ and as explicit conditioning.

We train with classifier-free guidance by randomly dropping observation-heavy conditions during training. In standard CFG for human motion,
the unconditional target is represented as a zero vector. Instead, we separate conditions related to the structural anchor
(bounding-box features, relative camera angles, and the torso-pose condition) from those related to non-torso articulation (keypoint rays and image features).
The unconditional target is then formed from the conditions related to the structural anchor, without information related to non-torso articulation.
This modified guidance strengthens observation-conditioned completion of ambiguity-prone non-torso articulation without discarding the structural anchor.

\subsection{Training}
We train the two stages sequentially. Stage~1 is first trained with a deterministic regression objective, using direct supervision on the variables in $\mathcal{A}$. Stage~2 is then trained with the masked flow-matching objective described in Sec.~3.3, conditioned on the Stage~1 anchor.
For each sequence, we sample a time $t$ and Gaussian source noise $\boldsymbol{\epsilon}$,
construct the masked intermediate state $\mathbf{z}_t$, and train the network to predict the velocity from the source distribution
to the target motion distribution. The flow-matching loss is applied only to generated coordinates,
while known torso/root coordinates are preserved by the mask and serve as structural anchors throughout training.

\textbf{Geometry-Aware Losses.} In addition to the generative objective, we use geometry-aware losses similar to GVHMR~\cite{shen2024gvhmr} for both stages. For Stage~1, these direct regression losses provide added supervision on the variables in $\mathcal{A}$. For Stage~2, we supervise normalized pose, joints, and root-motion targets together with camera-space and root-relative 3D joints,
projected 2D joints and vertices, root-relative vertices, camera-frame translation, and world translation recovered from the predicted local velocity.
This keeps the completed motion consistent with SMPL-X geometry, the image evidence, and the camera/world trajectory constraints, while preserving $\mathcal{A}$ as a fixed structural prior.

\textbf{Joint-Bone Consistency.} Unlike GVHMR, our Stage~2 target is over-parameterized with both rotations and explicit joint positions,
so the two branches can become inconsistent if trained independently. We therefore introduce a joint-bone consistency loss
that penalizes disagreement between the predicted joint-position branch and the forward-kinematics joints implied by the predicted
rotation branch, including both joint locations and bone vectors. This loss enforces self-consistency between the two predicted representations,
preventing the explicit joint-position branch from drifting away from the kinematics implied by the predicted rotations.
\begin{equation}
\mathcal{L}_{\mathrm{cons}}
=
\sum_j
\rho\!\left(
\hat{\mathbf{J}}_{c,j}
-
\mathbf{J}_{\mathrm{FK},j}
\right)
+
\sum_{(j,p)\in\Omega_{\mathrm{bone}}}
\rho\!\left(
\left(\hat{\mathbf{J}}_{c,j}-\hat{\mathbf{J}}_{c,p}\right)
-
\left(\mathbf{J}_{\mathrm{FK},j}-\mathbf{J}_{\mathrm{FK},p}\right)
\right)
\end{equation}
where $\Omega_{\mathrm{bone}}$ denotes the single-parent bone pairs of the SMPL-X body.

\textbf{Direct Projection Supervision.} We also introduce a direct projection loss for ambiguous non-torso joints.
Rather than relying only on broad vertex- or joint-level reprojection losses, we project visible, non-torso joints
from the predicted camera-space body into the image using the camera intrinsics, and apply a robust pixel-space loss
against the corresponding ground-truth projections. This term directly targets the limb articulation that Stage~2 is
designed to complete, and we gradually increase its weight during training to avoid destabilizing the early generative objective.
\begin{equation}
\mathcal{L}_{\mathrm{proj}}
=
\sum_{j\in\Omega_{\mathrm{proj}}}
\rho\!\left(
\Pi_K\!\left(\hat{\mathbf{J}}^{\,c}_j\right)
-
\mathbf{u}_j
\right)
\end{equation}
where $\Omega_{\mathrm{proj}}$ denotes the selected non-torso joints, $\Pi_K$ is projection with camera intrinsics $K$, and $\rho$ is a robust pixel-space penalty.


\subsection{Synthetic dataset generation pipeline}



\textbf{Motivation.} Motion-only corpora such as AMASS~\cite{mahmood2019amass} contain rich human dynamics, but they do not provide paired realistic videos and are therefore insufficient for training models for recovering human motion from imagery. Real videos offer much richer appearance variation, but their annotations are often noisy or incomplete. To address this, we construct a synthetic pipeline that preserves exact motion-capture supervision while using Uni3C~\cite{cao2025uni3c} as a video synthesis engine, which unifies camera and human-motion control from a single-view image across broader visual domains. Unlike conventional rendering pipelines~\cite{black2023bedlam}, which are constrained by available mesh textures, assets, and backgrounds, Uni3C allows us to sample a wider range of identities, clothing styles, and scene contexts, providing us camera-aware supervision with greater appearance diversity.

\textbf{Generation Pipeline.} Given a motion sequence and its SMPL-X parameters, we sample a camera trajectory with known camera parameters. We render geometric control signals, including SMPL-X body renderings, hand renderings, masks, and an OpenPose \cite{cao2019openpose}-style first-frame pose. The pose rendering is used to synthesize a reference image with identity and scene prompts using depth-conditioned FLUX.1~\cite{flux2024}. We then estimate monocular metric depth for the reference image using Depth Pro~\cite{Bochkovskii2024} and unproject it into a scene point cloud, which serves as the environment coordinate system. The SMPL-X motion and scene point cloud are rendered under the same target camera trajectory, producing aligned human-motion and camera-control conditions for Uni3C. The final RGB video is then generated by Uni3C from the reference image, the rendered point-cloud trajectory, and the rendered human motion. Additional implementation details are provided in the \textbf{Appendix}.

\section{Experiments}
\label{sec:experiments}


\paragraph{Model and Training.}
Both Stage~1 and Stage~2 are built on a 12-layer transformer architecture. Each layer uses rotary position embeddings, and each attention layer has 8 heads with a hidden dimension of 512. We train both models from scratch on a mixture of AMASS, BEDLAM, H36M, and 3DPW \cite{mahmood2019amass,black2023bedlam,Ionescu2014Human36MLS,vonmarcard20183dpw}.
Training uses mixed supervision: AMASS provides motion-only sequences, while BEDLAM, H36M, and 3DPW provide paired image-motion supervision. In addition, our Uni3C-based \cite{cao2025uni3c} synthetic sequences expand camera-aware coverage under dynamic viewpoints. We also apply training-time augmentation, including keypoint augmentation similar to WHAM \cite{shin2023wham}. Training was done with an Adam optimizer \cite{Kingma2014AdamAM}, takes approximately 36 hours on 4 A5000 GPUs.

\begin{table*}[t]
  \caption{Camera-space metrics. We evaluate the camera-space motion quality on the 3DPW, RICH and EMDB-1 datasets. $^*$ denotes models trained with the 3DPW training set.}
  \label{tab:camera_results}
  \centering
  \scriptsize
  \setlength{\tabcolsep}{1.6pt}
  \begin{tabular}{lccc ccc ccc}
    \toprule
    & \multicolumn{3}{c}{3DPW} & \multicolumn{3}{c}{RICH} & \multicolumn{3}{c}{EMDB} \\
    \cmidrule(lr){2-4} \cmidrule(lr){5-7} \cmidrule(l){8-10}
    Models & \shortstack{PA-MPJPE $\downarrow$} & MPJPE $\downarrow$ & PVE $\downarrow$ & \shortstack{PA-MPJPE $\downarrow$} & MPJPE $\downarrow$ & PVE $\downarrow$ & \shortstack{PA-MPJPE $\downarrow$} & MPJPE $\downarrow$ & PVE $\downarrow$ \\
    \midrule
    \multicolumn{10}{l}{\emph{Per-frame}} \\
    CLIFF$^{*}$ \cite{li2022cliff} & 43.0 & 69.0 & 81.2 & 56.6 & 102.6 & 115.0 & 68.1 & 103.3 & 128.0 \\
    HybrIK$^{*}$ \cite{li2021hybrik} & 41.8 & 71.6 & 82.3 & 56.4 & 96.8 & 110.4 & 65.6 & 103.0 & 122.2 \\
    HMR2.0 \cite{goel2023hmr2} & 44.4 & 69.8 & 82.2 & 48.1 & 96.0 & 110.9 & 60.6 & 98.0 & 120.3 \\
    ReFit$^{*}$ & 40.5 & 65.3 & 75.1 & 47.9 & 80.7 & 92.9 & 58.6 & 88.0 & 104.5 \\
    \midrule
    \multicolumn{10}{l}{\emph{Temporal}} \\
    VIBE$^{*}$ \cite{kocabas2020vibe} & 51.9 & 82.9 & 98.4 & 68.4 & 120.5 & 140.2 & 81.4 & 125.9 & 146.8 \\
    TRACE$^{*}$ \cite{sun2023trace} & 50.9 & 79.1 & 95.4 & -- & -- & -- & 70.9 & 109.9 & 127.4  \\
    SLAHMR & 55.9 & -- & -- & 52.5 & -- & -- & 69.5 & 93.5 & 110.7 \\
    PACE & -- & -- & --  & 49.3 & -- & -- & -- & -- & -- \\
    WHAM$^{*}$ \cite{shin2023wham} & \underline{35.9} & 57.8 & \underline{68.7} & 44.3 & 80.0 & 91.2 &  50.4 & 79.7 & 94.4 \\
    GVHMR$^{*}$ \cite{shen2024gvhmr} & 37.1 & \underline{56.9} & 68.8 & \underline{39.5} & \underline{66.0} & \textbf{74.4} & \textbf{44.6} & \underline{74.0} & \textbf{86.0} \\
    TRAM$^{*}$ \cite{wang2024tram} & \textbf{35.6} & 59.3 & 69.6 & -- & -- & -- & \underline{45.7} & 74.4 & 86.6 \\
    GENMO$^{*}$ \cite{li2025genmo} & 42.9 & 62.3 & 82.8 & -- & -- & -- & 47.6 & 81.2 & 94.6 \\
    \midrule
    Ours$^{*}$ & 36.9 & \textbf{56.2} & \textbf{68.3} & \textbf{39.1} & \textbf{65.8} & \underline{74.5} & \underline{45.7} & \textbf{73.6} & \underline{86.5} \\
    \bottomrule
  \end{tabular}
  \vspace{-12pt}
\end{table*}

\paragraph{Evaluation.}
Following recent world-grounded HMR benchmarks \cite{shin2023wham,wang2024tram,shen2024gvhmr}, we report camera-space metrics on 3DPW, RICH, and EMDB-1, and world-space metrics on RICH and EMDB-2 \cite{vonmarcard20183dpw,huang2022rich,kaufmann2023emdb}. To better probe robustness under visually challenging conditions, we also report results on 3DPW-XOCC \cite{li2023niki}, an occlusion-focused benchmark that simulates severe occlusion and truncation on 3DPW through random occlusion patches and frame truncation. Examples of 3DPW-XOCC are available at Fig.~\ref{fig:3dpwxocc_example} in \textbf{Appendix}.

We adopt the standard camera-space and world-space metrics used in recent world-grounded HMR work \cite{shin2023wham,wang2024tram,shen2024gvhmr,li2025genmo}; additional metric definitions are provided in Appendix~\ref{app:metrics}. We compare against a broad set of baselines, including per-frame methods such as CLIFF, HybrIK, and HMR2.0 \cite{li2022cliff,li2021hybrik,goel2023hmr2}, temporal camera-space or lifting-based methods such as VIBE, TRACE, WHAM, and TRAM \cite{kocabas2020vibe,sun2023trace,shin2023wham,wang2024tram}, and recent world-grounded or generative methods, most notably GVHMR and GENMO \cite{shen2024gvhmr,li2025genmo}. Because these benchmarks provide only one reference motion, they mainly test whether stochastic refinement improves single-reference recovery under ambiguity, rather than fully measuring calibration or sample diversity.

\subsection{Evaluation of motion estimation}

\paragraph{Camera-space Motion Recovery.}

Tab.~\ref{tab:camera_results} shows that our method remains highly competitive on standard camera-space benchmarks. MPJPE improves consistently across datasets, suggesting better absolute camera-space reconstruction, while PA-MPJPE and PVE remain broadly competitive. This is partly expected because evaluation under SMPL-based protocols introduces a small mismatch for SMPL-X predictions, indicated by GVHMR \cite{shen2024gvhmr}. Overall, the camera-space results show that our factorization can improve ambiguous pose articulation without sacrificing the stability of the deterministic torso output. Also, since these benchmarks contain limited severe occlusion and only a single reference motion, they likely understate the benefits of probabilistic refinement in genuinely ambiguous cases.

\paragraph{Regional Breakdown.}

Tab.~\ref{tab:stage1_camera_results} provides a targeted view of the factorized design by separating evaluation over torso and non-torso body regions. The \texttt{region=torso} setting evaluates the torso joints predicted by Stage~1, while the \texttt{region=non-torso} setting evaluates the non-torso joints refined by Stage~2. Under this split, Stage~1 improves all scores over GVHMR across the torso subset, suggesting that the deterministic stage is well matched to the more stable structural variables. On the non-torso subset, our method achieves the best MPJPE on all three datasets, consistent with the full camera-space results in Table~\ref{tab:camera_results}. This pattern matches the intended division of labor in the model: deterministic estimation handles the well-constrained torso-root structure, while probabilistic generation is more useful for the more ambiguous non-torso distal articulation.

\begin{table*}
  \caption{Regional camera-space evaluation against GVHMR. The left table reports the \texttt{torso} region predicted by Stage~1. The right table reports the \texttt{non-torso} region refined by Stage~2.}
  \label{tab:stage1_camera_results}
  \centering
  \scriptsize
  \begin{minipage}[t]{0.48\textwidth}
    \centering
    \textbf{Torso setup}

    \vspace{2pt}
    \setlength{\tabcolsep}{2.8pt}
    \resizebox{\linewidth}{!}{%
    \begin{tabular}{lcccccc}
      \toprule
      & \multicolumn{2}{c}{3DPW} & \multicolumn{2}{c}{RICH} & \multicolumn{2}{c}{EMDB} \\
      \cmidrule(lr){2-3} \cmidrule(lr){4-5} \cmidrule(l){6-7}
      Models & \shortstack{PA-MPJPE $\downarrow$} & MPJPE $\downarrow$ & \shortstack{PA-MPJPE $\downarrow$} & MPJPE $\downarrow$ & \shortstack{PA-MPJPE $\downarrow$} & MPJPE $\downarrow$ \\
      \midrule
      GVHMR \cite{shen2024gvhmr} & 5.96 & 27.31 & 9.03 & 36.79 & 14.96 & 51.56 \\
      Ours & \textbf{5.72} & \textbf{27.21} & \textbf{9.00} & \textbf{35.82} & \textbf{14.26} & \textbf{50.03} \\
      \bottomrule
    \end{tabular}%
    }
  \end{minipage}
  \hfill
  \begin{minipage}[t]{0.48\textwidth}
    \centering
    \textbf{Non-torso setup}

    \vspace{2pt}
    \setlength{\tabcolsep}{2.8pt}
    \resizebox{\linewidth}{!}{%
    \begin{tabular}{lcccccc}
      \toprule
      & \multicolumn{2}{c}{3DPW} & \multicolumn{2}{c}{RICH} & \multicolumn{2}{c}{EMDB} \\
      \cmidrule(lr){2-3} \cmidrule(lr){4-5} \cmidrule(l){6-7}
      Models & \shortstack{PA-MPJPE $\downarrow$} & MPJPE $\downarrow$ & \shortstack{PA-MPJPE $\downarrow$} & MPJPE $\downarrow$ & \shortstack{PA-MPJPE $\downarrow$} & MPJPE $\downarrow$ \\
      \midrule
      GVHMR \cite{shen2024gvhmr} & \textbf{39.5} & 64.9 & 45.6 & 80.7 & \textbf{51.6} & 85.4 \\
      Ours & 39.6 & \textbf{64.7} & \textbf{45.4} & \textbf{79.6} & 53.0 & \textbf{84.6} \\
      \bottomrule
    \end{tabular}%
    }
  \end{minipage}
\end{table*}

\paragraph{Occlusion-specific Benchmark.}

Tab.~\ref{tab:xocc_results} reports the camera-space metrics on 3DPW-XOCC \cite{li2023niki}. Our method achieves the best performance on all reported metrics in this setting, indicating that the factorized probabilistic stage is particularly beneficial when image evidence is heavily degraded, improving robustness by more reliably completing ambiguous body configurations under occlusion.

\paragraph{World-space Motion Recovery.}

In world space, the clearest gain appears on the drift-sensitive metric W-MPJPE in Tab.~\ref{tab:world_results}. Since W-MPJPE allows only a short initial alignment, the improvement suggests that our method better preserves long-horizon motion coherence rather than benefiting primarily from full-sequence registration. While the remaining world-space metrics are more mixed relative to GENMO \cite{li2025genmo} and GVHMR \cite{shen2024gvhmr}, the results consistently point to stronger drift control with broadly competitive overall performance.

{
\begin{table*}[t]
  \centering
  \begin{minipage}[t]{0.67\textwidth}\vspace{0pt}
    \centering
    \captionof{table}{World-space reconstruction on EMDB and RICH datasets.}
    \label{tab:world_results}

    \vspace{-6pt}
    \scriptsize
    \setlength{\tabcolsep}{1.8pt}
    \resizebox{\linewidth}{!}{%
    \begin{tabular}{lccccc ccccc}
      \toprule
      & \multicolumn{5}{c}{EMDB} & \multicolumn{5}{c}{RICH} \\
      \cmidrule(lr){2-6} \cmidrule(l){7-11}
      Models & \shortstack{WA-\\MPJPE $\downarrow$} & \shortstack{W-\\MPJPE $\downarrow$} & \shortstack{Root\\translation $\downarrow$} & \shortstack{Jitter\\$\downarrow$} & \shortstack{Foot-\\sliding $\downarrow$} & \shortstack{WA-\\MPJPE $\downarrow$} & \shortstack{W-\\MPJPE $\downarrow$} & \shortstack{Root\\translation $\downarrow$} & \shortstack{Jitter\\$\downarrow$} & \shortstack{Foot-\\sliding $\downarrow$} \\
      \midrule
      TRACE \cite{sun2023trace} & 529.0 & 1702.3 & 17.7 & 2987.6 & 370.7 & 238.1 & 925.4 & 610.4 & 1578.6 & 230.7 \\
      GLAMR & 280.8 & 726.6 & 11.4 & 46.3 & 20.7 & 129.4 & 236.2 & 3.8 & 49.7 & 18.1 \\
      SLAHMR & 326.9 & 776.1 & 10.2 & 31.3 & 14.5 & 98.1 & 186.4 & 28.9 & 34.3 & 5.1 \\
      WHAM \cite{shin2023wham} & 135.6 & 354.8 & 6.0 & 22.5 & \underline{4.4} & 109.9 & 184.6 & 4.1 & 19.7 & \underline{3.3} \\
      GVHMR \cite{shen2024gvhmr} & 111.0 & 276.5 & 2.0 & \underline{16.7} & \textbf{3.5} & \textbf{78.8} & \underline{126.3} & \textbf{2.4} & \underline{12.8} & \underline{3.0} \\
      TRAM \cite{wang2024tram} & 76.4 & 222.4 & \underline{1.4} & 18.1 & 11.0 & -- & -- & -- & -- & -- \\
      GENMO \cite{li2025genmo} & \textbf{65.1} & \underline{210.9} & \textbf{1.0} & \textbf{9.58} & 8.3 & \underline{80.7} & 127.2 & 2.6 & \textbf{8.6} & \textbf{2.7} \\
      \midrule
      Ours & \underline{70.5} & \textbf{192.5} & 1.5 & 17.7 & 9.3 & 86.6 & \textbf{123.3} & \underline{2.5} & 17.7 & 8.6 \\
      \bottomrule
    \end{tabular}%
    }
  \end{minipage}
  \hfill
  \begin{minipage}[t]{0.3\textwidth}
    \centering
    \captionof{table}{Occlusion-specific evaluation on the 3DPW-XOCC dataset.}
    \label{tab:xocc_results}

    \vspace{-2pt}
    \scriptsize
    \setlength{\tabcolsep}{1.8pt}
    \resizebox{\linewidth}{!}{%
    \begin{tabular}{lccc}
      \toprule
      Methods & MPJPE $\downarrow$ & PA-MPJPE $\downarrow$ & PVE $\downarrow$ \\
      \midrule
      HybrIK \cite{li2021hybrik} & 148.3 & 98.7 & 164.5 \\
      PARE \cite{kocabas2021pare} & 114.2 & 67.7 & 133.0 \\
      PARE + VIBE \cite{kocabas2021pare,kocabas2020vibe} & 97.3 & 60.2 & 114.9 \\
      NIKI (frame-based) \cite{li2023niki} & 110.7 & 60.5 & 128.6\\
      NIKI (temporal) \cite{li2023niki} & 88.9 & 52.1 & 98.0 \\
      GENMO \cite{li2025genmo} & 76.2 & 48.4 & 94.2 \\
      \midrule
      Ours & \textbf{66.3} &\textbf{ 45.1} & \textbf{81.2} \\
      \bottomrule
    \end{tabular}%
    }
  \end{minipage}
  \vspace{-16pt}
\end{table*}
}

\subsection{Ablation studies}

\begin{figure}[t]
  \centering

  \begin{minipage}[t]{0.57\linewidth}\vspace{9pt}
    \centering
    \fontsize{9}{10}\selectfont
    \resizebox{\linewidth}{!}{%
    \begin{tabular}{p{0.48\linewidth}cccc}
      \toprule
      Variant & PA-MPJPE $\downarrow$ & MPJPE $\downarrow$ & PVE $\downarrow$  \\
      \midrule
      Stage 1 Model & 65.0 & 77.0 & 122.4 \\
      + Stage 2 Model & 39.9 & 60.3 & 73.3 \\
      + Composite Representation & 39.4 & 59.2 & 71.8 \\
      + Geometry-aware losses & 37.2 & 57.2 & 69.2 \\
      + CFG scaling & 37.1 & 56.5 & 68.5  \\
      + synthetic dataset & 36.9 & 56.2 & 68.3 \\
      \bottomrule
    \end{tabular}%
    }
  \end{minipage}
  \hfill
  \begin{minipage}[t]{0.42\linewidth}
    \centering
    \begin{minipage}[t]{0.24\linewidth}
      \parbox[t][8pt][c]{\linewidth}{\centering\scriptsize Input}
      \includegraphics[width=\linewidth]{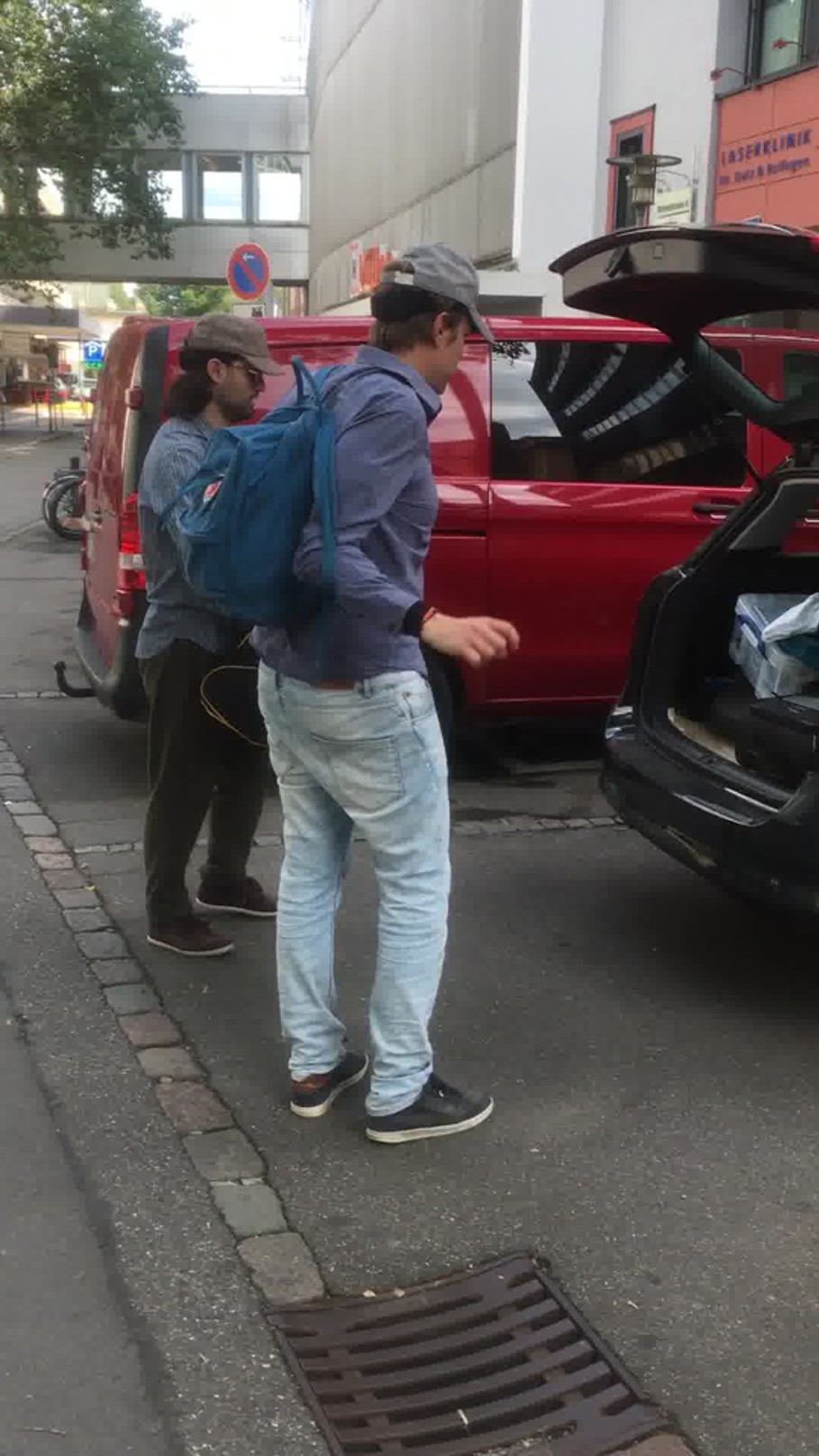}\\[-1pt]
    \end{minipage}\hfill
    \begin{minipage}[t]{0.24\linewidth}
      \parbox[t][8pt][c]{\linewidth}{\centering\scriptsize GT}
      \includegraphics[width=\linewidth]{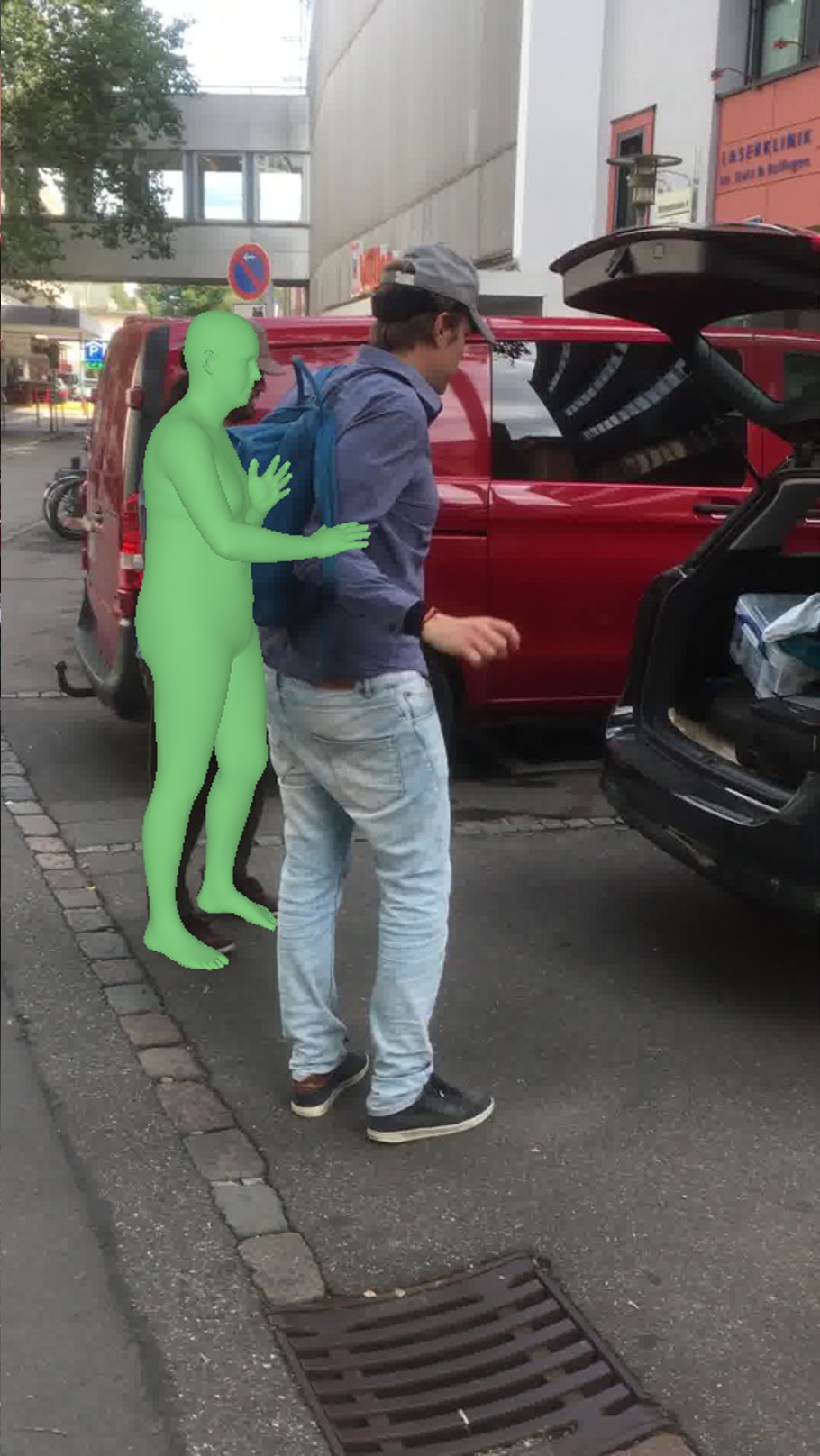}\\[-1pt]
    \end{minipage}\hfill
    \begin{minipage}[t]{0.24\linewidth}
      \parbox[t][8pt][c]{\linewidth}{\centering\scriptsize GVHMR}
      \includegraphics[width=\linewidth]{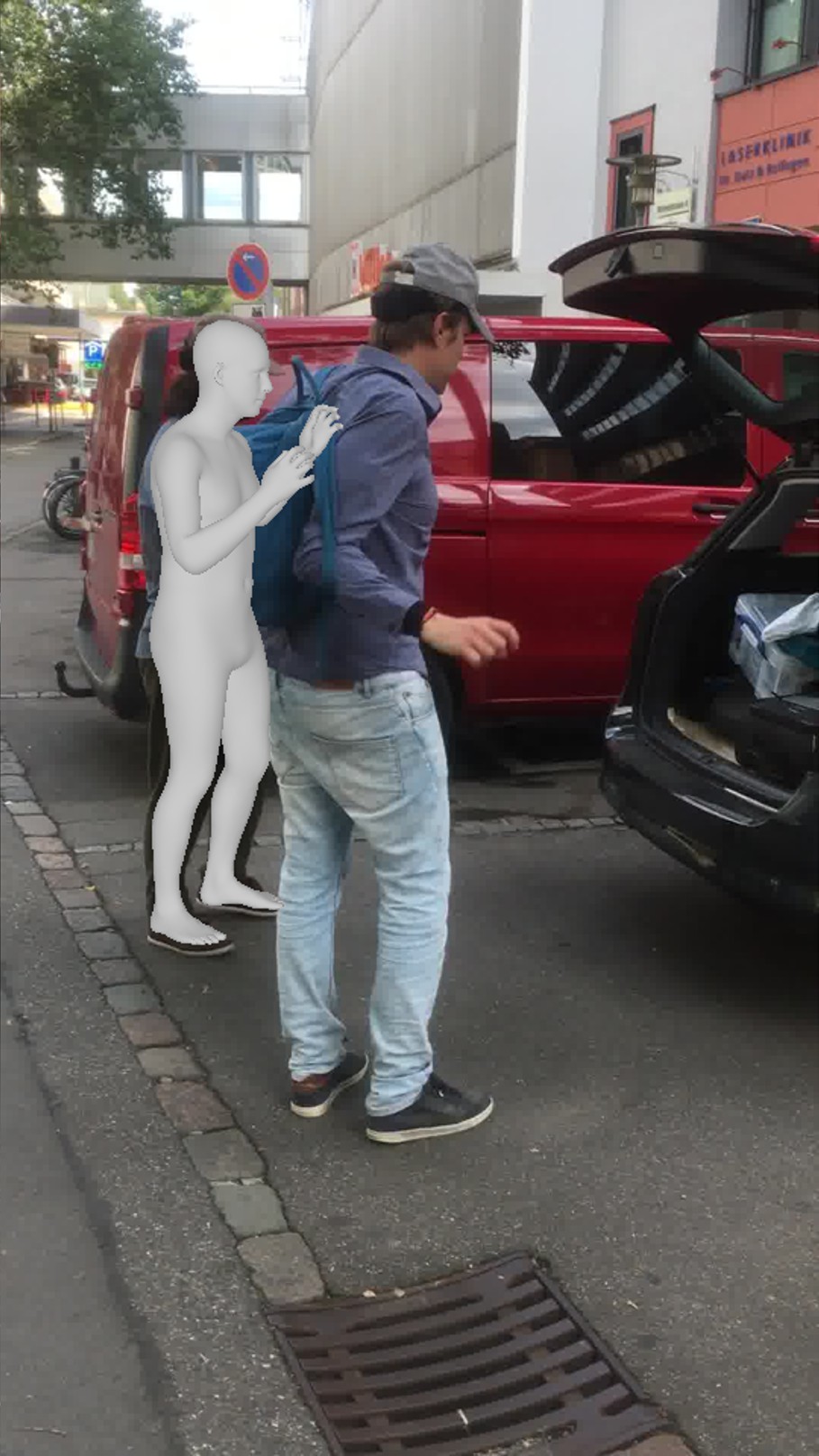}\\[-1pt]
    \end{minipage}\hfill
    \begin{minipage}[t]{0.24\linewidth}
      \parbox[t][8pt][c]{\linewidth}{\centering\scriptsize Ours}
      \includegraphics[width=\linewidth]{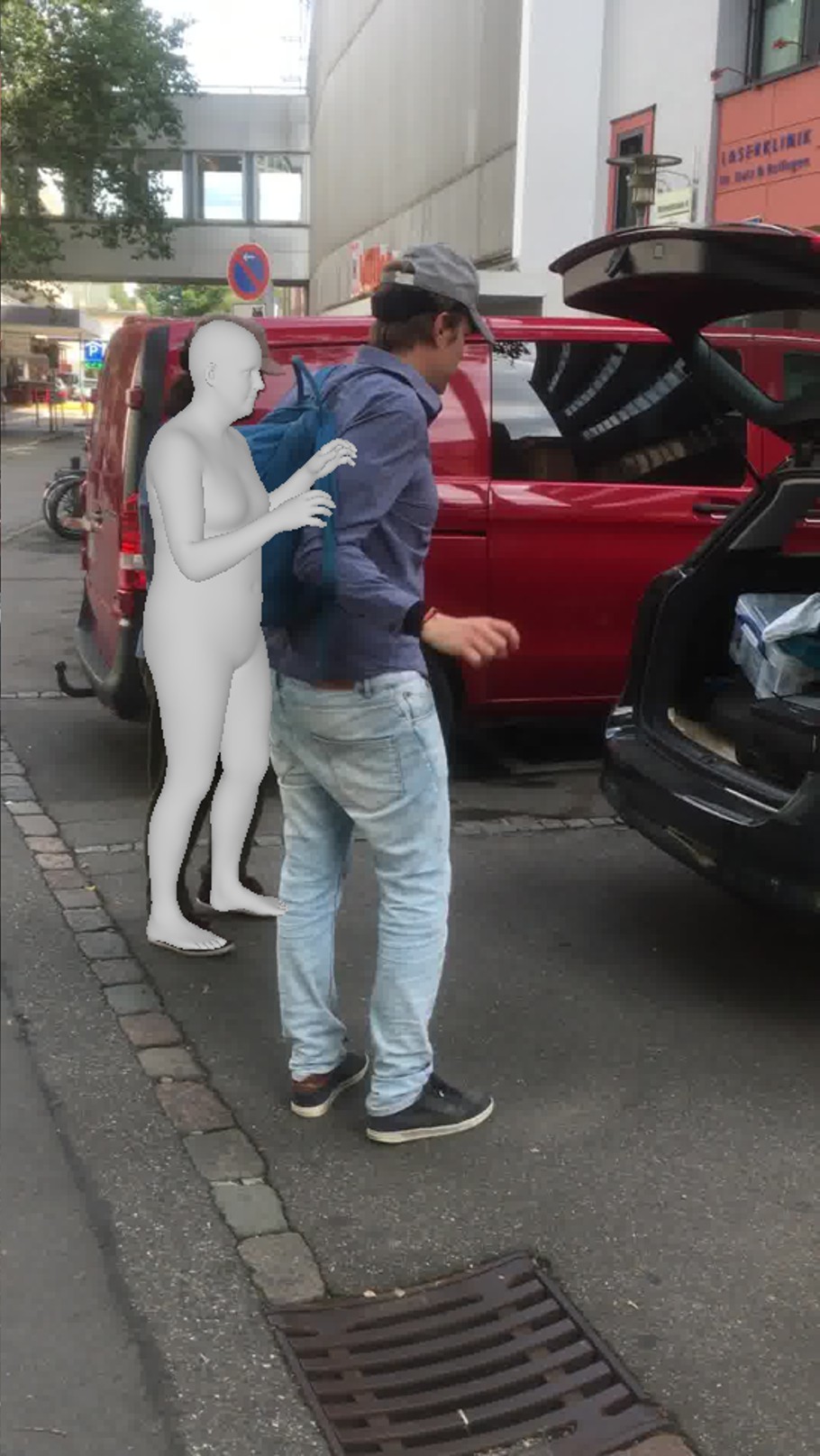}\\[-1pt]
    \end{minipage}
  \end{minipage}

  \vspace{-8pt}
  
  \begin{minipage}[t]{0.57\linewidth}
    \captionof{table}{Progressive ablation on camera-space reconstruction metrics, compared on 3DPW \cite{vonmarcard20183dpw}.}
    \label{tab:ablation_results}
  \end{minipage}
  \hfill
  \begin{minipage}[t]{0.42\linewidth}
    \captionof{figure}{Body-shape recovery under heavy occlusion. FactorizedHMR better preserves body volume.}
    \label{fig:bodyshape}
  \end{minipage}
  \vspace{-18pt}
\end{figure}

Tab.~\ref{tab:ablation_results} presents a progressive ablation on 3DPW \cite{vonmarcard20183dpw} using camera-space metrics. Note that the Stage~1 model alone predicts only the torso anchor, so Stage~2 is required to recover a complete body pose. Adding Stage~2 yields the largest improvement, showing that the torso-root anchor
alone is insufficient for full-body recovery. Composite representation and geometry-aware losses then provide consistent gains across all three metrics, and CFG scaling yields an additional reduction on each metric. Finally, adding the synthetic dataset brings a smaller but still consistent improvement, indicating that its contribution is complementary to the architectural and objective-level changes. Overall, the table supports the importance of Stage~2, composite representation, geometry-aware training, guidance scaling, and synthetic data for improving video HMR. We further include a controlled deterministic--probabilistic diagnostic in the \textbf{Appendix}.

\textbf{Runtime Analysis.} Compared with GVHMR on a 1193-frame sequence, our full two-stage model requires 3.92 $\pm$ 0.03 seconds per sequence versus 0.41 $\pm$ 0.01 seconds for GVHMR, while using 1057.0 MB versus 303.6 MB of peak allocated GPU memory. All models were tested on an NVIDIA RTX A5000 GPU, with denoising steps of 50. Still, the full pipeline runs at 304.45 frames per second, and can be reduced directly by using fewer denoising steps at inference time.

\begin{figure}[t]
    \centering

    \begin{minipage}[t]{0.99\linewidth}
        \centering
        \begin{minipage}[t]{0.24\linewidth}
            {\scriptsize GT}
            \vspace{6pt}
            \centering
            \includegraphics[width=\linewidth]{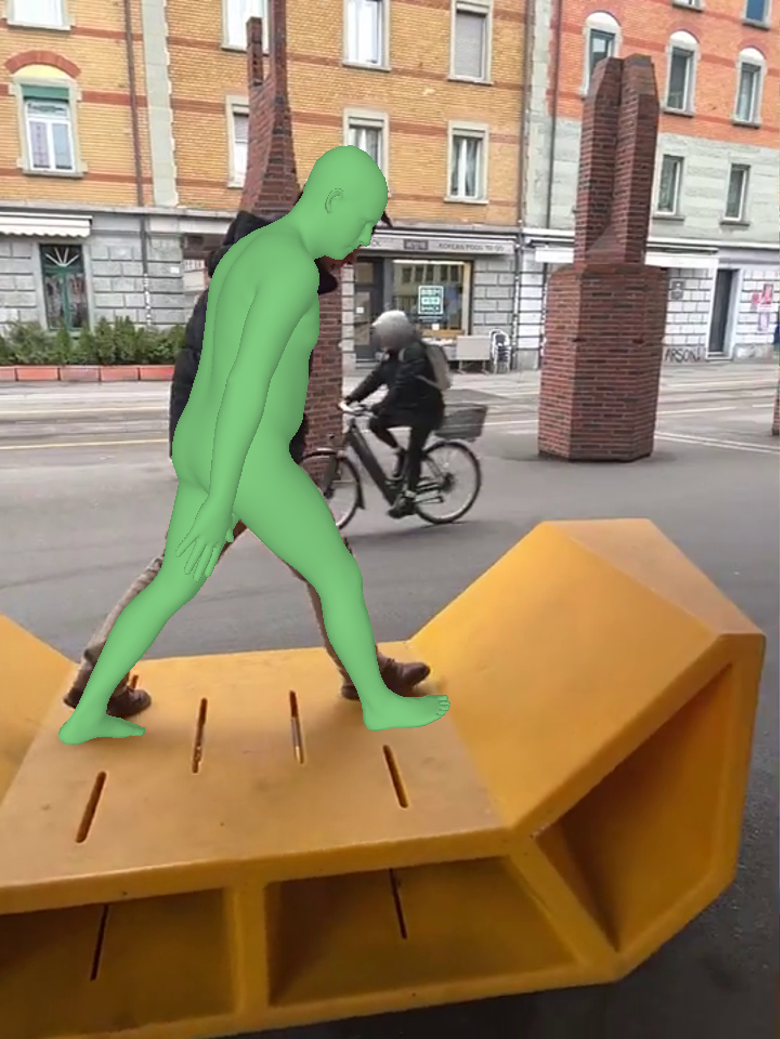}\\[-1pt]
        \end{minipage}\hfill
        \begin{minipage}[t]{0.24\linewidth}
            {\scriptsize GVHMR}
            \vspace{6pt}
            \centering
            \includegraphics[width=\linewidth]{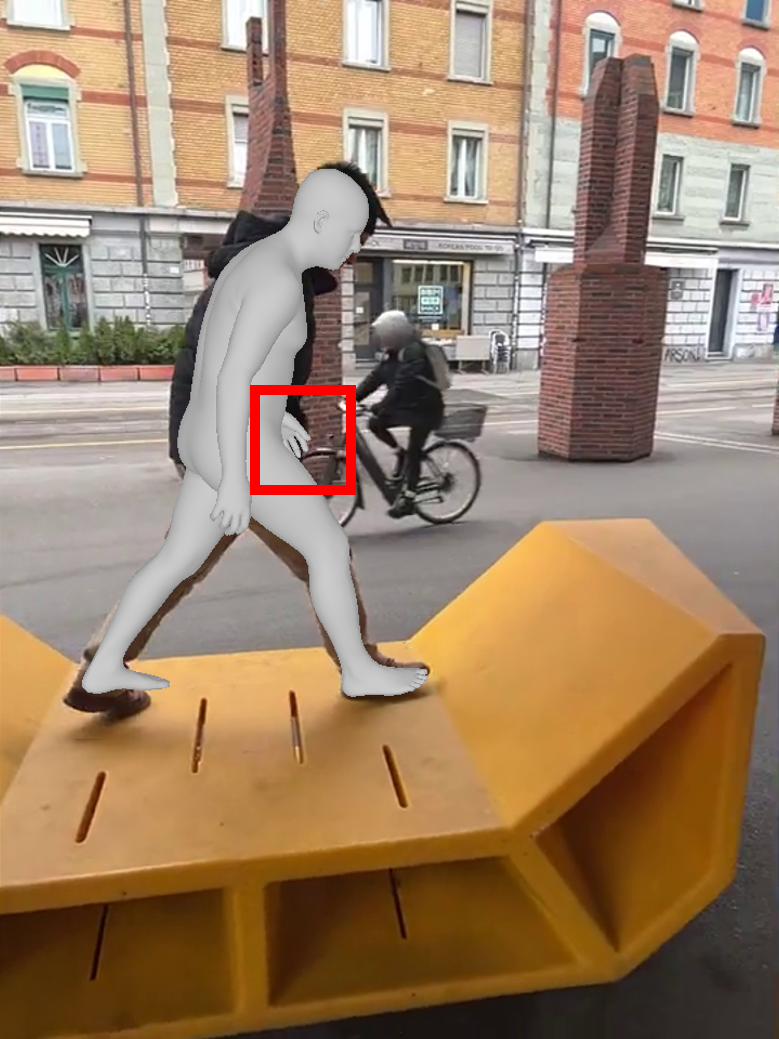}\\[-1pt]
        \end{minipage}\hfill
        \begin{minipage}[t]{0.24\linewidth}
            {\scriptsize GENMO}
            \vspace{6pt}
            \centering
            \includegraphics[width=\linewidth]{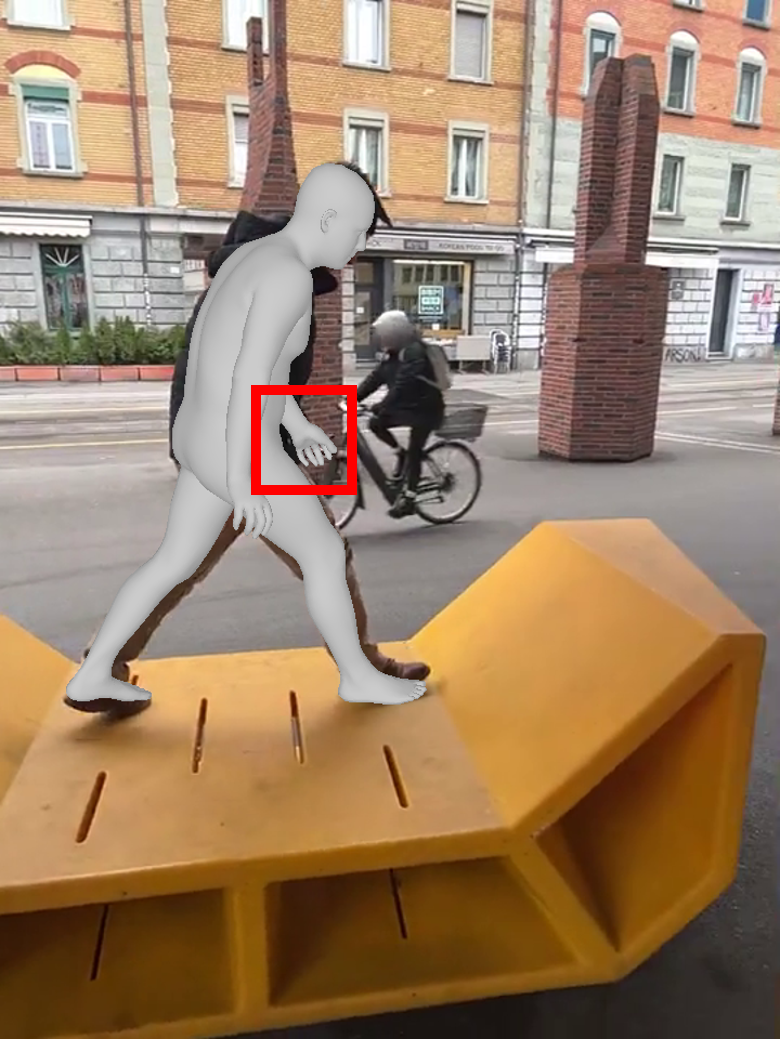}\\[-1pt]
        \end{minipage}\hfill
        \begin{minipage}[t]{0.24\linewidth}
            {\scriptsize Ours}
            \vspace{6pt}
            \centering
            \includegraphics[width=\linewidth]{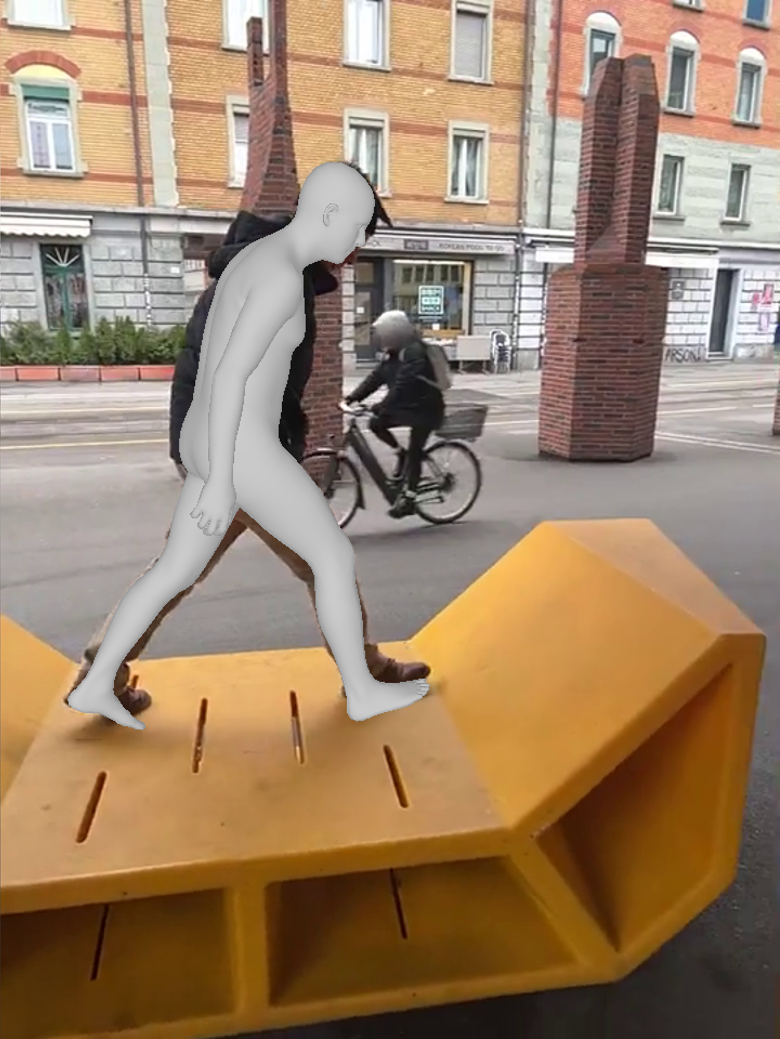}\\[-1pt]
        \end{minipage}
    \end{minipage}
    \hfill

    \vspace{0.1em}

    \begin{minipage}[t]{0.49\linewidth}
        \centering
        \begin{minipage}[t]{0.33\linewidth}
            {\scriptsize GT}
            \centering
            \includegraphics[width=\linewidth]{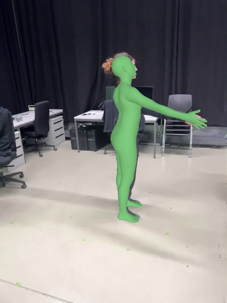}\\[-1pt]
        \end{minipage}\hfill
        \begin{minipage}[t]{0.33\linewidth}
            {\scriptsize GVHMR}
            \centering
            \includegraphics[width=\linewidth]{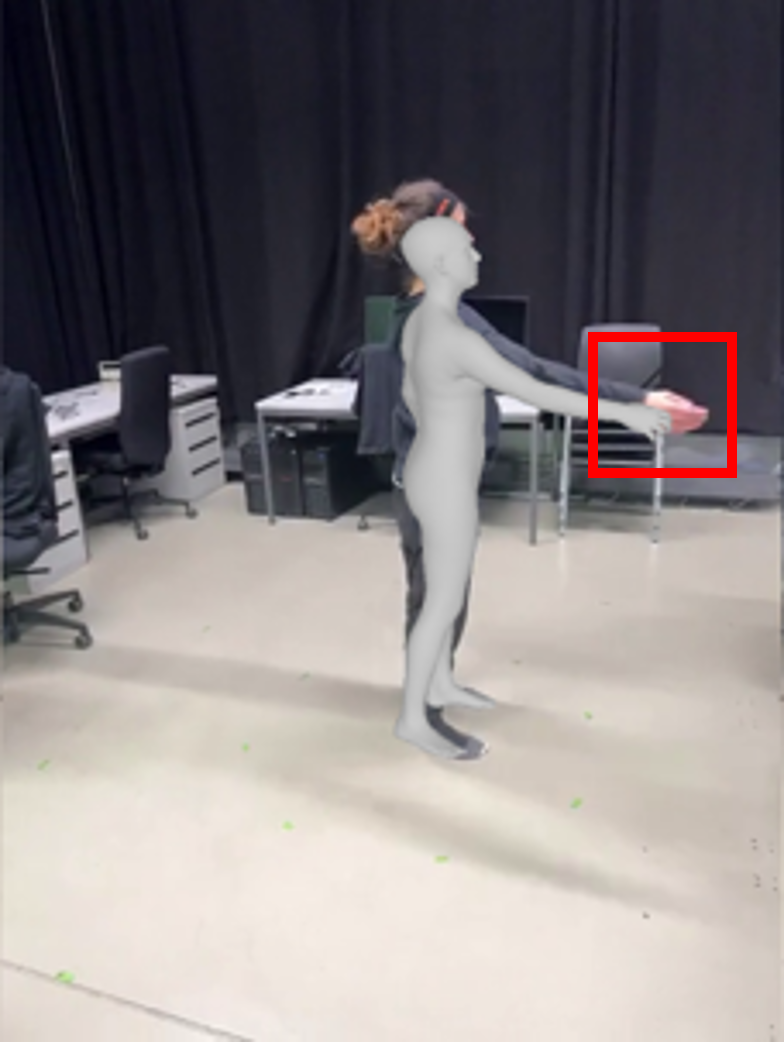}\\[-1pt]
        \end{minipage}\hfill
        \begin{minipage}[t]{0.33\linewidth}
            {\scriptsize Ours}
            \centering
            \includegraphics[width=\linewidth]{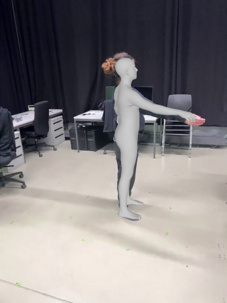}\\[-1pt]
        \end{minipage}
    \end{minipage}
    \hfill
    \begin{minipage}[t]{0.49\linewidth}
        \centering
        \begin{minipage}[t]{0.33\linewidth}
            {\scriptsize GT}
            \centering
            \includegraphics[width=\linewidth]{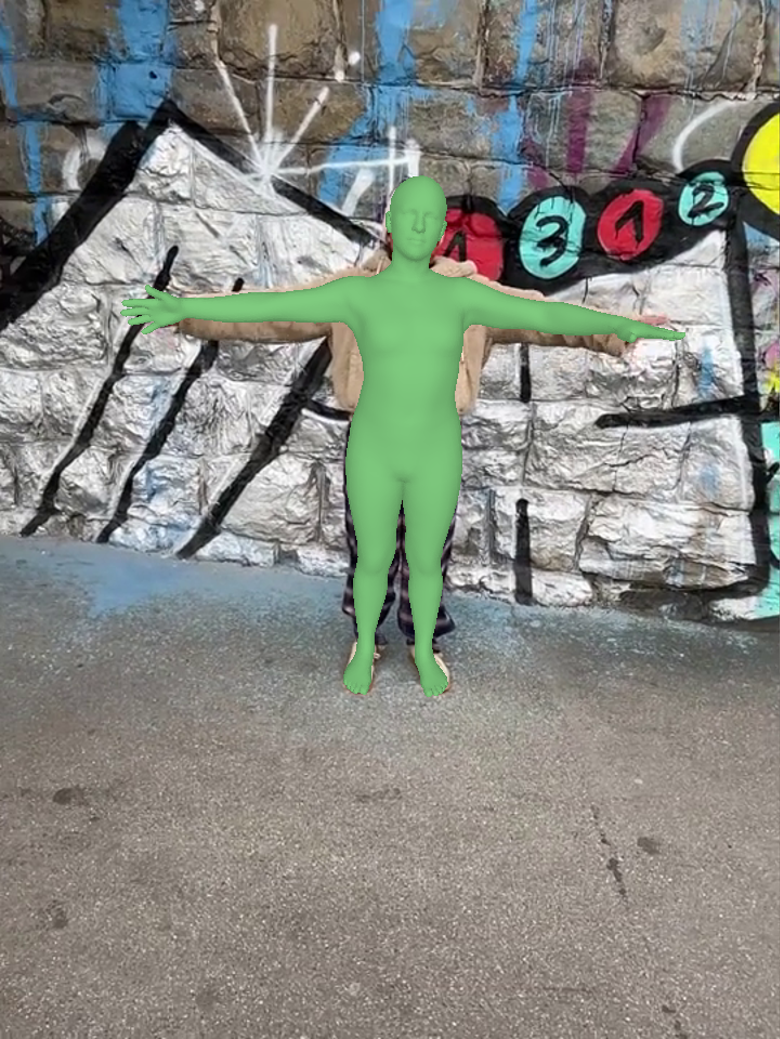}\\[-1pt]
        \end{minipage}\hfill
        \begin{minipage}[t]{0.33\linewidth}
            {\scriptsize GVHMR}
            \centering
            \includegraphics[width=\linewidth]{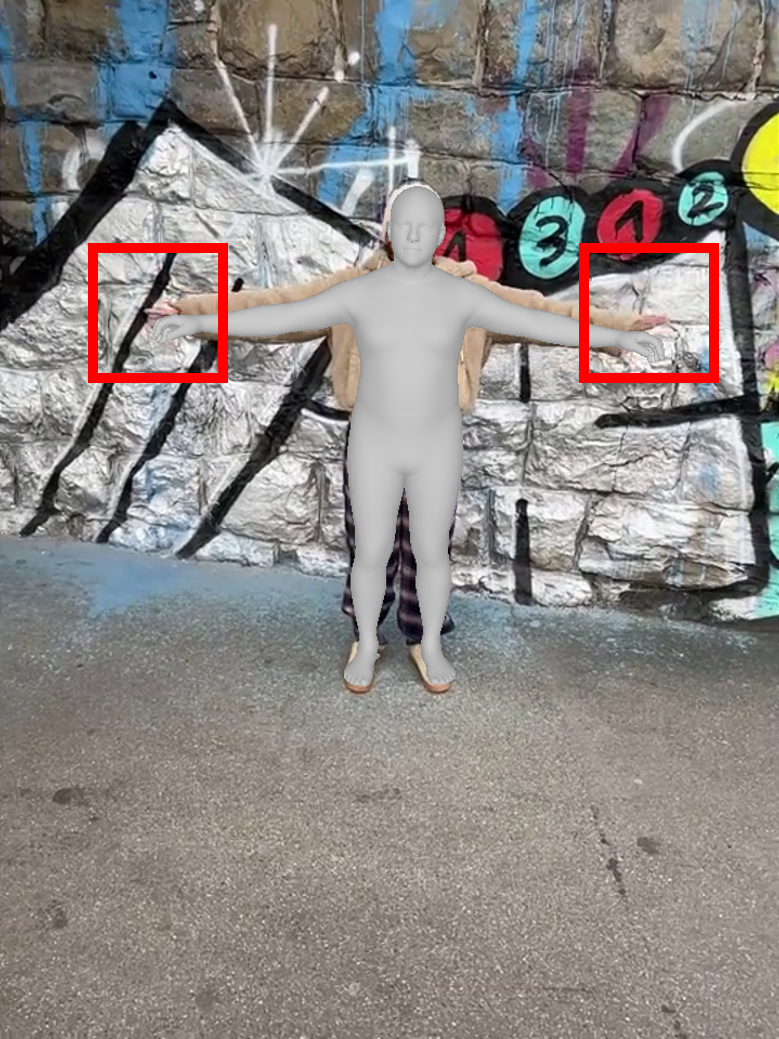}\\[-1pt]
        \end{minipage}\hfill
        \begin{minipage}[t]{0.33\linewidth}
            {\scriptsize Ours}
            \centering
            \includegraphics[width=\linewidth]{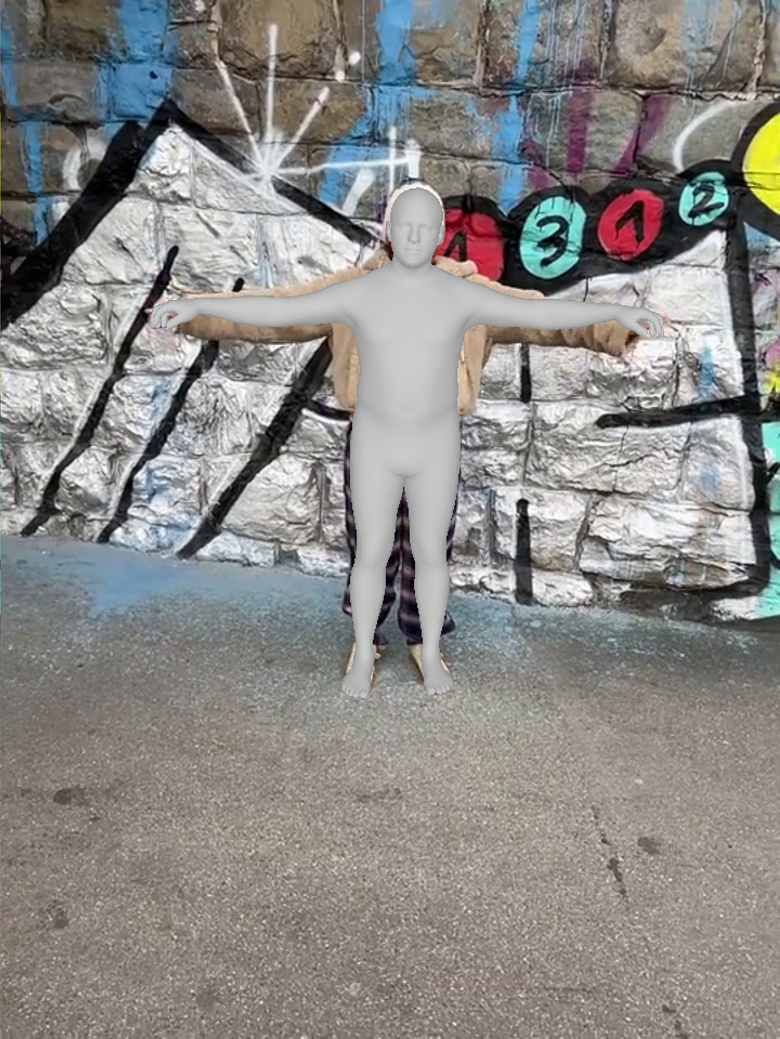}\\[-1pt]
        \end{minipage}
    \end{minipage}

    \caption{Body-pose recovery in ambiguity-heavy scenes. The red squares in the GVHMR \cite{shen2024gvhmr} and GENMO \cite{li2025genmo} results indicate incorrectly predicted body poses. }
    \label{fig:posecompare}
    \vspace{-16pt}
\end{figure}

\paragraph{Qualitative Results.}

The qualitative examples visualize failure modes for previous benchmarks, where body poses are hallucinated due to the tendencies of deterministic pipelines to average multiple possibilities. Fig.~\ref{fig:bodyshape} shows that our method better preserves body volume under missing evidence, while Fig.~\ref{fig:posecompare} shows more accurate distal body poses, consistent with the given visual evidence.

\paragraph{Limitations and Future Work.}

The current uncertainty partition between Stage~1 and Stage~2 is fixed, which can still fail under extreme truncation or rapid camera motion: if the Stage~1 structural anchor is biased, Stage~2 may inherit that error rather than correct it. In addition, probabilistic refinement increases inference cost, which limits practical deployment. Future work could make the factorization adaptive, improve sampling efficiency, and develop ambiguity-aware benchmarks for calibrated multi-hypothesis video HMR.

\section{Conclusions}
\label{sec:conclusions}

We presented FactorizedHMR, a video HMR framework that uses an uncertainty-aware factorization: a deterministic stage estimates a stable torso-root anchor, and a probabilistic flow-matching stage completes ambiguity-prone non-torso articulation and world motion. This design, together with geometry-aware refinement and camera-aware synthetic supervision, improves reconstruction while preserving stability. Across camera-space and world-space benchmarks, FactorizedHMR remains strong against both deterministic and generative baselines and displays clear qualitative advantages in ambiguous scenes. More broadly, our results suggest that probabilistic completion in HMR is most useful when targeted to ambiguity-prone variables, and future work can further improve joint refinement while developing benchmarks for plausible multi-solution recovery under occlusion.

\newpage 



\medskip

{
    \small
    \bibliographystyle{ieeenat_fullname}
    \bibliography{main}
}


\appendix

\section{Technical appendices and supplementary material}

Section~\ref{app:metrics} summarizes the evaluation metrics and sampling protocols used in our experiments. Section~\ref{app:per_joint_error} provides a per-joint baseline error analysis to further motivate the uncertainty factorization in our method. Section~\ref{app:dataset} describes the synthetic data generation pipeline in greater detail and includes additional qualitative comparisons that compare our method with other benchmarks. Section ~\ref{app:social} discusses the potential positive and negative societal impacts of our project.

\subsection{Evaluation}
\label{app:metrics}

For camera-space reconstruction, MPJPE and PVE are computed after pelvis alignment, while PA-MPJPE additionally applies a similarity transform. For world-space reconstruction, we follow the standard 100-frame chunk protocol used in recent work \cite{shin2023wham,wang2024tram,shen2024gvhmr,li2025genmo}. WA-MPJPE aligns the full chunk, whereas W-MPJPE aligns only the first two frames and therefore penalizes subsequent drift. We further report root translation error (RTE), computed after rigid alignment and normalized by the total trajectory displacement, together with jitter and foot sliding, which measure higher-order temporal smoothness and predicted foot motion on ground-truth contact frames \cite{shin2023wham,shen2024gvhmr,li2025genmo}. 

\begin{figure}[!b]
  \centering
  \captionsetup{
    skip=0pt,
    belowskip=-8pt,
    aboveskip=0pt
  }

  \begin{minipage}[t]{0.49\linewidth}
    \begin{minipage}[t]{\linewidth}
      \centering
      \caption{Effect of the number of ODE sampling steps on MPJPE and WA-MPJPE metrics. Both metrics improve rapidly from very small step counts and then saturate, with only minor differences beyond roughly 20 steps. We use 50 steps as a conservative near-converged default.}
      \label{fig:ablation_steps}
    \end{minipage}
    \centering
    \vspace{8pt}
    \includegraphics[width=\linewidth]{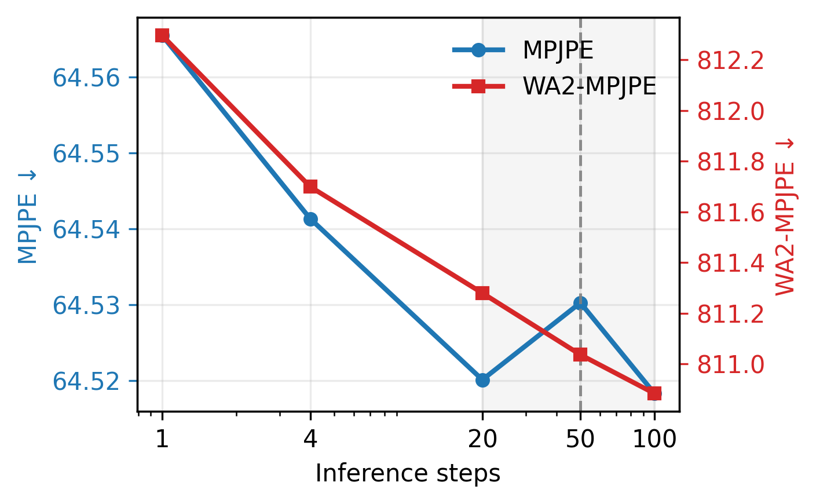}
  \end{minipage}
  \hfill
  \begin{minipage}[t]{0.49\linewidth}
    \begin{minipage}[t]{\linewidth}
      \centering
      \captionof{figure}{Example of 3DPW-XOCC's occluded frames (right) compared to the original 3DPW dataset (left).}
      \label{fig:3dpwxocc_example}
    \end{minipage}
    \vspace{4pt}
    \centering
    \includegraphics[width=0.7\linewidth]{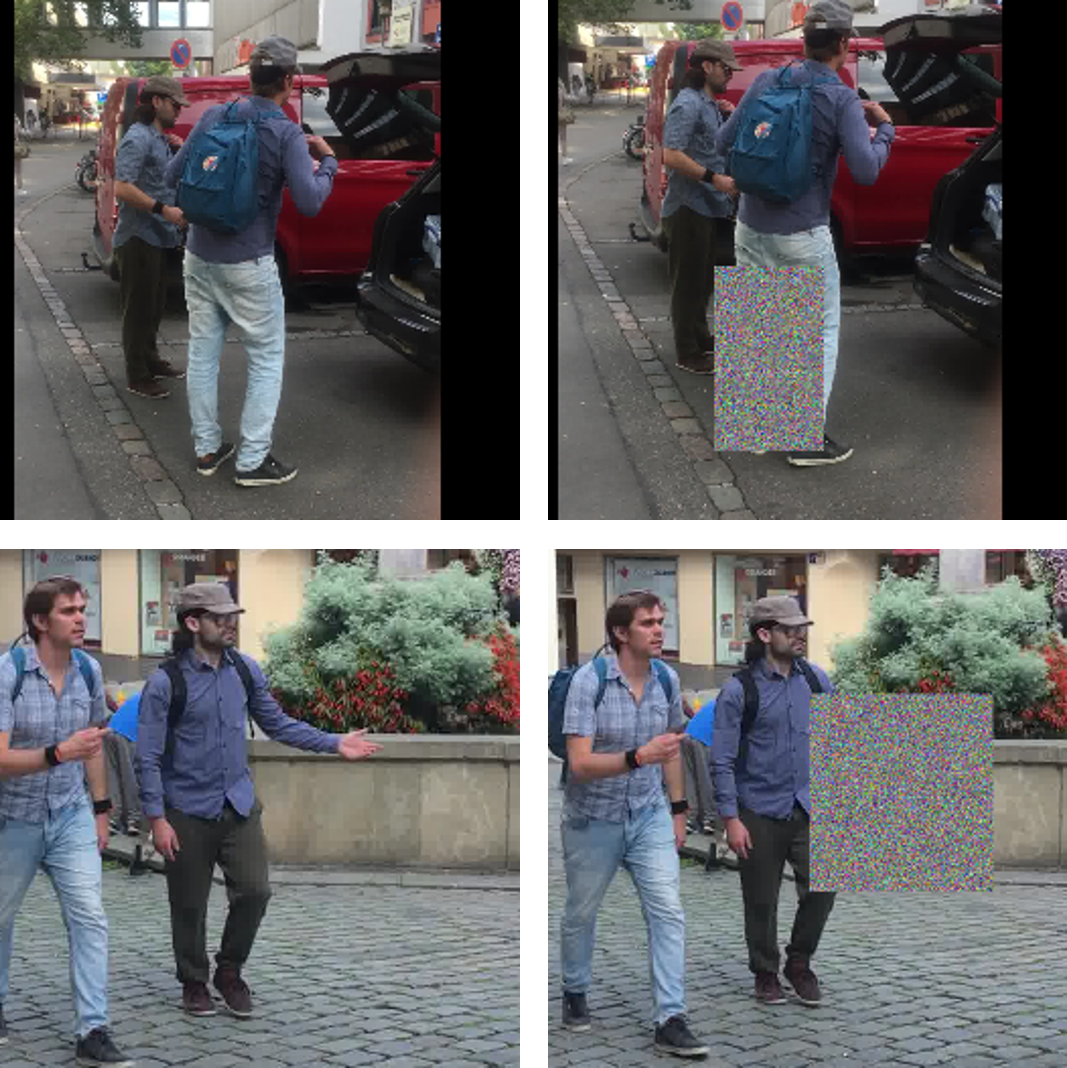}
  \end{minipage}
\end{figure}

For Stage~2 inference, we use the flow-matching ODE sampler with the default Euler solver and 50 integration steps (i.e., approximately 50 NFEs) unless otherwise noted. We apply classifier-free guidance during sampling, and in practice, we select the CFG scale from a narrow range of 1.25 to 1.75, with 1.5 used as the default operating point in most experiments. 
We include an ablation study that compares the number of ODE integration steps during inference at Fig.~\ref{fig:ablation_steps}. Performance improves from very small step counts to moderate ones, then quickly saturates. The differences among 20, 50, and 100 steps are minor, and no single setting is uniformly best across all metrics: 20 steps gives the best PA-MPJPE, while 100 steps is best on MPJPE, PVE, and world-space metrics. We therefore use 50 steps as a conservative near-converged default for the main experiments.


\subsection{Per-joint baseline error analysis}
\label{app:per_joint_error}

To further motivate the proposed factorization, Table~\ref{tab:gvhmr_per_joint_mpJPE} summarizes per-joint MPJPE of GVHMR across 3DPW, EMDB, and RICH. The table separates relatively stable torso joints with non-torso joints to examine whether the baseline error pattern is consistent with the non-uniform uncertainty assumption behind our method. Because the joint information used for evaluation differs by dataset (3DPW : 14, EMDB-1 and RICH : 24), joints that don't exist in 3DPW are marked as blank.

The results show a clear and consistent separation between these regions. Averaged over datasets, the torso-root anchor subset has a mean MPJPE of 39.83, whereas the distal subset reaches 84.58, with dataset-wise gaps of 32.68 on 3DPW, 50.01 on EMDB-1, and 51.55 on RICH. This supports our design choice to treat torso-root variables as a more stable structural anchor and to reserve probabilistic modeling for the more ambiguous non-torso articulation.

\subsection{Synthetic data details}
\label{app:dataset}

We generate 340 synthetic training videos from AMASS motion clips, each converted to a 120-frame SMPL-X sequence with known camera intrinsics and a sampled camera trajectory. In our implementation, camera motion is parameterized by yaw, pitch, roll, and translation components $(t_x,t_y,t_z)$ and rendered into paired control signals including body renderings, hand renderings, masks, first-frame OpenPose renderings, and point-cloud trajectory renderings. Identity and appearance are produced from a curated prompt bank of 10 identity/clothing prompts and 10 scene prompts, combined randomly for each clip. Occlusion and truncation are not injected by separate synthetic occluders; instead, they arise naturally from body self-occlusion, camera viewpoint, and framing under the sampled trajectories. We use the generated videos as paired observations with fixed pre-generation SMPL-X motion and camera labels.

The synthetic set is used only for training and is concatenated with the training datasets \cite{mahmood2019amass, black2023bedlam, Ionescu2014Human36MLS, vonmarcard20183dpw}. Its purpose is to improve camera-aware structural supervision rather than to serve as a standalone benchmark or replacement for real video. We will release the dataset generation code, subject to the licenses of the upstream components used in generation; in particular, release and redistribution are constrained by the terms of FLUX.1 \cite{flux2024}, Uni3C \cite{cao2025uni3c}, and Depth Pro \cite{Bochkovskii2024}, which we review explicitly in the project release.

\begin{figure}[!b]
  \centering
  \captionsetup{
    skip=2pt,
    belowskip=0pt,
    aboveskip=0pt
  }

  \newlength{\capboxheight}
  \setlength{\capboxheight}{1.25cm}

  \begin{minipage}[t]{\linewidth}
    \begin{minipage}[t][\capboxheight][t]{\linewidth}
      \centering
      \captionof{table}{Per-joint MPJPE calculated by GVHMR \cite{shen2024gvhmr} across test datasets. Lower values indicate joints that are more reliably recovered by a strong deterministic baseline.}
      \label{tab:gvhmr_per_joint_mpJPE}
    \end{minipage}

    \vspace{-4pt}
    \centering
    \scriptsize
    \renewcommand{\arraystretch}{0.92}
    \setlength{\tabcolsep}{4pt}
    \resizebox{0.6\linewidth}{!}{%
      \begin{tabular}{lcccc}
        \toprule
        Joint group / joint & 3DPW & EMDB-1 & RICH & Mean \\
        \midrule
        \multicolumn{5}{l}{\emph{Torso joints}} \\
        Pelvis / root & -- & 14.71 & 10.74 & 12.73 \\
        Left hip & 16.97 & 9.32 & 8.37 & 11.55 \\
        Right hip & 16.97 & 9.32 & 8.37 & 11.55 \\
        Spine / torso & -- & 55.04 & 35.71 & 45.38 \\
        Neck & 47.98 & 78.48 & 62.67 & 63.04 \\
        Left shoulder & 52.20 & 76.67 & 59.89 & 62.92 \\
        Right shoulder & 48.17 & 75.60 & 55.66 & 59.81 \\
        \midrule
        \multicolumn{5}{l}{\emph{Non-torso joints}} \\
        Left elbow & 69.64 & 84.90 & 73.89 & 76.14 \\
        Right elbow & 61.82 & 78.72 & 69.69 & 70.08 \\
        Left wrist & 84.70 & 109.27 & 89.17 & 94.38 \\
        Right wrist & 82.34 & 99.39 & 86.99 & 89.57 \\
        Left hand & -- & 125.99 & 101.72 & 113.86 \\
        Right hand & -- & 115.88 & 99.08 & 107.48 \\
        Left knee & 46.53 & 55.66 & 66.98 & 56.39 \\
        Right knee & 44.57 & 56.07 & 62.71 & 54.45 \\
        Left ankle & 83.42 & 89.24 & 108.91 & 93.86 \\
        Right ankle & 80.06 & 92.56 & 106.87 & 93.16 \\
        Left foot & -- & 104.42 & 115.53 & 109.98 \\
        Right foot & -- & 107.93 & 113.44 & 110.69 \\
        \midrule
        Anchor mean & 36.46 & 43.33 & 39.7 & 39.83 \\
        Distal mean & 69.14 & 93.34 & 91.25 & 84.58 \\
        Distal $-$ anchor & 32.68 & 50.01 & 51.55 & 44.75 \\
        \bottomrule
      \end{tabular}%
    }
  \end{minipage}
\end{figure}

\subsection{Controlled Deterministic--Probabilistic Factorization}
\label{app:controlled_factorization}

To verify FactorizedHMR's design choice of estimating the torso/root variables deterministically while generating the non-torso articulation probabilistically, we compare several matched variants on the 3DPW dataset \cite{vonmarcard20183dpw}. The variants differ in whether torso and non-torso variables are estimated deterministically or probabilistically, and whether the model is trained as a single-stage predictor with a part-aware loss.

\begin{table}[t]
\centering
\caption{
Controlled factorization diagnostic on 3DPW. Fully deterministic prediction does not model ambiguity in non-torso variables, while fully probabilistic prediction destabilizes the torso/root variables. Part-aware single-stage losses improve over naive same-paradigm settings, but do not match the proposed deterministic torso/root and probabilistic non-torso factorization.
}
\label{tab:controlled_factorization}
\resizebox{\linewidth}{!}{
\begin{tabular}{lccc}
\toprule
\textbf{Variant} & \textbf{MPJPE $\downarrow$} & \textbf{PA-MPJPE $\downarrow$} & \textbf{PVE $\downarrow$} \\
\midrule
Deterministic torso / Deterministic non-torso
& 60.2 & 40.7 & 74.0 \\
Deterministic single-stage with part-aware loss
& 58.4 & 37.8 & 70.6 \\
Probabilistic torso / Probabilistic non-torso
& 68.9 & 45.5 & 81.0 \\
Probabilistic single-stage with part-aware loss
& 59.8 & 39.5 & 74.1 \\
\textbf{Ours: Deterministic torso / Probabilistic non-torso}
& \textbf{56.2} & \textbf{36.9} & \textbf{68.3} \\
\bottomrule
\end{tabular}
}
\end{table}

As shown in Table~\ref{tab:controlled_factorization}, using one estimation
paradigm for all variables is suboptimal, as the fully deterministic variant loses the benefit of probabilistic generation for ambiguity-prone non-torso variables, whereas the fully probabilistic variant performs worse because the stable torso/root variables should act as an anchor rather than be generated stochastically. The part-aware single-stage variants show that reweighting difficult body regions is useful, but not sufficient. Our proposed factorization achieves the best result because it assigns deterministic estimation to stable variables and probabilistic generation to ambiguous variables.

\subsection{Additional Generative-HMR Baseline Context}
\label{app:generative_hmr_context}

We also compare with other probabilistic HMR pipelines where possible. These methods are not always directly matched to our video-level camera/world-space protocol, so the numbers in Table~\ref{tab:generative_baselines} should be read with the corresponding protocol caveats.

\begin{table}[t]
\centering
\caption{
Additional generative/probabilistic HMR baseline context on 3DPW. RoHM~\cite{Zhang2024RoHMRH} is a best-effort reproduction with released weights and our evaluation path, since the original work does not provide a straightforward matched evaluation path for our video-level camera-coordinate setting. ProHMR~\cite{kolotouros2021prohmr} is the official PA-MPJPE reported in its paper, not our reproduction.}
\label{tab:generative_baselines}
\begin{tabular}{lccc}
\toprule
\textbf{Method} & \textbf{MPJPE $\downarrow$} & \textbf{PA-MPJPE $\downarrow$} & \textbf{PVE $\downarrow$} \\
\midrule
RoHM & 63.9 & 41.2 & 76.0 \\
ProHMR & -- & 59.8 & -- \\
\bottomrule
\end{tabular}
\end{table}

For RoHM~\cite{Zhang2024RoHMRH}, the released weights allow evaluation, but the original codebase does not provide a direct evaluation path matching our video-level camera-coordinate protocol on 3DPW, EMDB, or RICH. We therefore report a best-effort reproduction and mark it as such. For ProHMR~\cite{kolotouros2021prohmr}, we report the official PA-MPJPE from the paper. This comparison provides context for related probabilistic HMR methods, while avoiding a claim that all methods are evaluated under a fully matched protocol.

\subsection{Broader societal impact statement}
\label{app:social}

This work on video human mesh recovery has potential positive societal impact in areas such as animation, virtual reality, sports analysis, rehabilitation, and human-computer interaction, where accurate 3D human motion understanding can support more accessible and expressive tools. At the same time, the ability to recover detailed human pose and motion from video also raises ethical concerns. In particular, similar systems could be used for surveillance, tracking, or behavioral analysis without a subject's knowledge or consent. Such technologies may also create unequal harms if they are deployed in settings involving sensitive populations or if their outputs are used to make judgments about individuals. We therefore believe that research on human motion recovery should be accompanied by careful consideration of privacy, consent, and the broader social context in which these systems may be used.

\subsection{Acknowledgments.} FactorizedHMR was supported in part by the US National Science Foundation under Award No. 2549260

\begin{figure}[t]
    \centering
    \newcommand{\apptriprowwidth}{1.20\linewidth}
    \newcommand{\apptripcolwidth}{0.53\linewidth}
    \newcommand{\apptripgap}{0.015\linewidth}

    \makebox[\linewidth][c]{%
    \begin{minipage}[t]{\apptripcolwidth}
        \centering
        \begin{minipage}[t]{0.32\linewidth}
            \parbox[t][8pt][c]{\linewidth}{\centering\scriptsize GT}\par
            \vspace{2pt}
            \includegraphics[width=\linewidth]{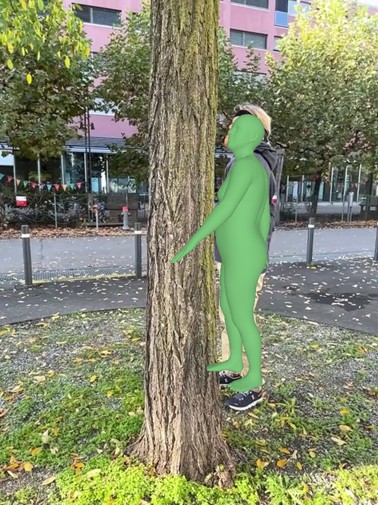}\\[-1pt]
        \end{minipage}\hfill
        \begin{minipage}[t]{0.32\linewidth}
            \parbox[t][8pt][c]{\linewidth}{\centering\scriptsize GVHMR}\par
            \vspace{2pt}
            \includegraphics[width=\linewidth]{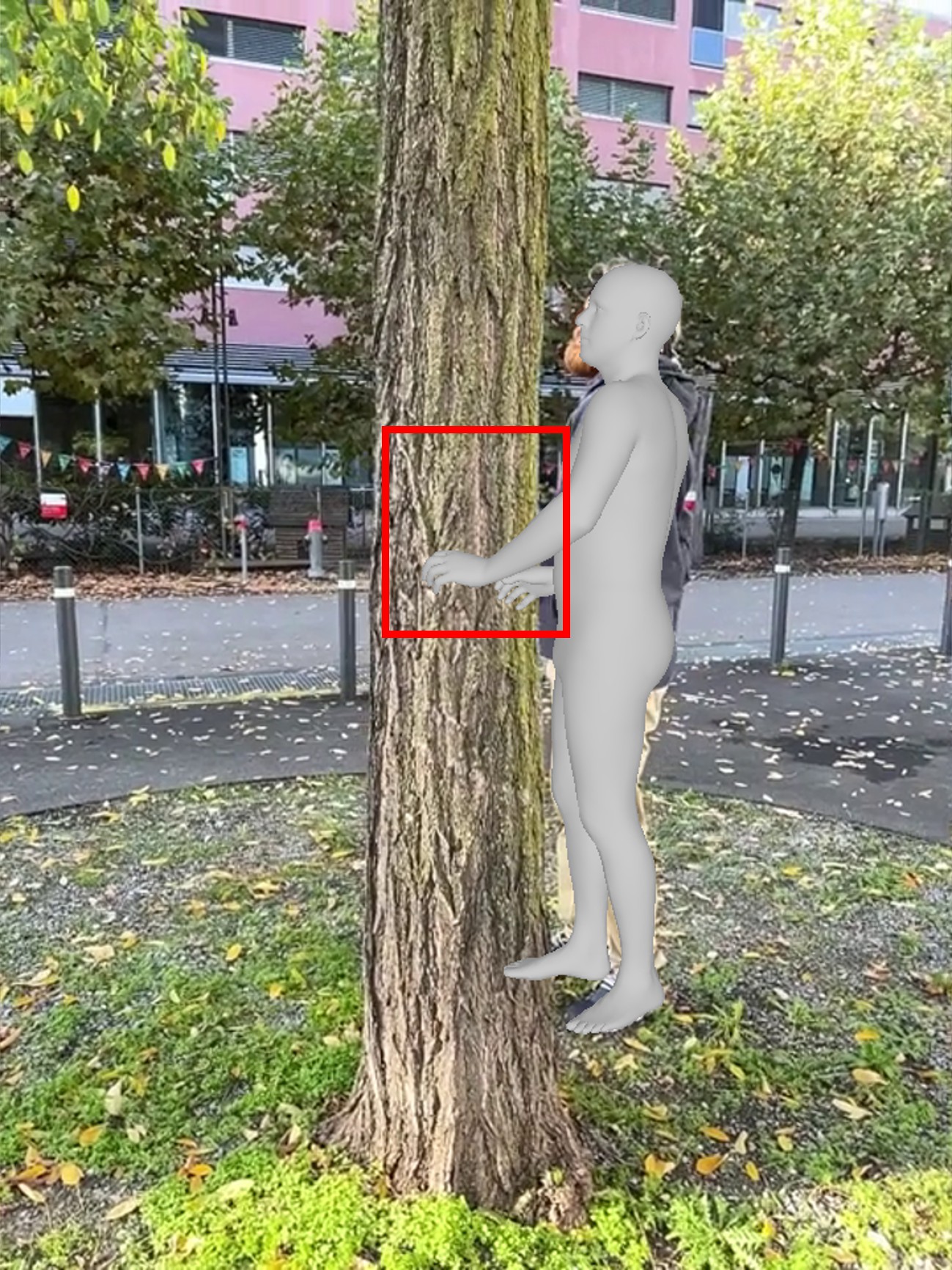}\\[-1pt]
        \end{minipage}\hfill
        \begin{minipage}[t]{0.32\linewidth}
            \parbox[t][8pt][c]{\linewidth}{\centering\scriptsize Ours}\par
            \vspace{2pt}
            \includegraphics[width=\linewidth]{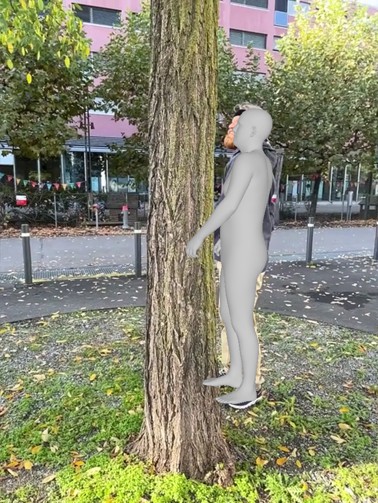}\\[-1pt]
        \end{minipage}
        \vfill
    \end{minipage}\hspace{\apptripgap}

    \begin{minipage}[t]{\apptripcolwidth}
        \centering
        \begin{minipage}[t]{0.32\linewidth}
            \parbox[t][8pt][c]{\linewidth}{\centering\scriptsize GT}\par
            \vspace{2pt}
            \includegraphics[width=\linewidth]{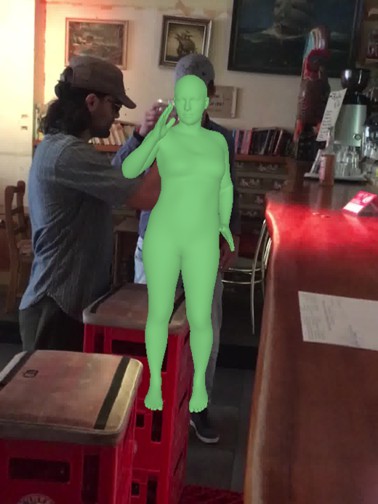}\\[-1pt]
        \end{minipage}\hfill
        \begin{minipage}[t]{0.32\linewidth}
            \parbox[t][8pt][c]{\linewidth}{\centering\scriptsize GVHMR}\par
            \vspace{2pt}
            \includegraphics[width=\linewidth]{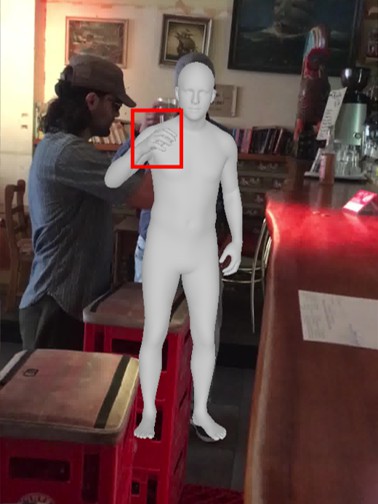}\\[-1pt]
        \end{minipage}\hfill
        \begin{minipage}[t]{0.32\linewidth}
            \parbox[t][8pt][c]{\linewidth}{\centering\scriptsize Ours}\par
            \vspace{2pt}
            \includegraphics[width=\linewidth]{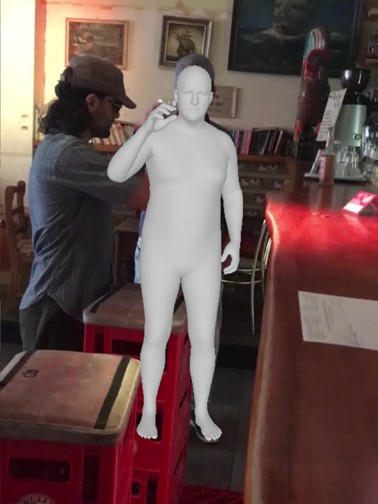}\\[-1pt]
        \end{minipage}
        \vfill
    \end{minipage}}

    \vspace{-0.4em}

    \makebox[\linewidth][c]{%
    \begin{minipage}[t]{\apptripcolwidth}
        \centering
        \begin{minipage}[t]{0.32\linewidth}
            \centering
            \includegraphics[width=\linewidth]{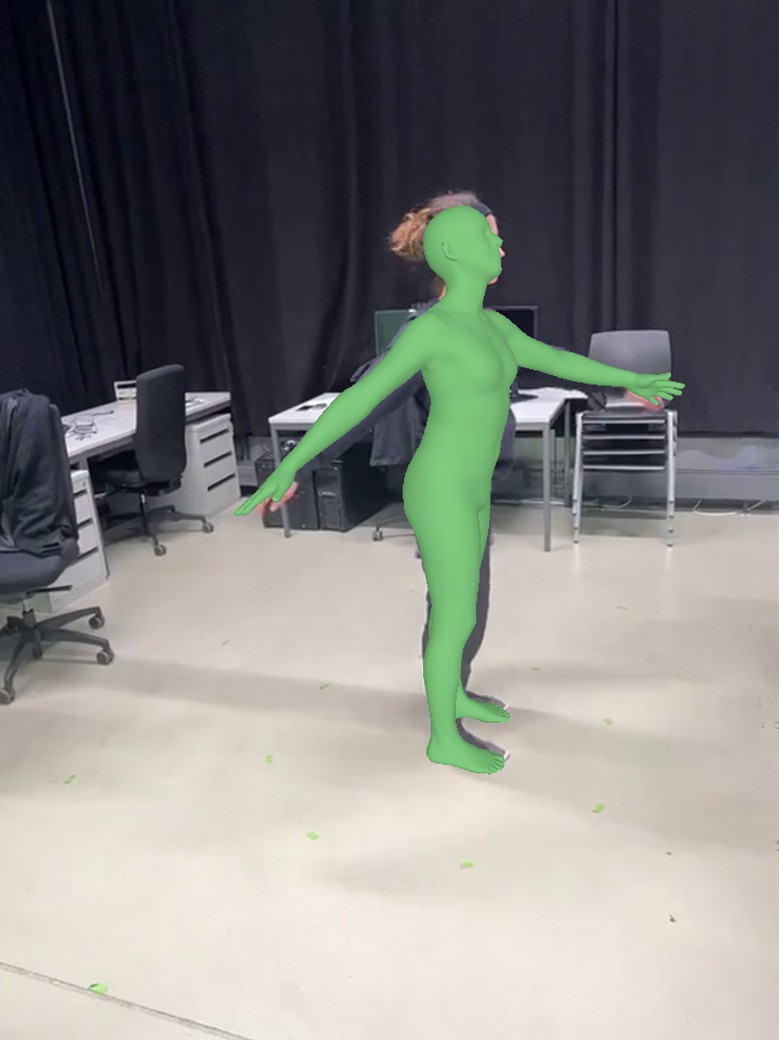}\\[-1pt]
        \end{minipage}\hfill
        \begin{minipage}[t]{0.32\linewidth}
            \centering
            \includegraphics[width=\linewidth]{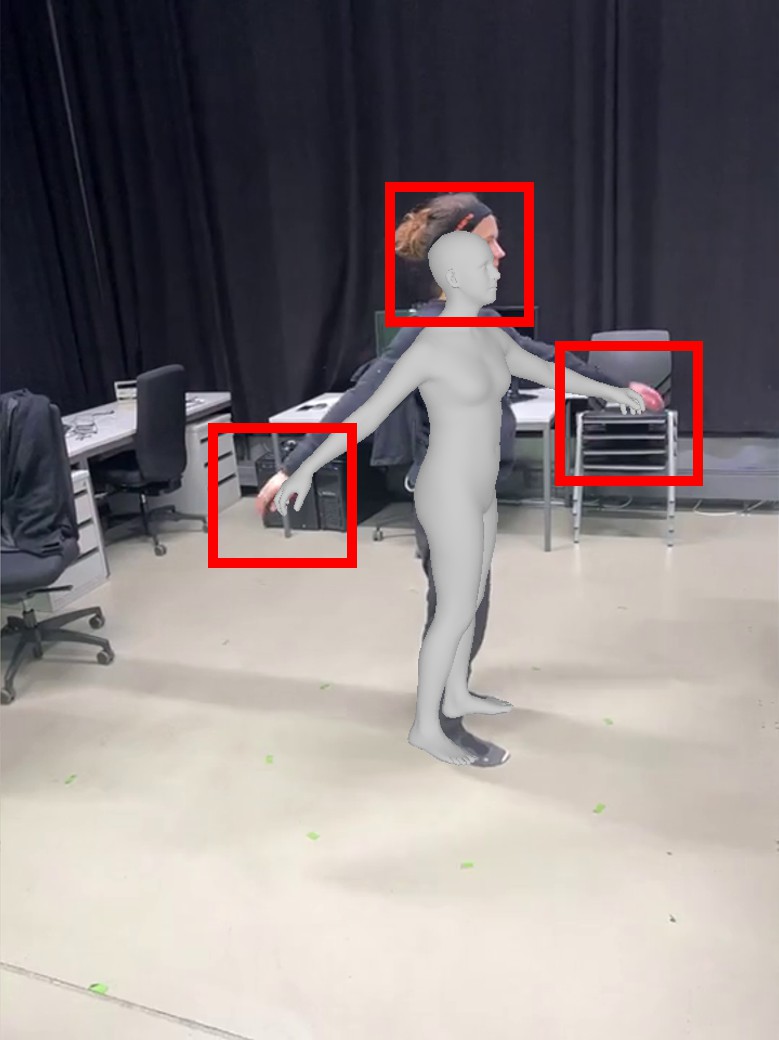}\\[-1pt]
        \end{minipage}\hfill
        \begin{minipage}[t]{0.32\linewidth}
            \centering
            \includegraphics[width=\linewidth]{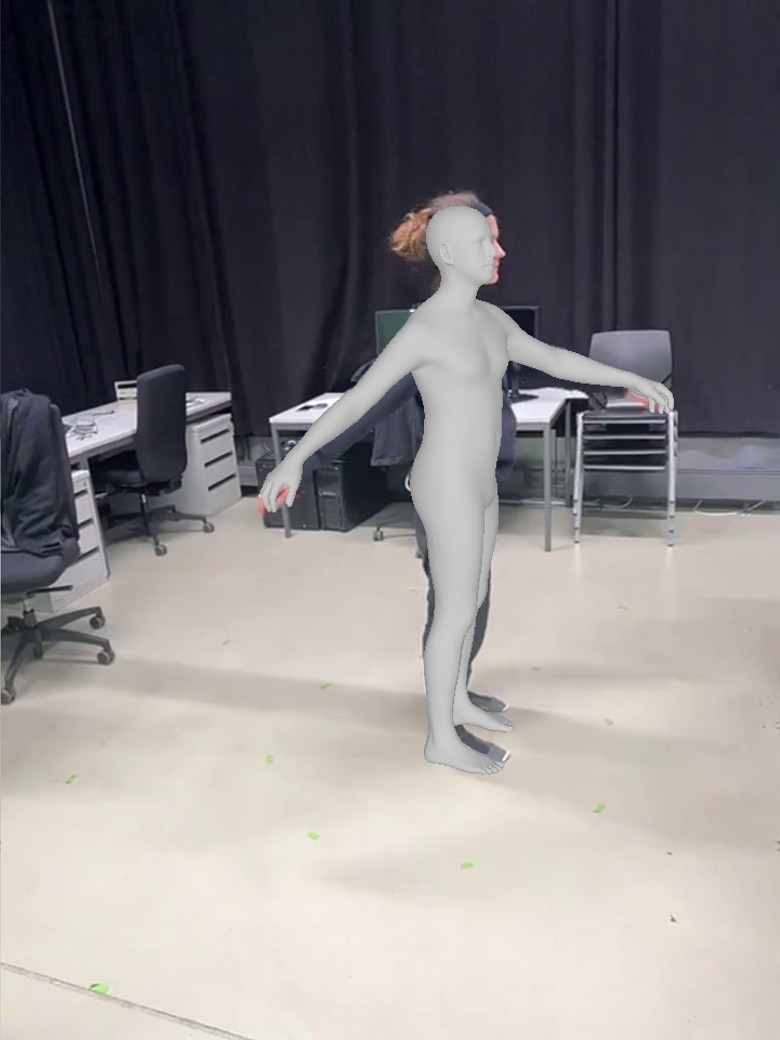}\\[-1pt]
        \end{minipage}
    \end{minipage}\hspace{\apptripgap}
    \begin{minipage}[t]{\apptripcolwidth}
        \centering
        \begin{minipage}[t]{0.32\linewidth}
            \centering
            \includegraphics[width=\linewidth]{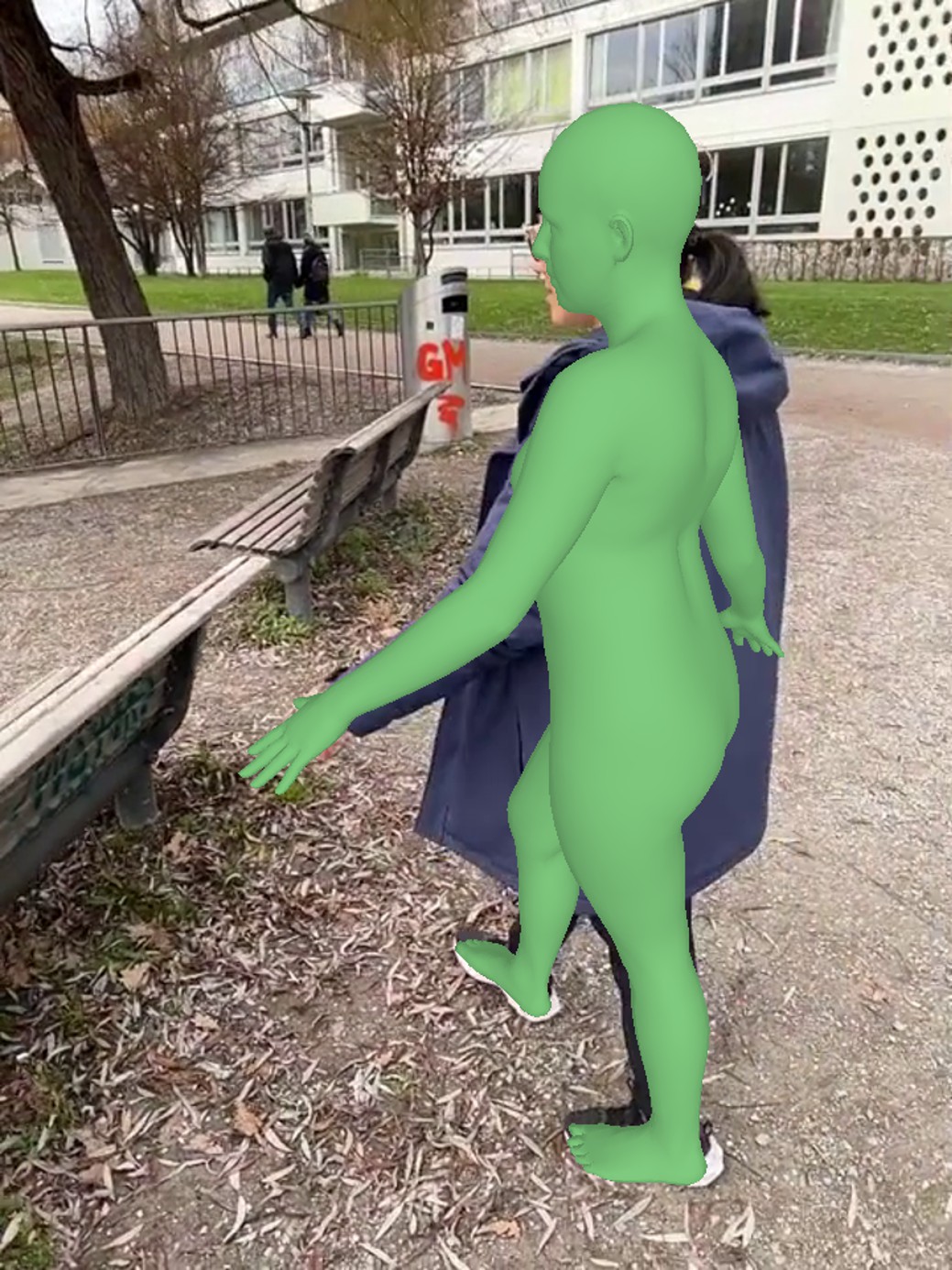}\\[-1pt]
        \end{minipage}\hfill
        \begin{minipage}[t]{0.32\linewidth}
            \centering
            \includegraphics[width=\linewidth]{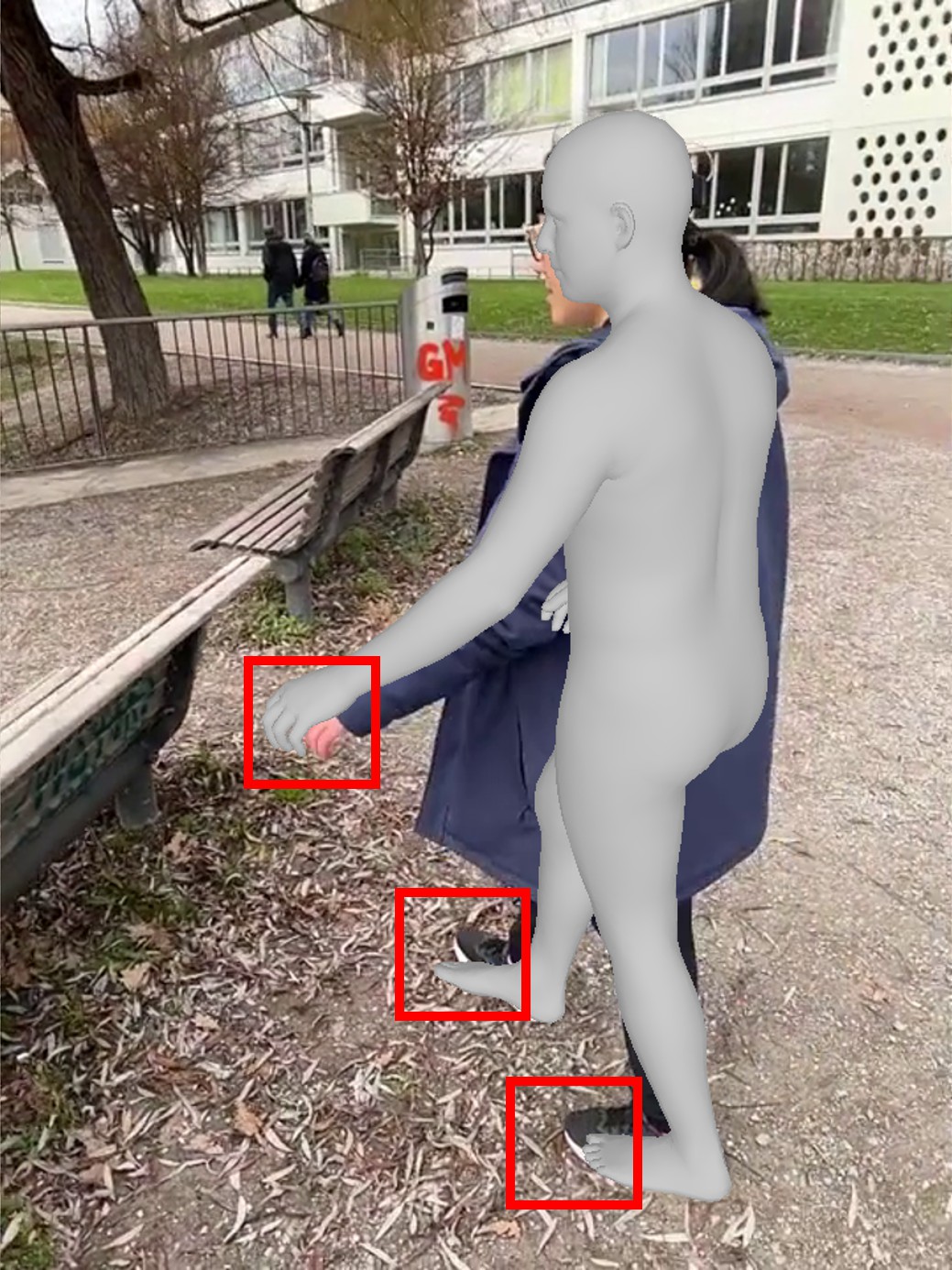}\\[-1pt]
        \end{minipage}\hfill
        \begin{minipage}[t]{0.32\linewidth}
            \centering
            \includegraphics[width=\linewidth]{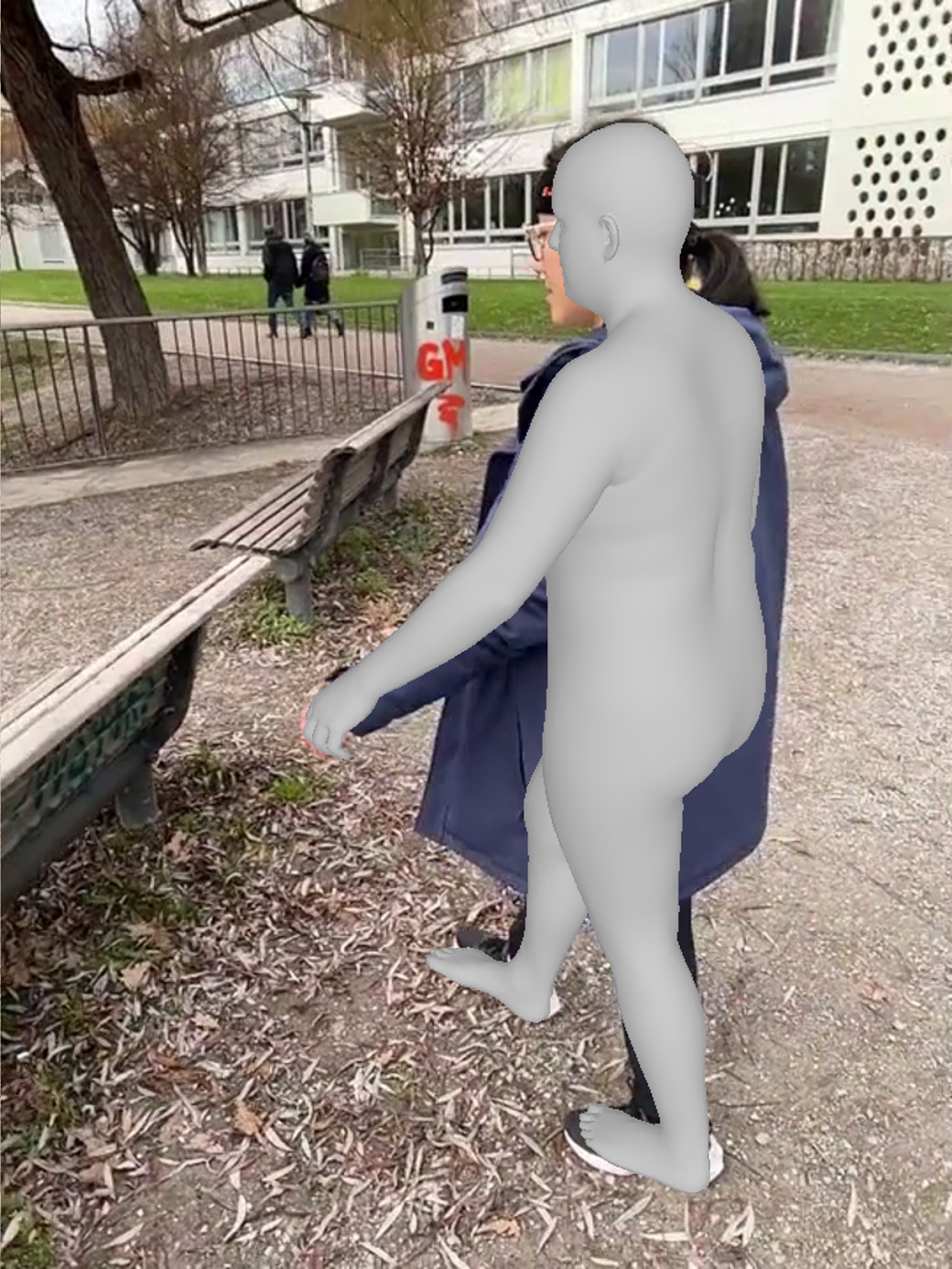}\\[-1pt]
        \end{minipage}
    \end{minipage}}

    \vspace{0.4em}

    \makebox[\linewidth][c]{%
    \begin{minipage}[t]{\apptripcolwidth}
        \centering
        \begin{minipage}[t]{0.32\linewidth}
            \centering
            \includegraphics[width=\linewidth]{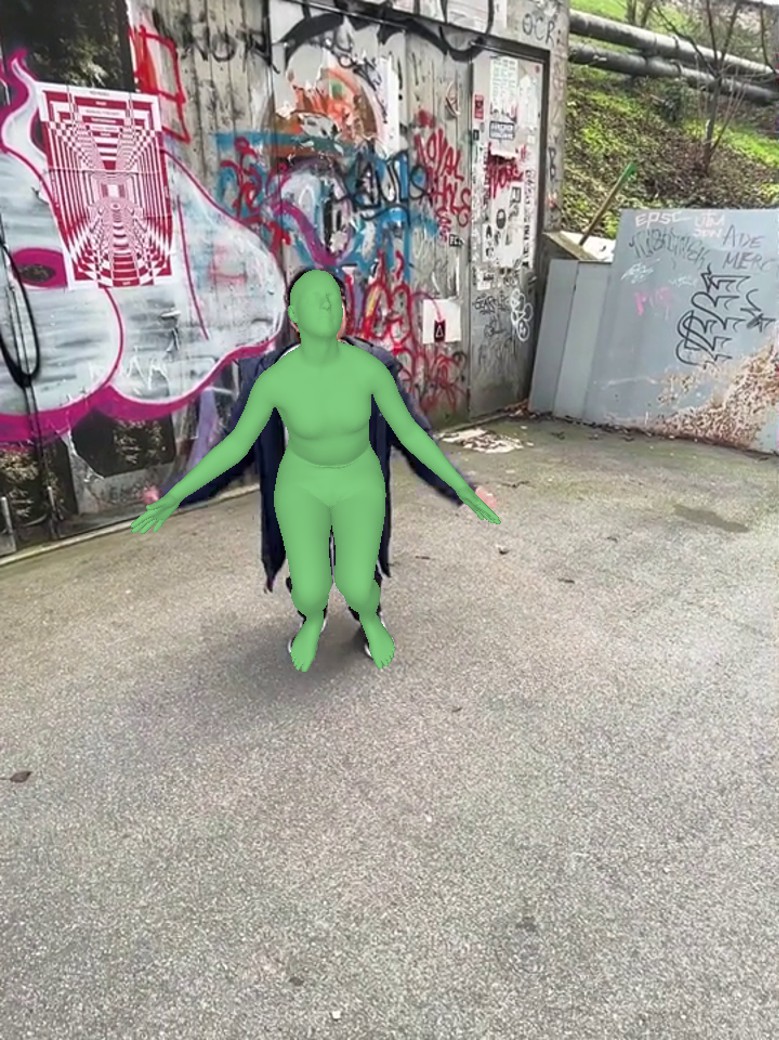}\\[-1pt]
        \end{minipage}\hfill
        \begin{minipage}[t]{0.32\linewidth}
            \centering
            \includegraphics[width=\linewidth]{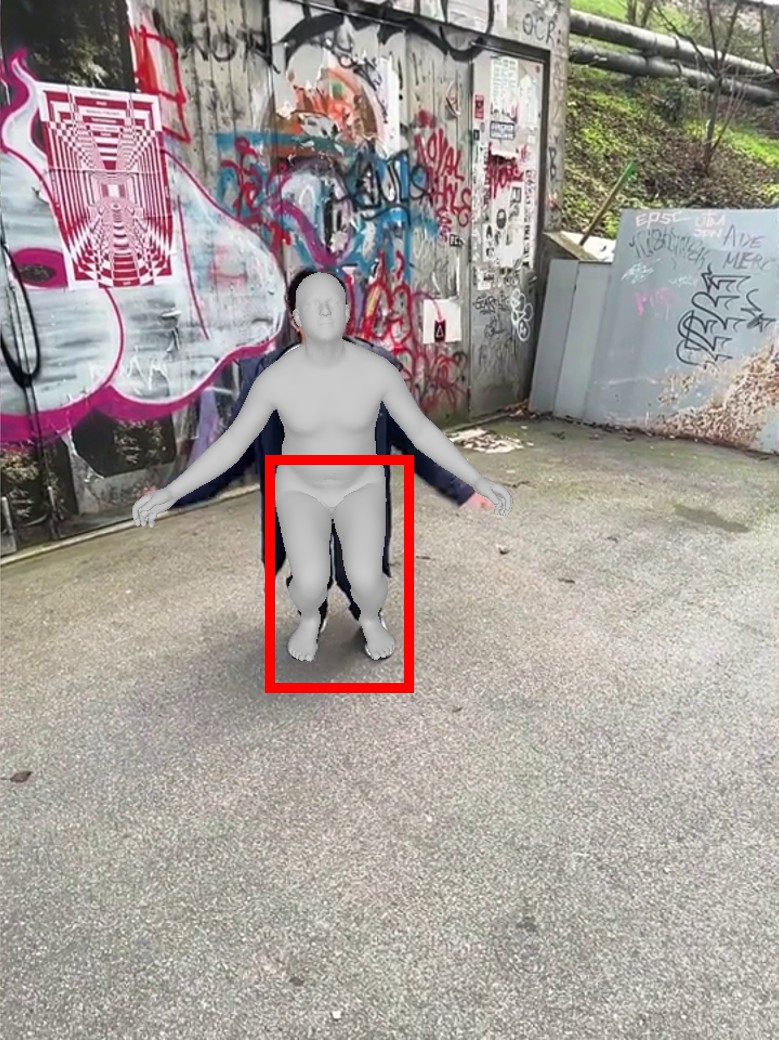}\\[-1pt]
        \end{minipage}\hfill
        \begin{minipage}[t]{0.32\linewidth}
            \centering
            \includegraphics[width=\linewidth]{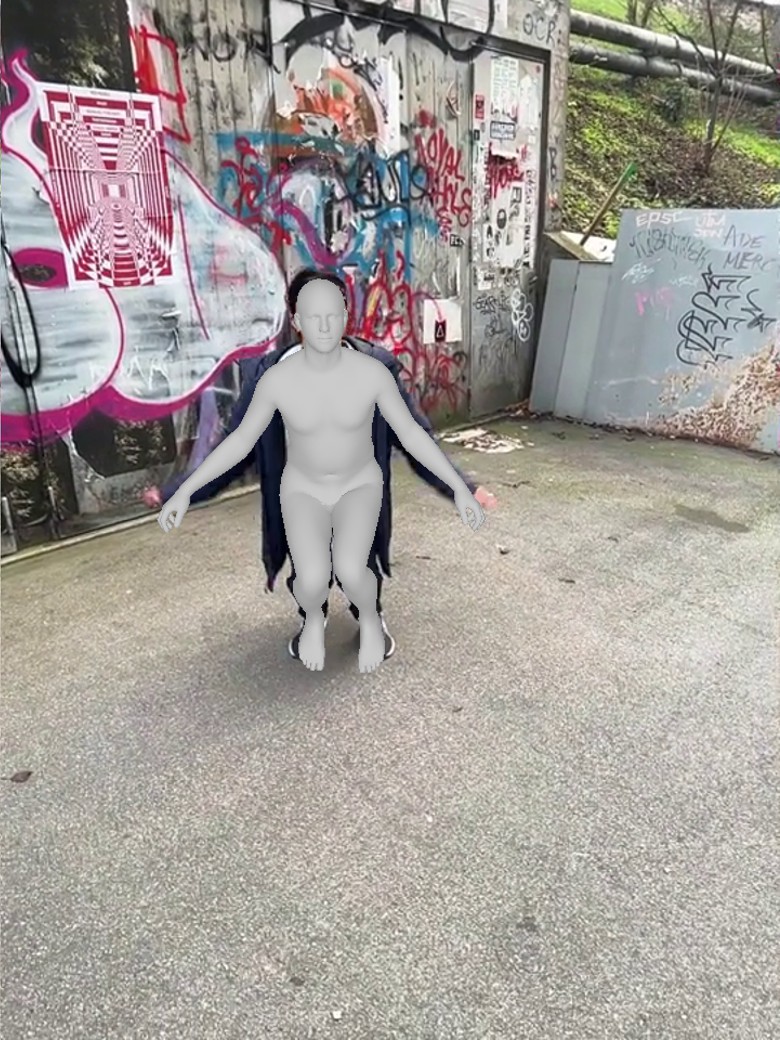}\\[-1pt]
        \end{minipage}
    \end{minipage}\hspace{\apptripgap}
    \begin{minipage}[t]{\apptripcolwidth}
        \centering
        \begin{minipage}[t]{0.32\linewidth}
            \centering
            \includegraphics[width=\linewidth]{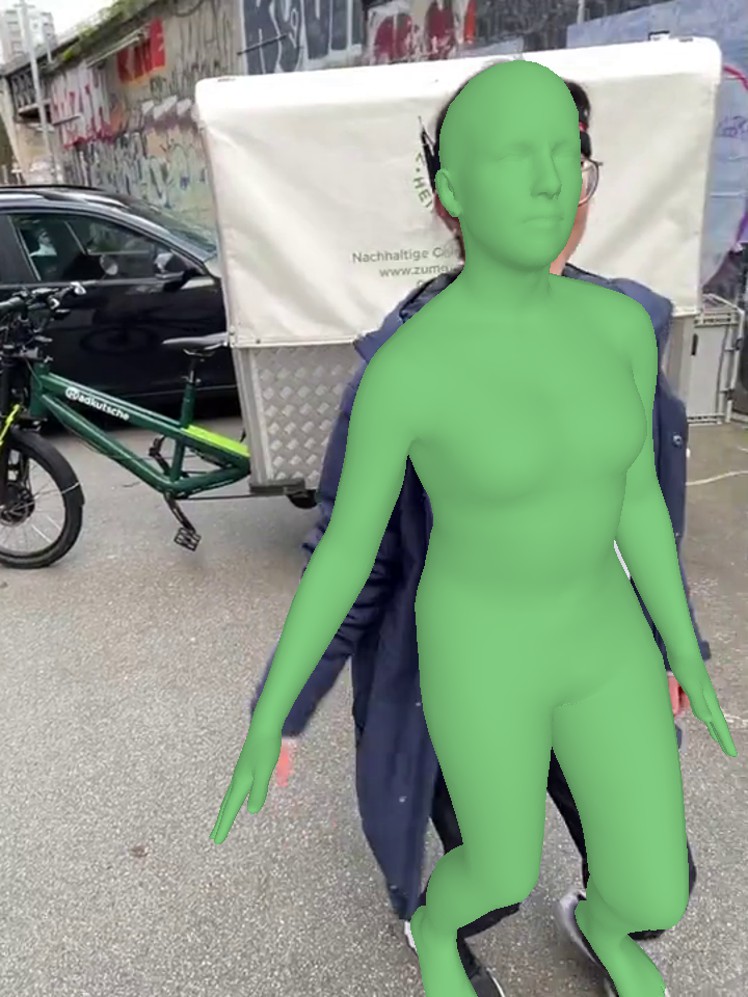}\\[-1pt]
        \end{minipage}\hfill
        \begin{minipage}[t]{0.32\linewidth}
            \centering
            \includegraphics[width=\linewidth]{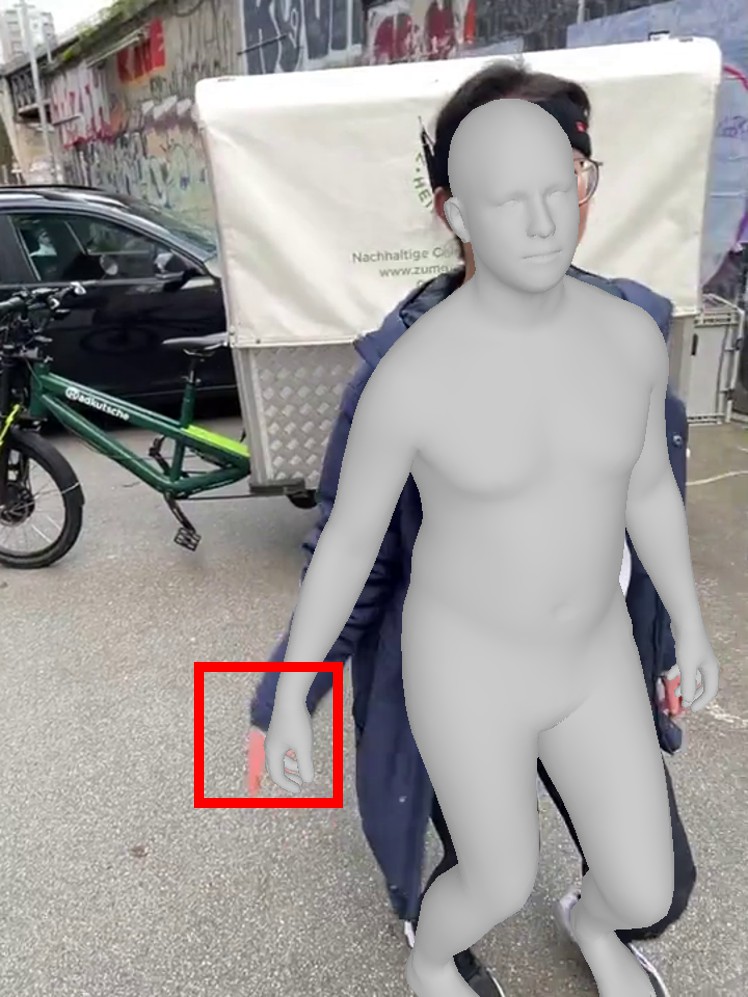}\\[-1pt]
        \end{minipage}\hfill
        \begin{minipage}[t]{0.32\linewidth}
            \centering
            \includegraphics[width=\linewidth]{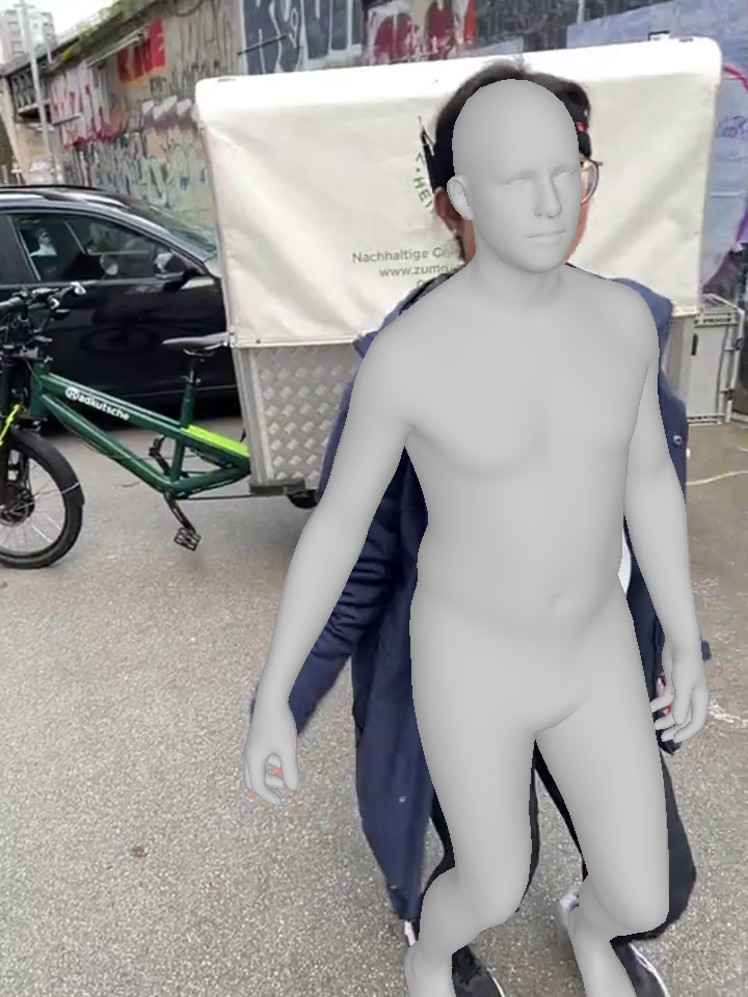}\\[-1pt]
        \end{minipage}
    \end{minipage}}

    \vspace{0.4em}

    \makebox[\linewidth][c]{%
    \begin{minipage}[t]{\apptripcolwidth}
        \centering
        \begin{minipage}[t]{0.32\linewidth}
            \centering
            \includegraphics[width=\linewidth]{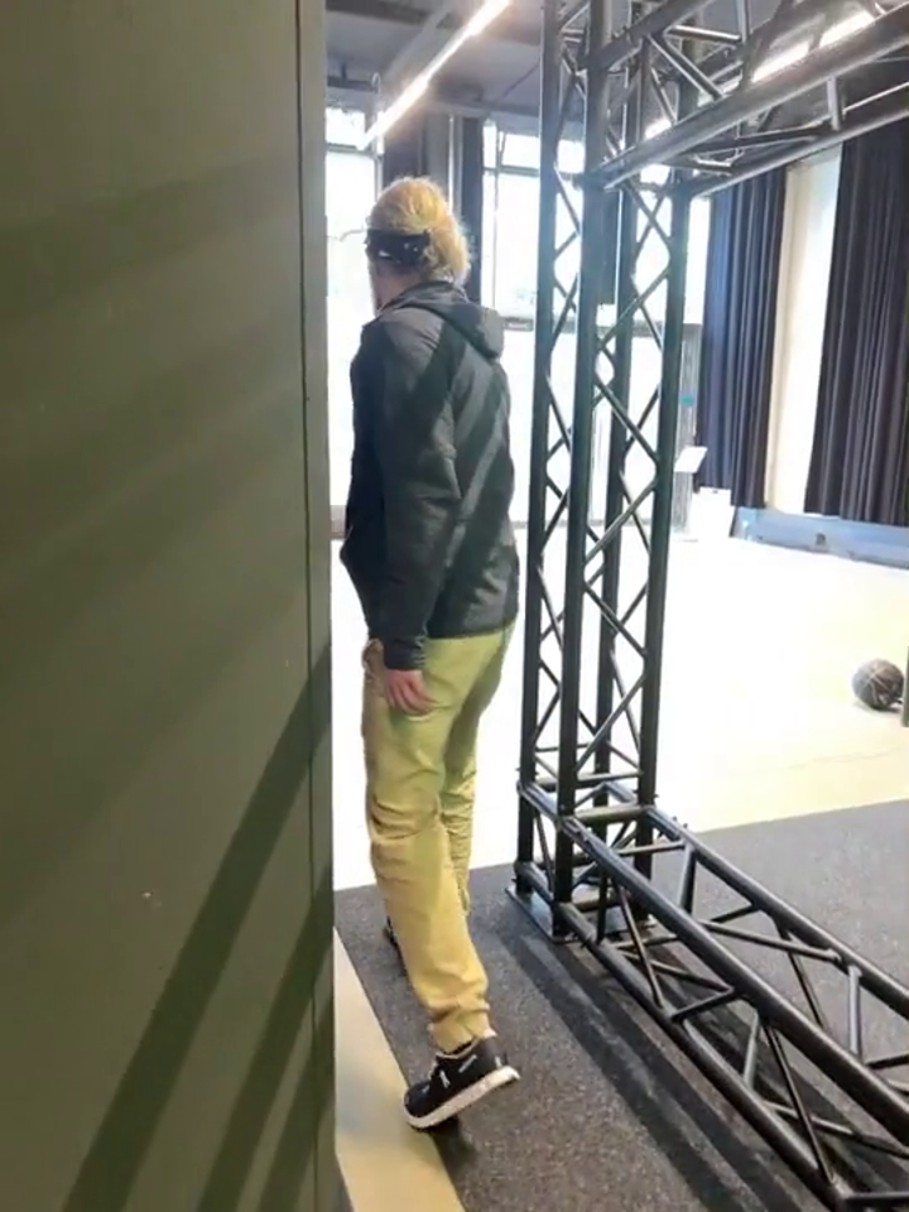}\\[-1pt]
        \end{minipage}\hfill
        \begin{minipage}[t]{0.32\linewidth}
            \centering
            \includegraphics[width=\linewidth]{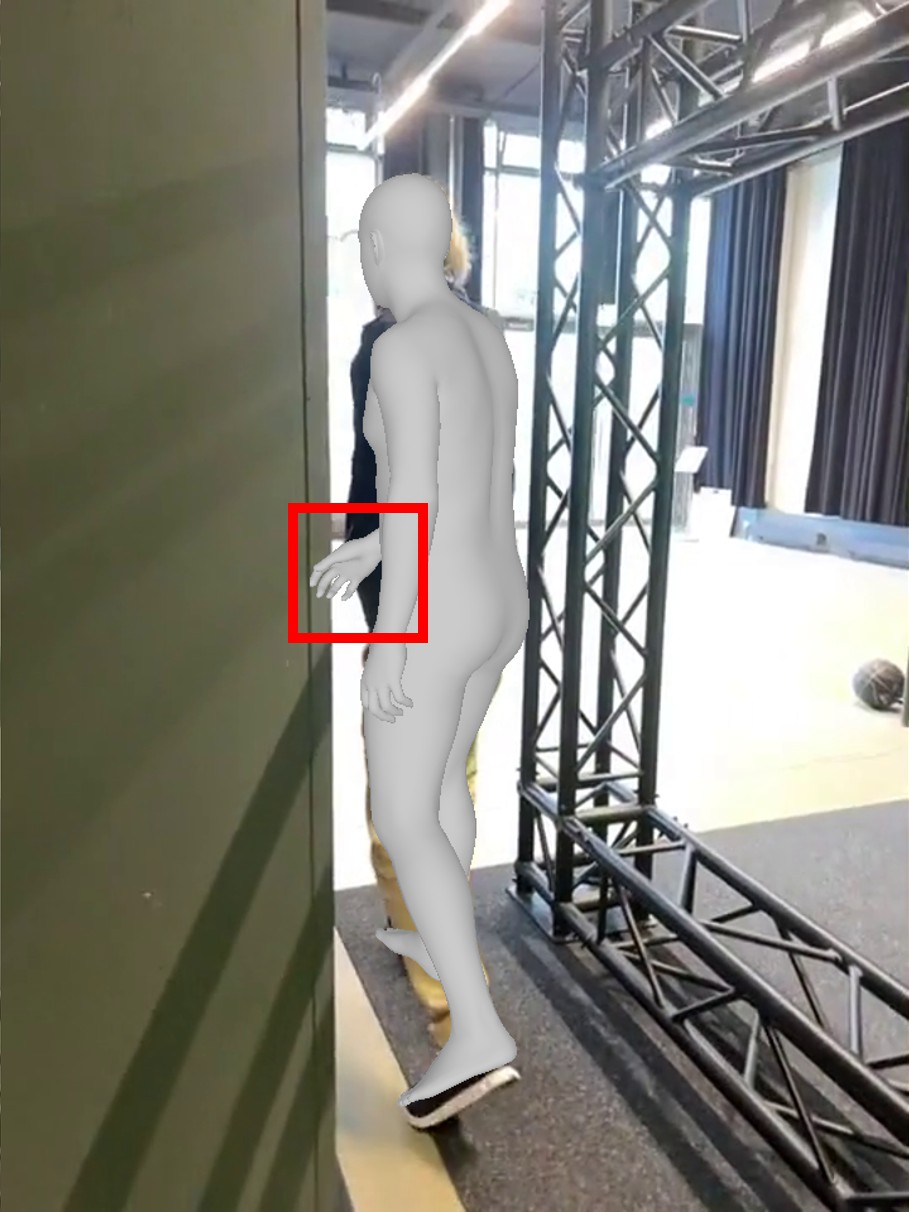}\\[-1pt]
        \end{minipage}\hfill
        \begin{minipage}[t]{0.32\linewidth}
            \centering
            \includegraphics[width=\linewidth]{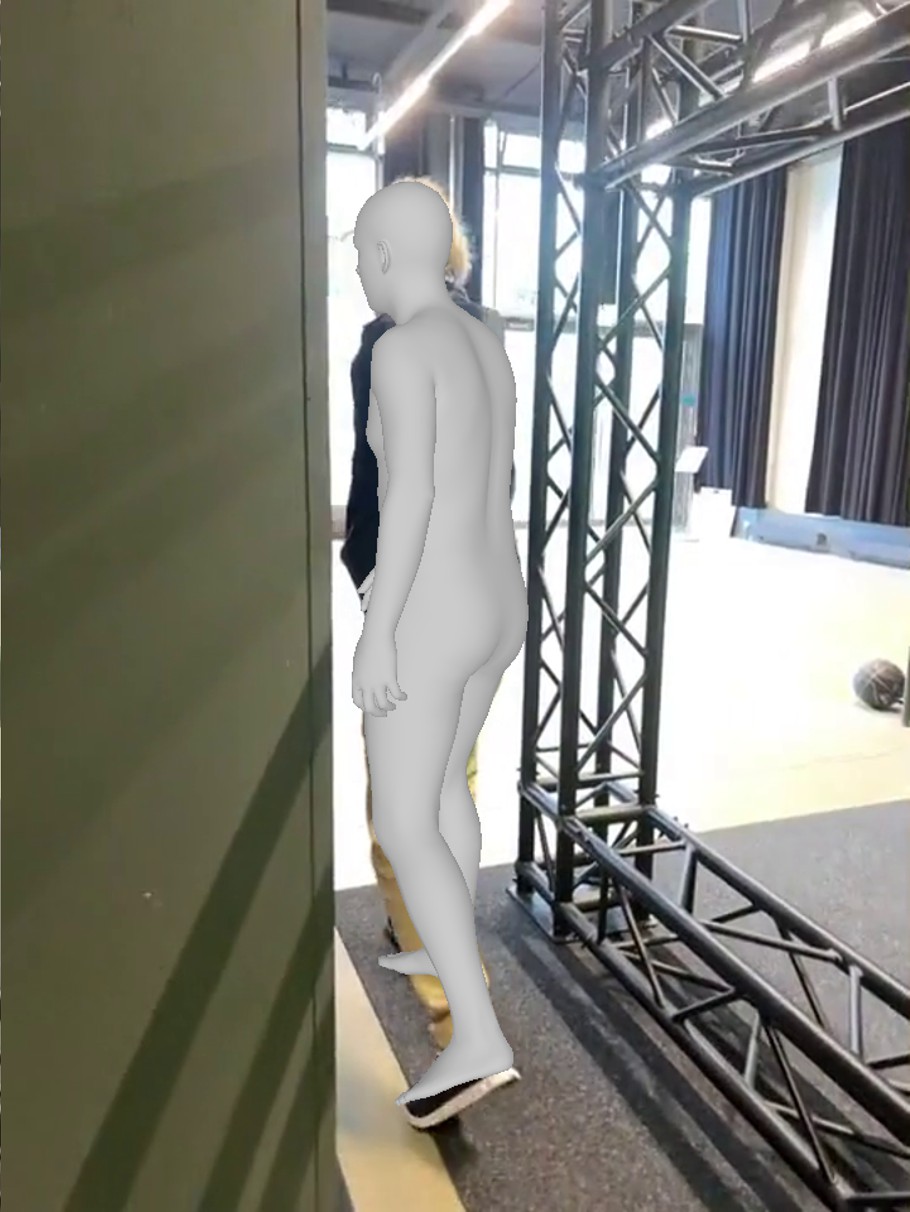}\\[-1pt]
        \end{minipage}
    \end{minipage}\hspace{\apptripgap}
    \begin{minipage}[t]{\apptripcolwidth}
        \centering
        \begin{minipage}[t]{0.32\linewidth}
            \centering
            \includegraphics[width=\linewidth]{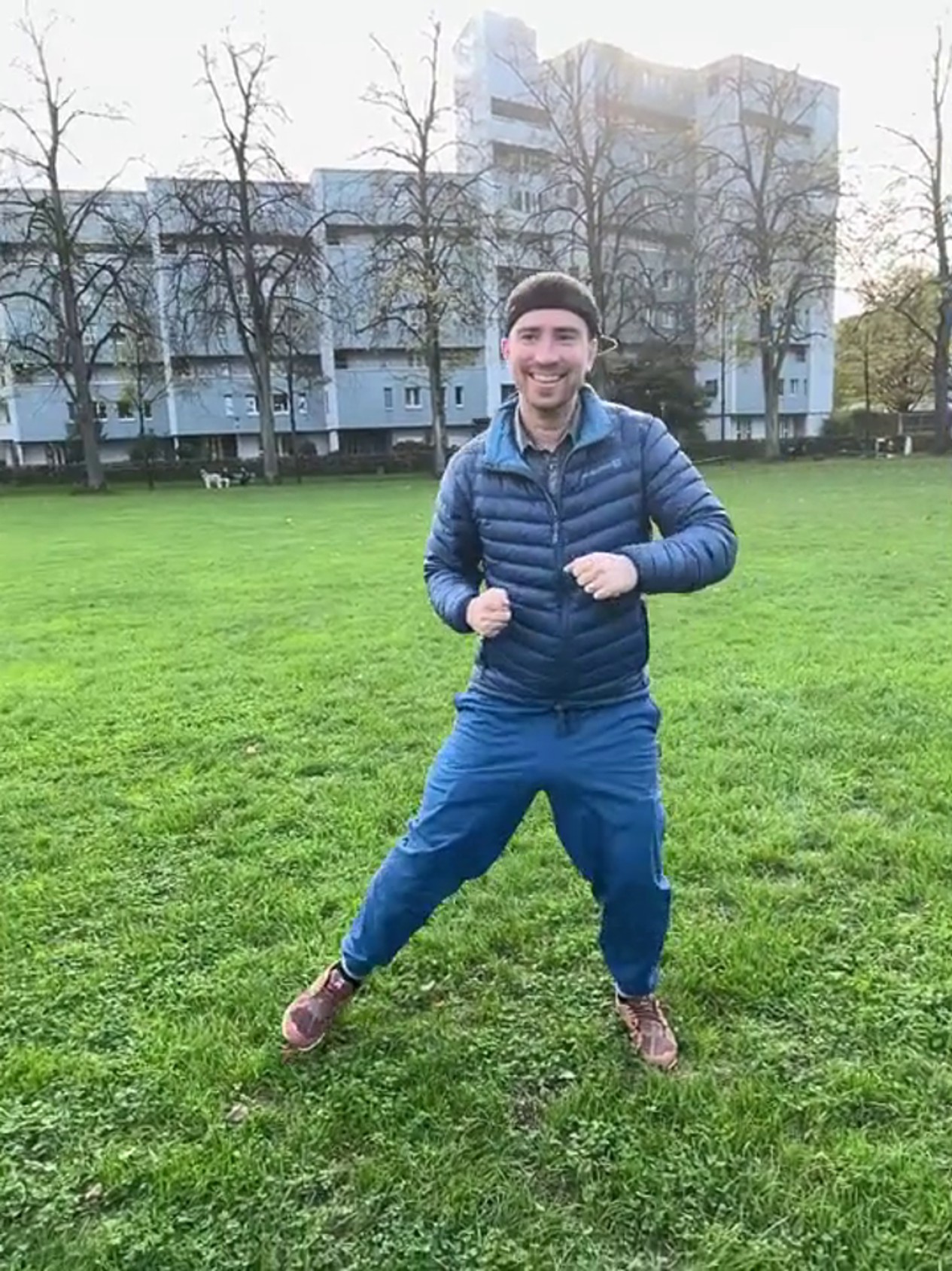}\\[-1pt]
        \end{minipage}\hfill
        \begin{minipage}[t]{0.32\linewidth}
            \centering
            \includegraphics[width=\linewidth]{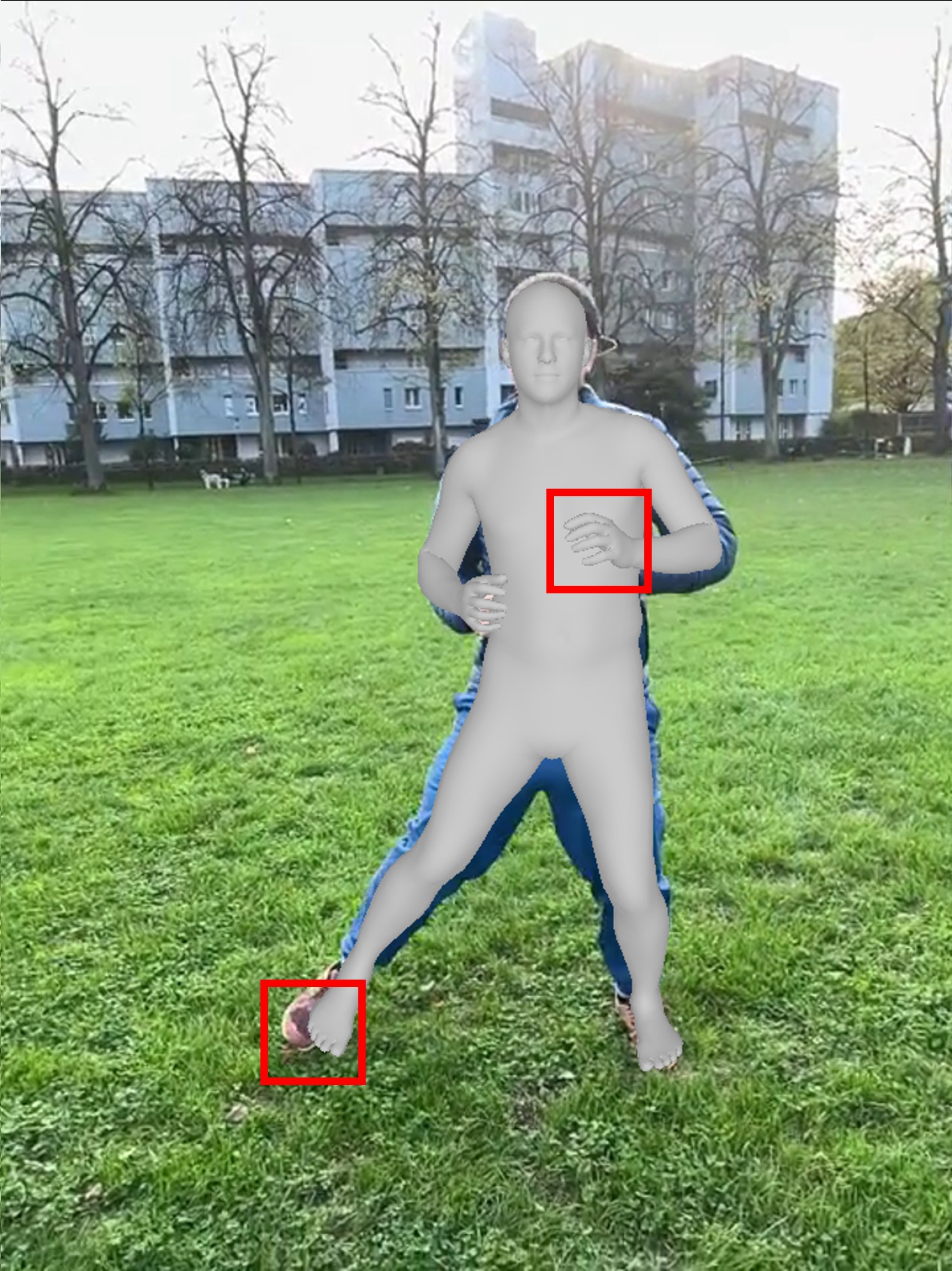}\\[-1pt]
        \end{minipage}\hfill
        \begin{minipage}[t]{0.32\linewidth}
            \centering
            \includegraphics[width=\linewidth]{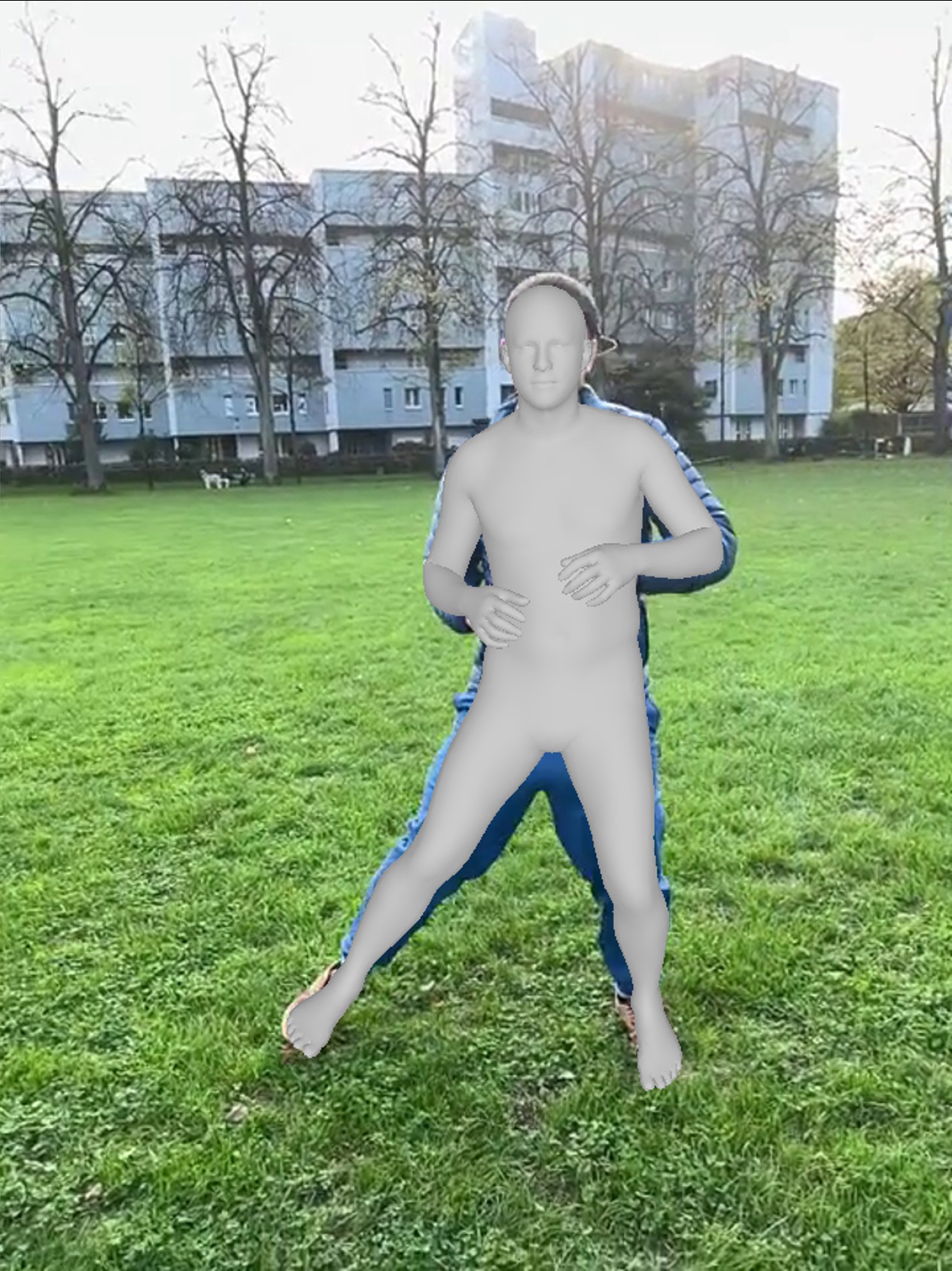}\\[-1pt]
        \end{minipage}
    \end{minipage}}

    \caption{Additional qualitative comparisons in the appendix. Each example is shown left-to-right as GT, baseline, and Ours. Across a broader set of ambiguity-heavy scenes, FactorizedHMR produces distal-limb completions that remain more consistent with the visible evidence under occlusion, truncation, and clutter.}
    \label{fig:appendix_qualitative_grid}
\end{figure}
\clearpage

\begin{figure}[p]
    \centering
    \newcommand{\roww}{1.2\linewidth}
    \def\appfigsevencolwidth{0.31\linewidth}
    \def\appfigmargin{0.01\linewidth}

    \makebox[\linewidth][c]{%
        \begin{minipage}[t]{\roww}
            \centering
            \begin{minipage}[t]{\appfigsevencolwidth}
                \parbox[t][8pt][c]{\linewidth}{\centering\scriptsize Input}\par
                \vspace{2pt}
                \includegraphics[width=\linewidth]{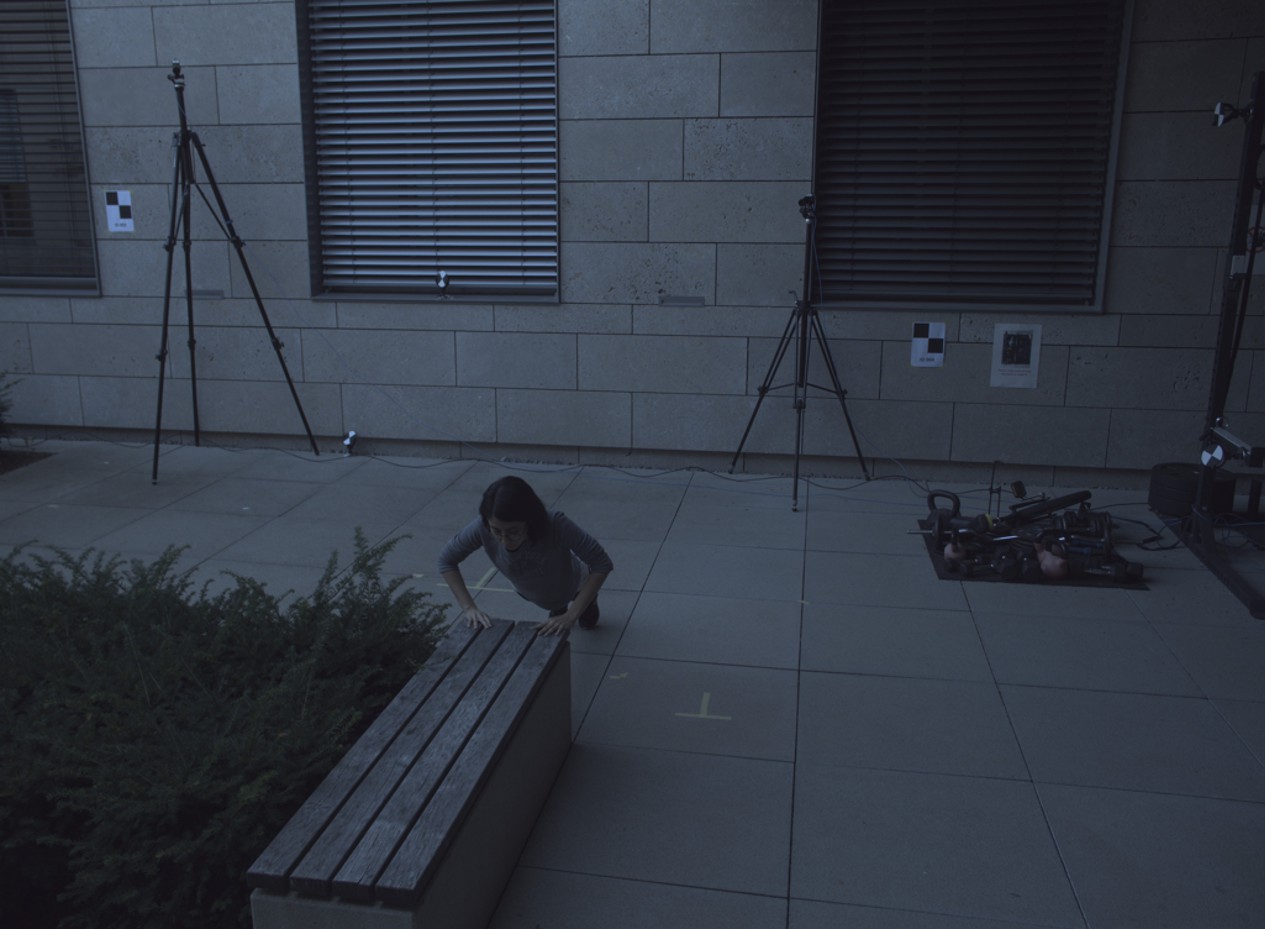}
            \end{minipage}\hspace{\appfigmargin}
            \begin{minipage}[t]{\appfigsevencolwidth}
                \parbox[t][8pt][c]{\linewidth}{\centering\scriptsize GVHMR}\par
                \vspace{2pt}
                \includegraphics[width=\linewidth]{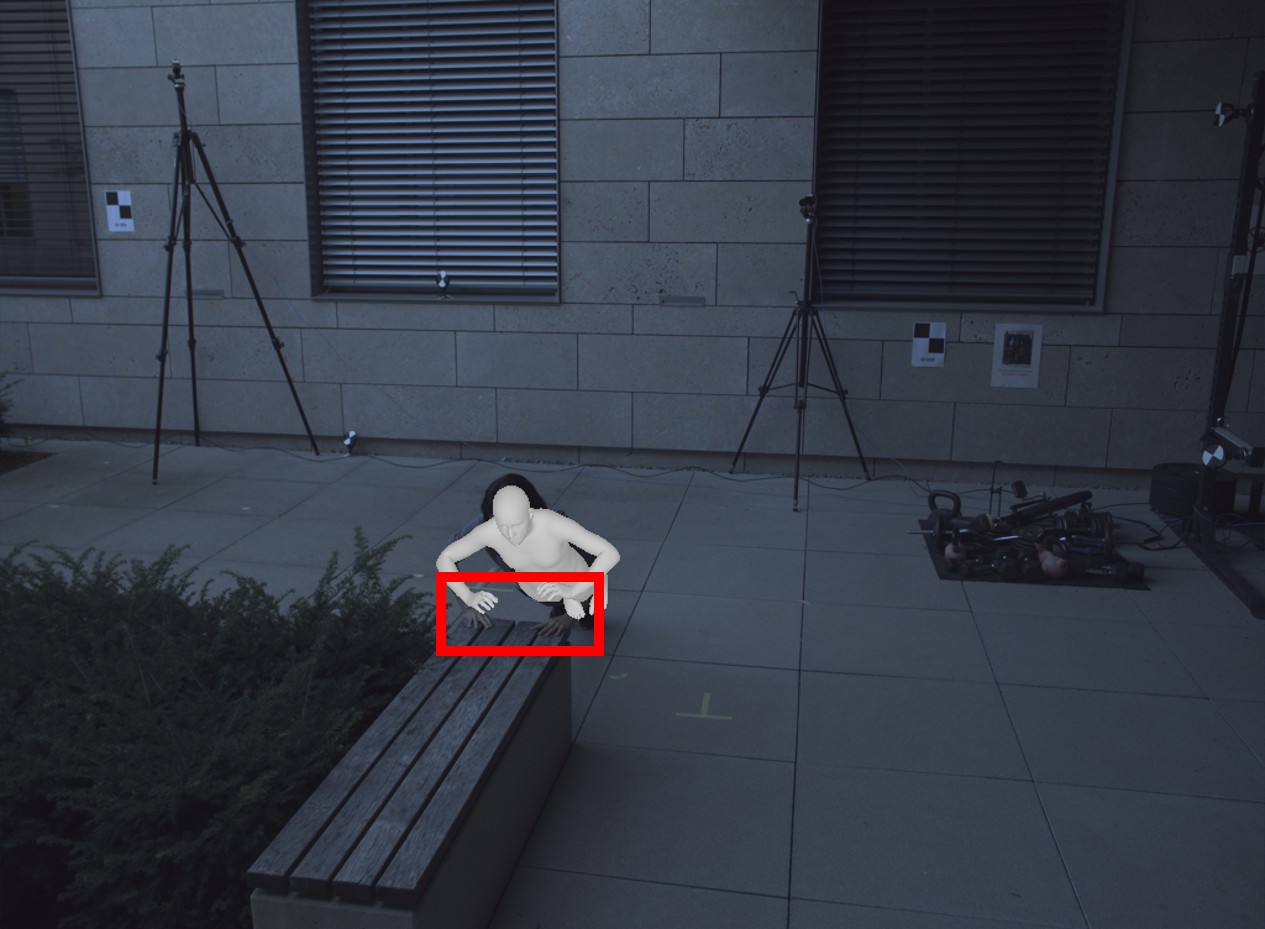}\\[-1pt]
            \end{minipage}\hspace{\appfigmargin}
            \begin{minipage}[t]{\appfigsevencolwidth}
                \parbox[t][8pt][c]{\linewidth}{\centering\scriptsize Ours}\par
                \vspace{2pt}
                \includegraphics[width=\linewidth]{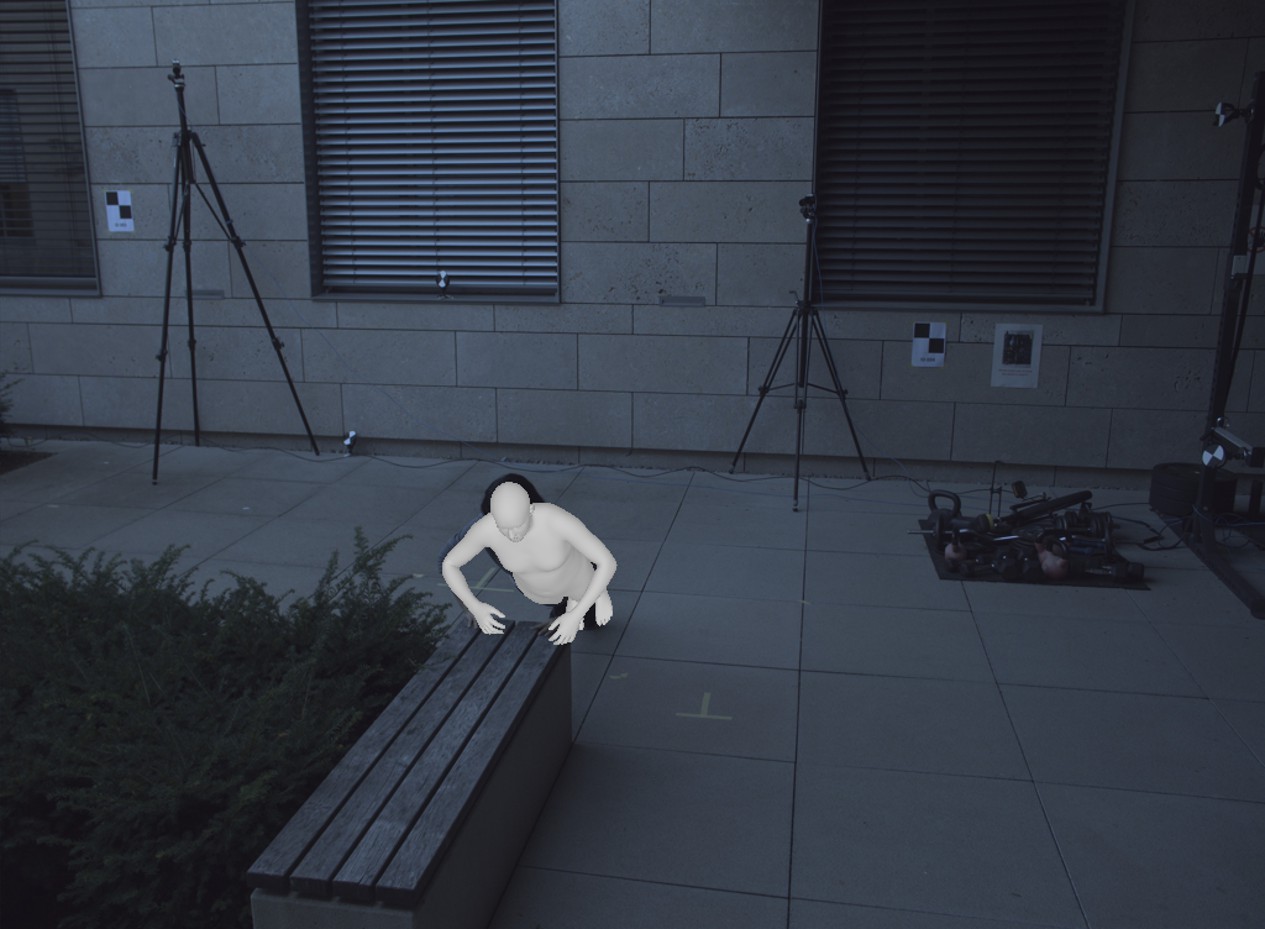}\\[-1pt]
            \end{minipage}
        \end{minipage}%
    }
    
    \vspace{-0.3em}

    \makebox[\linewidth][c]{%
        \begin{minipage}[t]{\roww}
            \centering
            \begin{minipage}[t]{\appfigsevencolwidth}
                \centering
                \includegraphics[width=\linewidth]{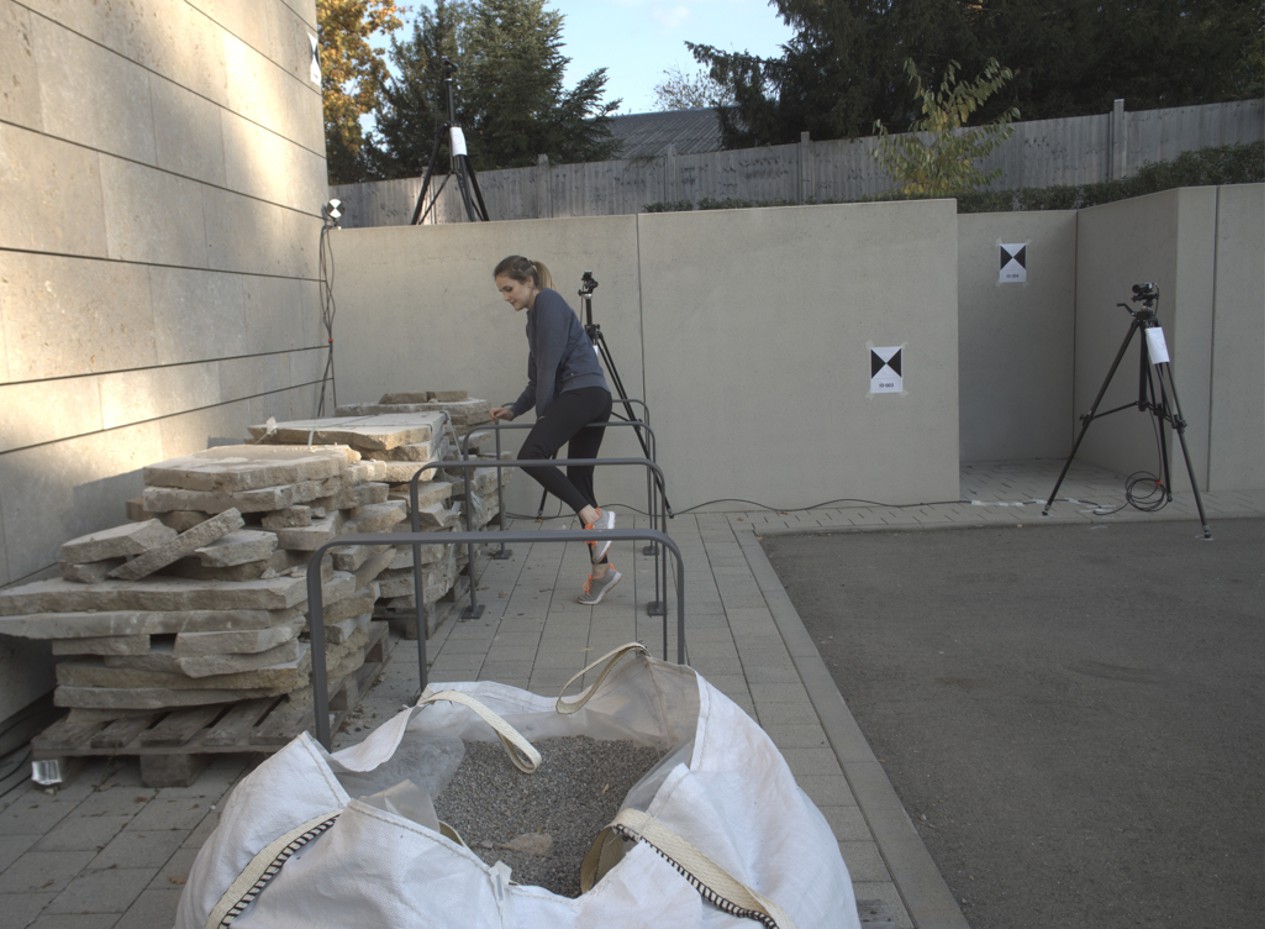}\\[-1pt]
            \end{minipage}\hspace{\appfigmargin}
            \begin{minipage}[t]{\appfigsevencolwidth}
                \centering
                \includegraphics[width=\linewidth]{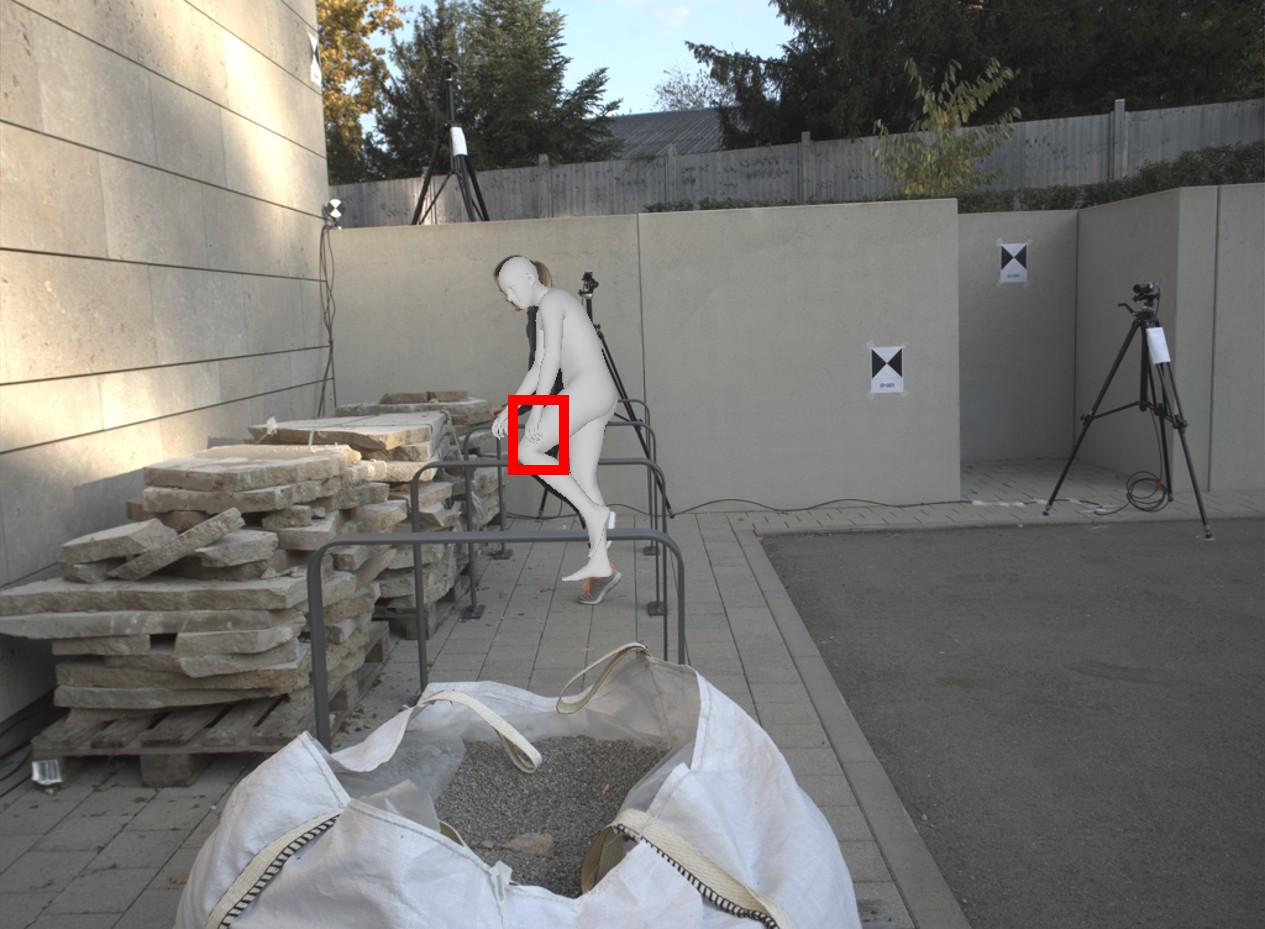}\\[-1pt]
            \end{minipage}\hspace{\appfigmargin}
            \begin{minipage}[t]{\appfigsevencolwidth}
                \centering
                \includegraphics[width=\linewidth]{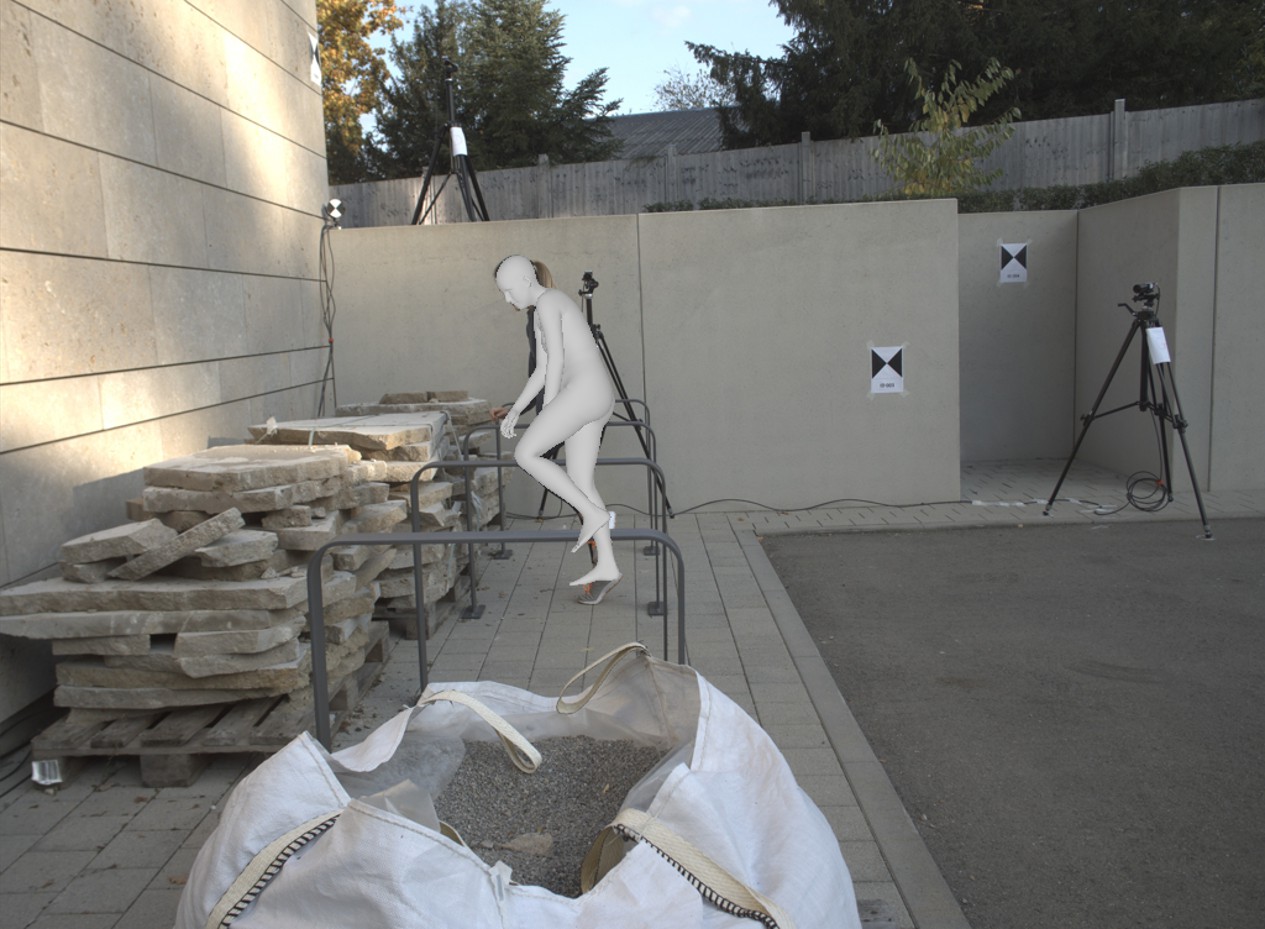}\\[-1pt]
            \end{minipage}
        \end{minipage}%
    }

    \vspace{0.5em}

    \makebox[\linewidth][c]{%
        \begin{minipage}[t]{\roww}
            \centering
            \begin{minipage}[t]{\appfigsevencolwidth}
                \centering
                \includegraphics[width=\linewidth]{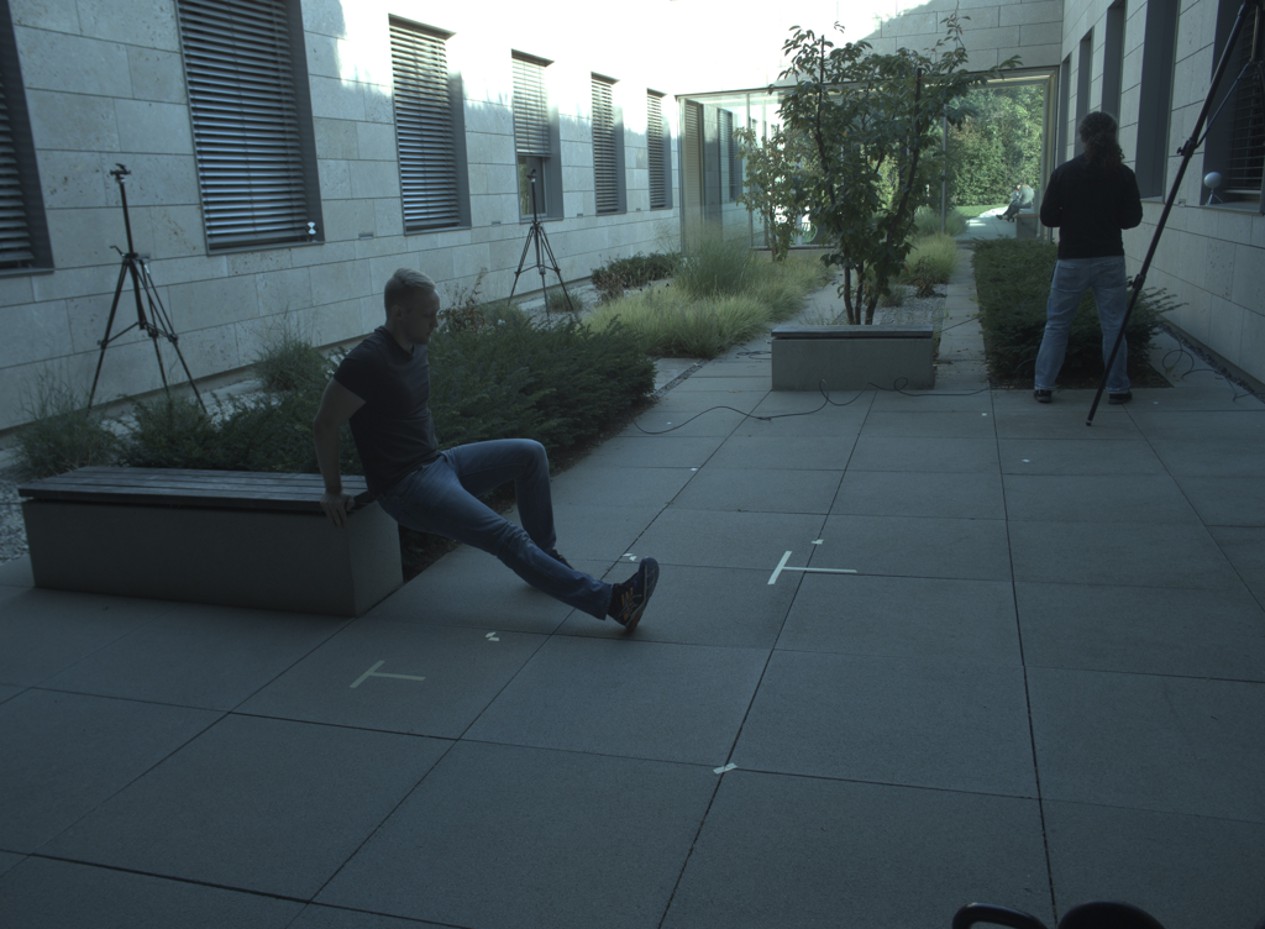}\\[-1pt]
            \end{minipage}\hspace{\appfigmargin}
            \begin{minipage}[t]{\appfigsevencolwidth}
                \centering
                \includegraphics[width=\linewidth]{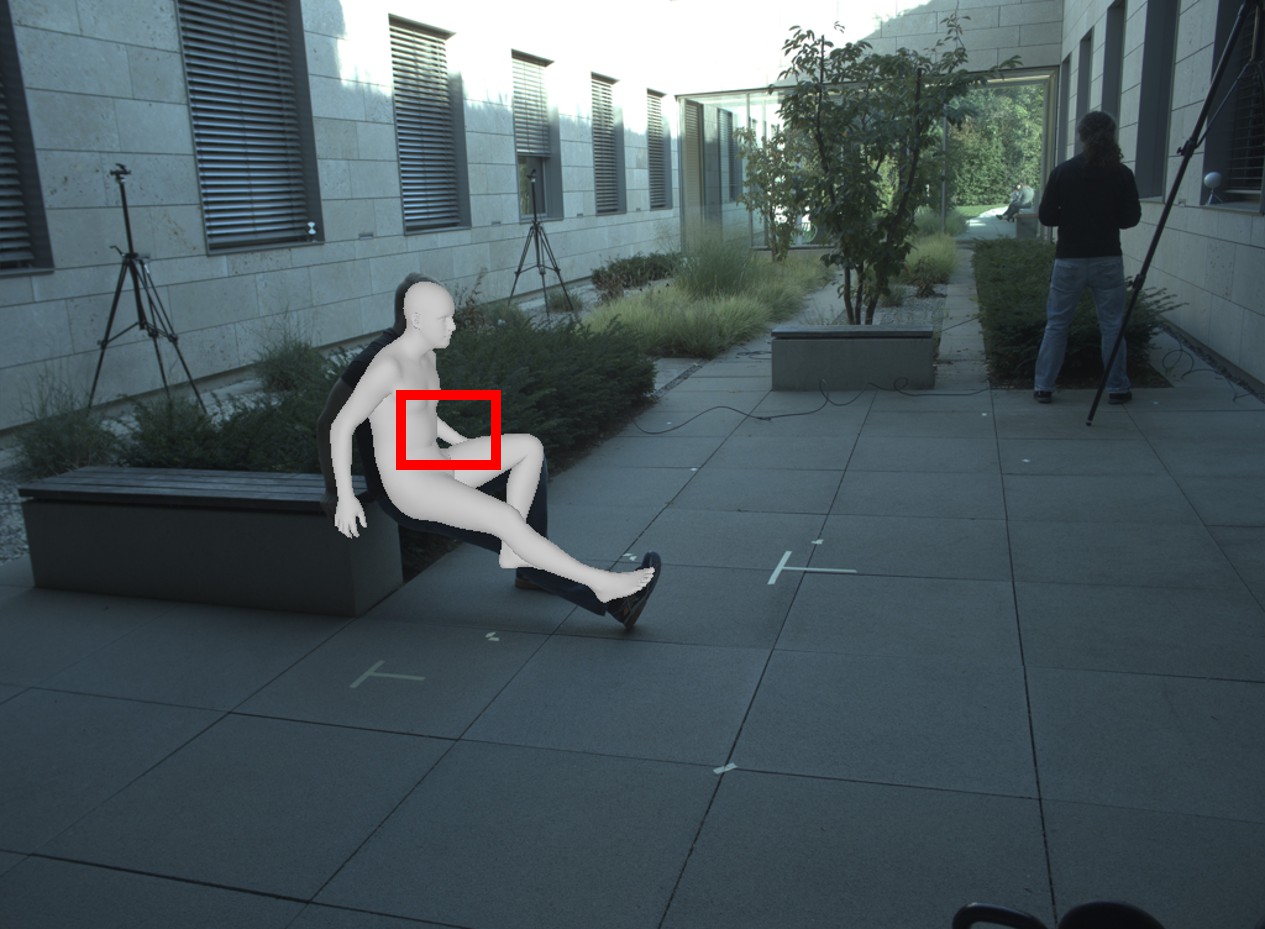}\\[-1pt]
            \end{minipage}\hspace{\appfigmargin}
            \begin{minipage}[t]{\appfigsevencolwidth}
                \centering
                \includegraphics[width=\linewidth]{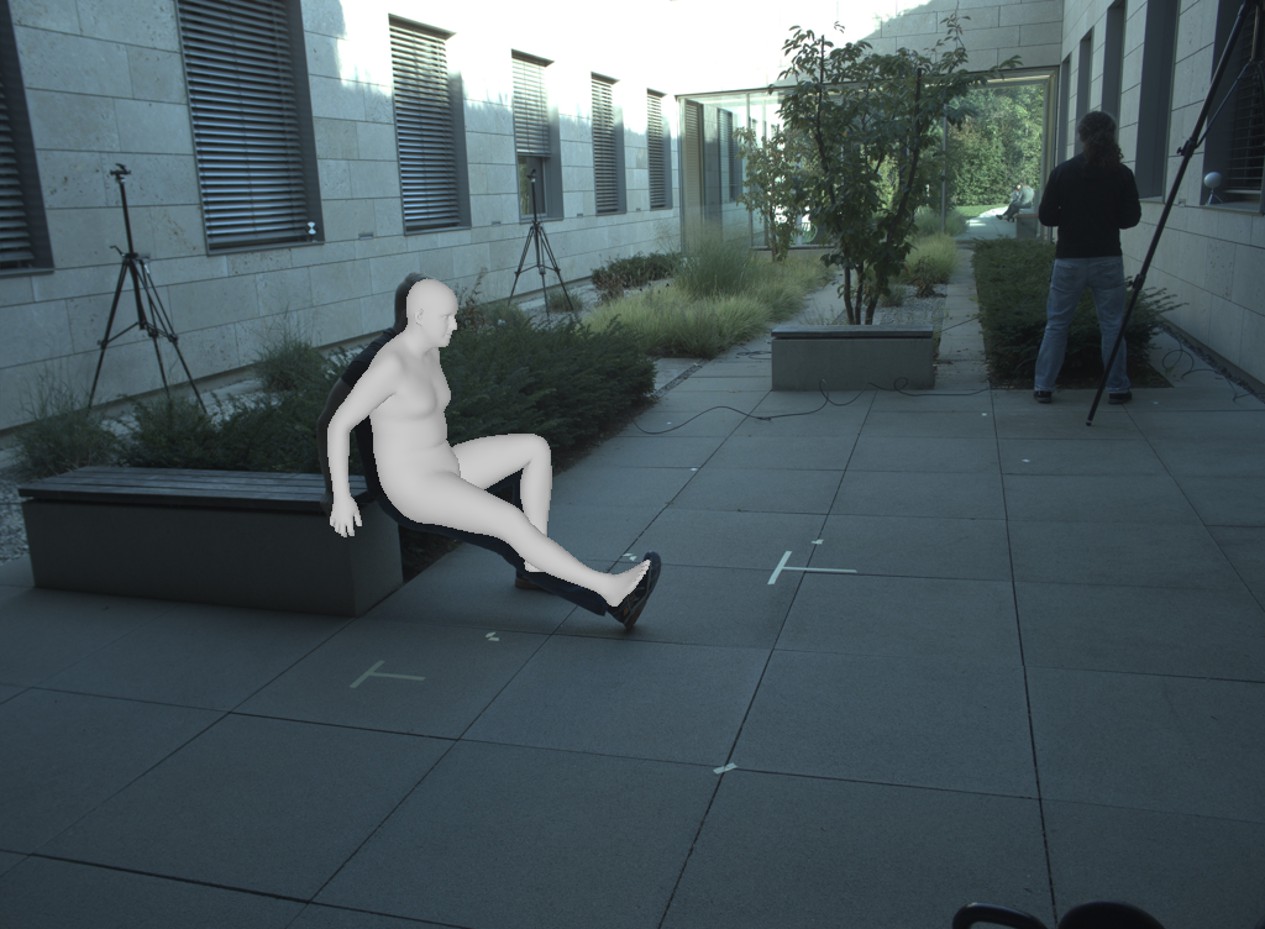}\\[-1pt]
            \end{minipage}
        \end{minipage}%
    }

    \vspace{0.5em}

    \makebox[\linewidth][c]{%
        \begin{minipage}[t]{\roww}
            \centering
            \begin{minipage}[t]{\appfigsevencolwidth}
                \centering
                \includegraphics[width=\linewidth]{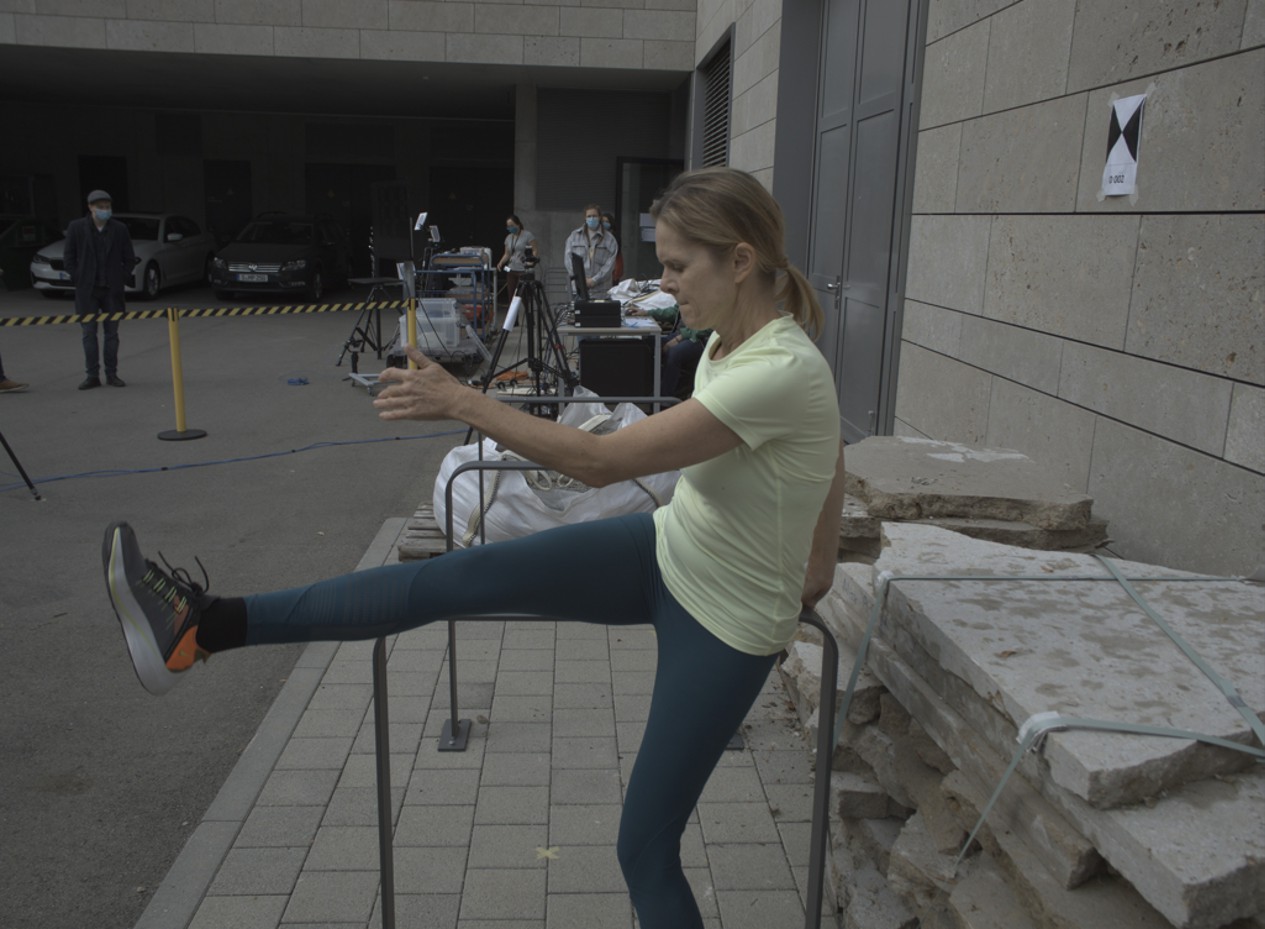}\\[-1pt]
            \end{minipage}\hspace{\appfigmargin}
            \begin{minipage}[t]{\appfigsevencolwidth}
                \centering
                \includegraphics[width=\linewidth]{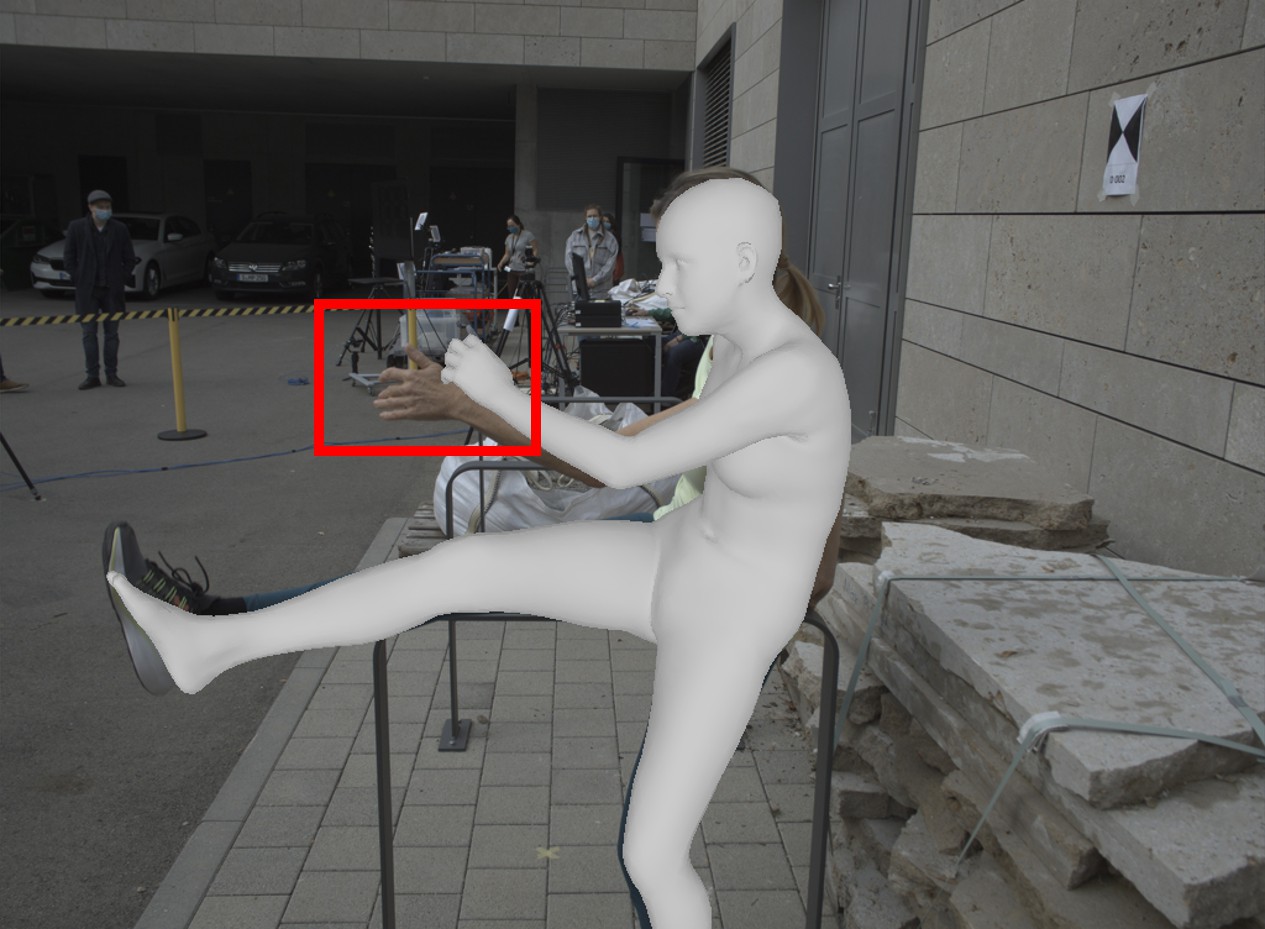}\\[-1pt]
            \end{minipage}\hspace{\appfigmargin}
            \begin{minipage}[t]{\appfigsevencolwidth}
                \centering
                \includegraphics[width=\linewidth]{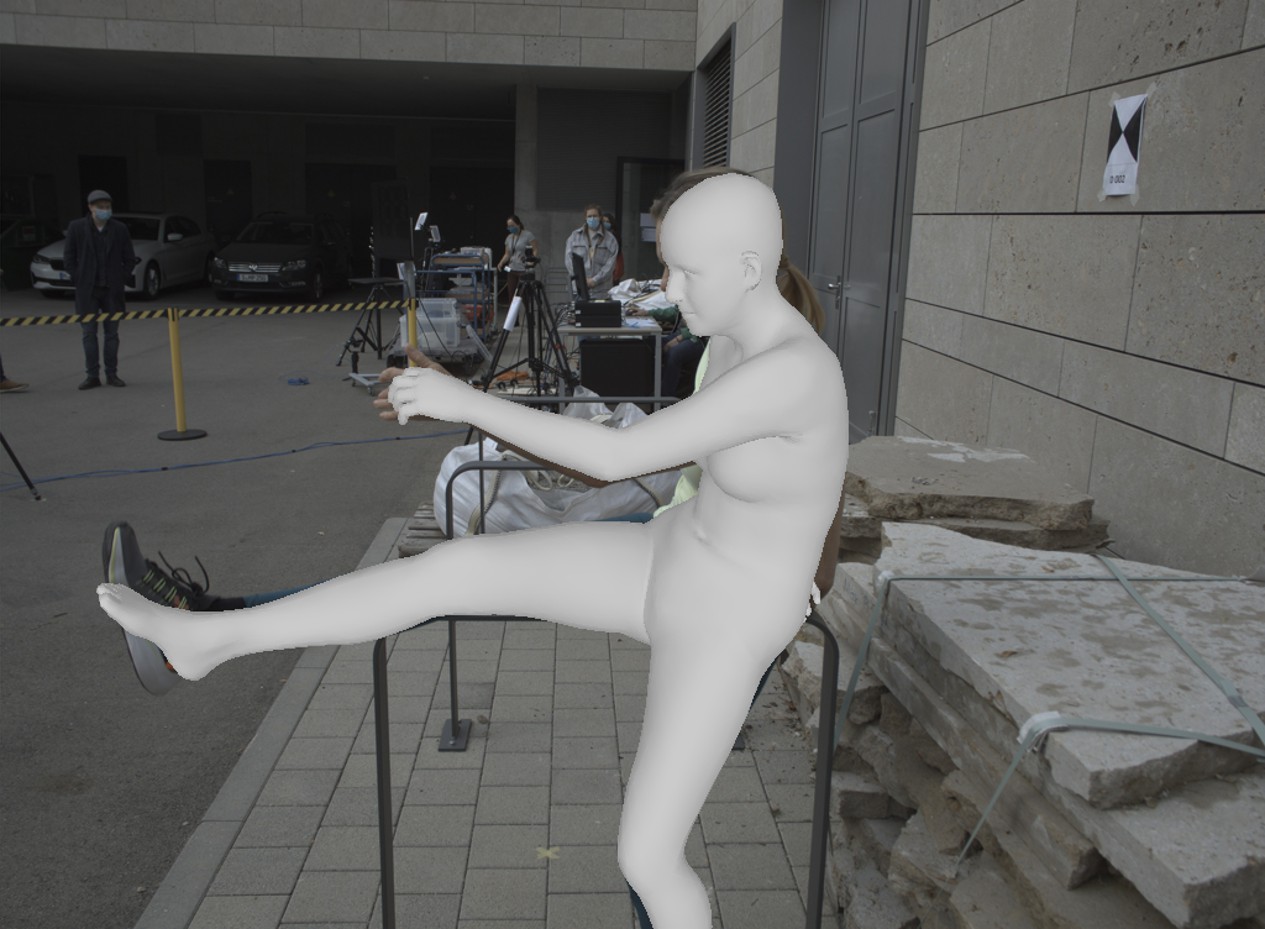}\\[-1pt]
            \end{minipage}
        \end{minipage}%
    }

    \caption{Additional body-pose recovery comparisons between our method and GVHMR \cite{shen2024gvhmr}}
    \label{fig:appendix_qualitative_single_row_1}
\end{figure}
\clearpage

\begin{figure}[p]
    \centering
    \newcommand{\roww}{1.2\linewidth}
    \def\appfigsevencolwidth{0.31\linewidth}
    \def\appfigmargin{0.01\linewidth}

    \makebox[\linewidth][c]{%
        \begin{minipage}[t]{\roww}
            \centering
            \begin{minipage}[t]{\appfigsevencolwidth}
                \parbox[t][8pt][c]{\linewidth}{\centering\scriptsize Input}\par
                \vspace{2pt}
                \includegraphics[width=\linewidth]{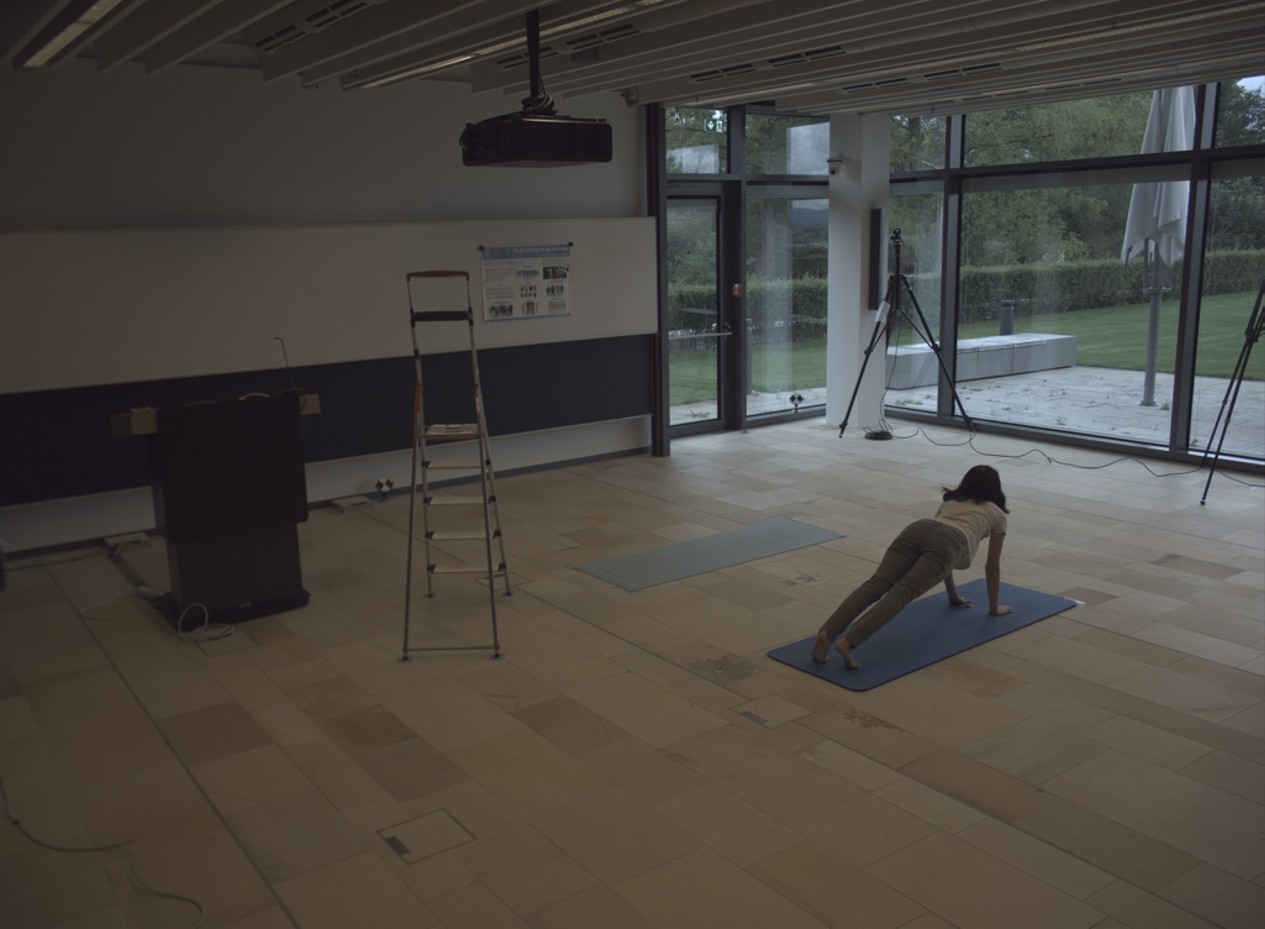}
            \end{minipage}\hspace{\appfigmargin}
            \begin{minipage}[t]{\appfigsevencolwidth}
                \parbox[t][8pt][c]{\linewidth}{\centering\scriptsize GVHMR}\par
                \vspace{2pt}
                \includegraphics[width=\linewidth]{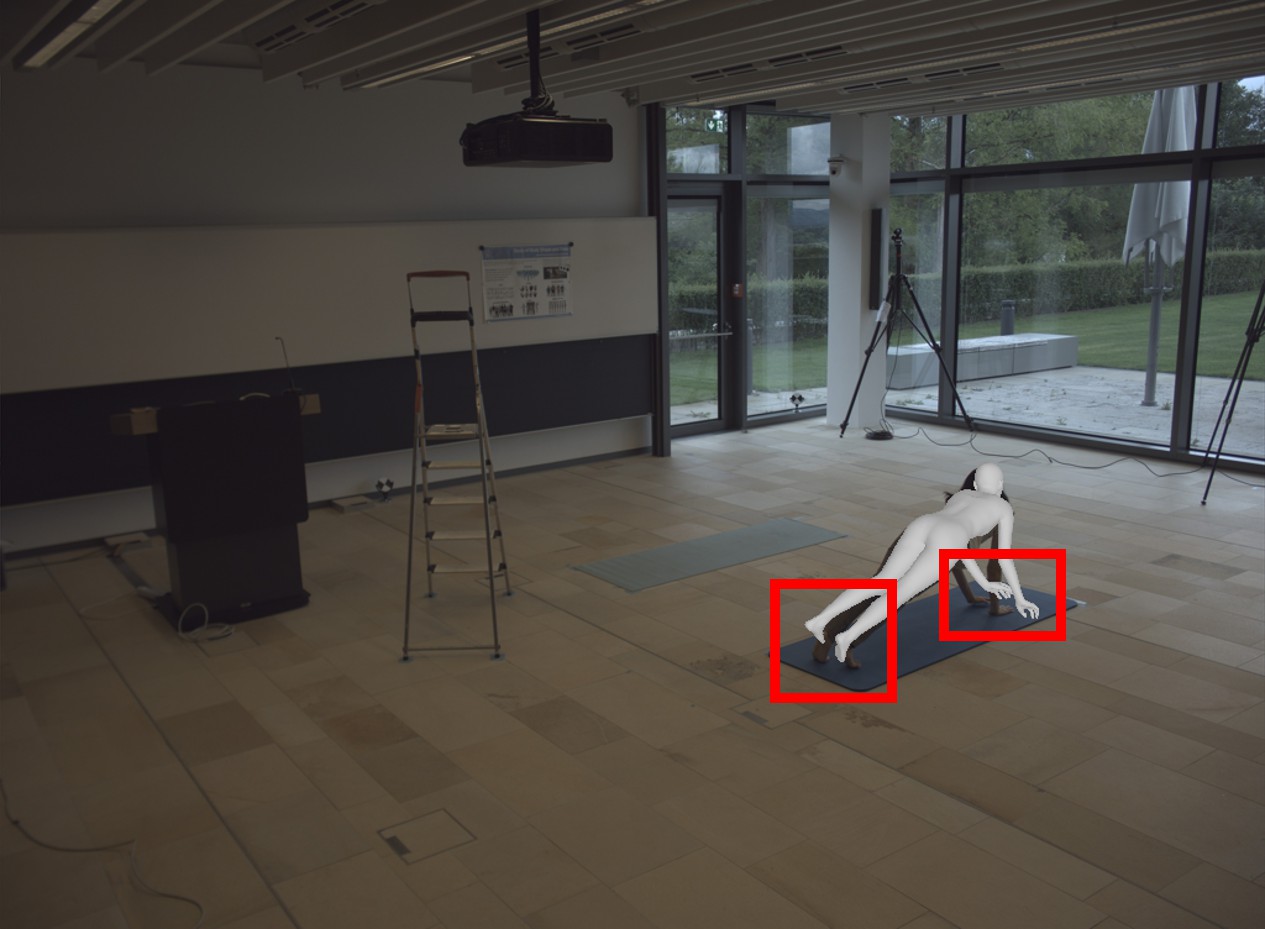}\\[-1pt]
            \end{minipage}\hspace{\appfigmargin}
            \begin{minipage}[t]{\appfigsevencolwidth}
                \parbox[t][8pt][c]{\linewidth}{\centering\scriptsize Ours}\par
                \vspace{2pt}
                \includegraphics[width=\linewidth]{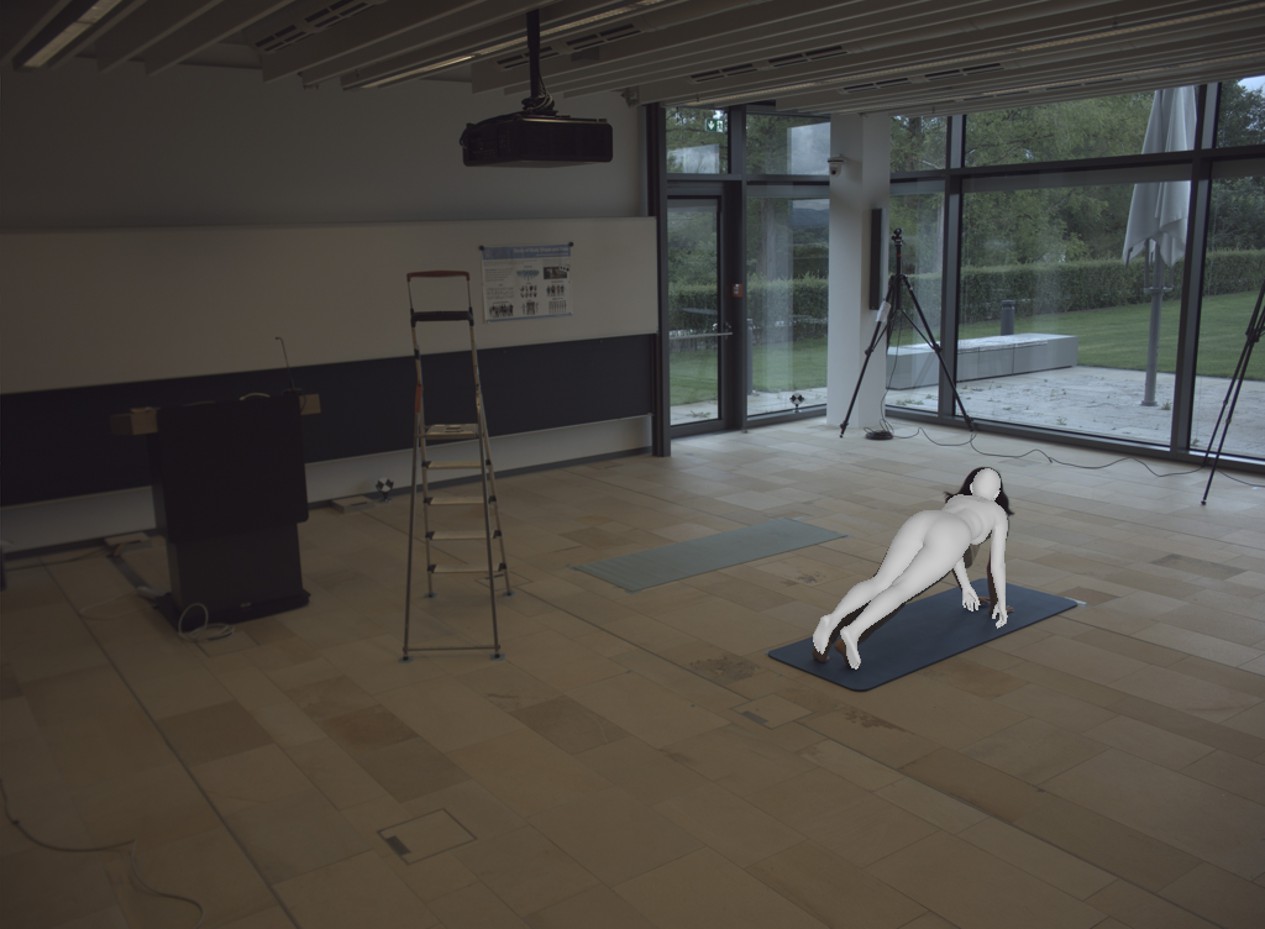}\\[-1pt]
            \end{minipage}
        \end{minipage}%
    }
    
    \vspace{-0.3em}

    \makebox[\linewidth][c]{%
        \begin{minipage}[t]{\roww}
            \centering
            \begin{minipage}[t]{\appfigsevencolwidth}
                \centering
                \includegraphics[width=\linewidth]{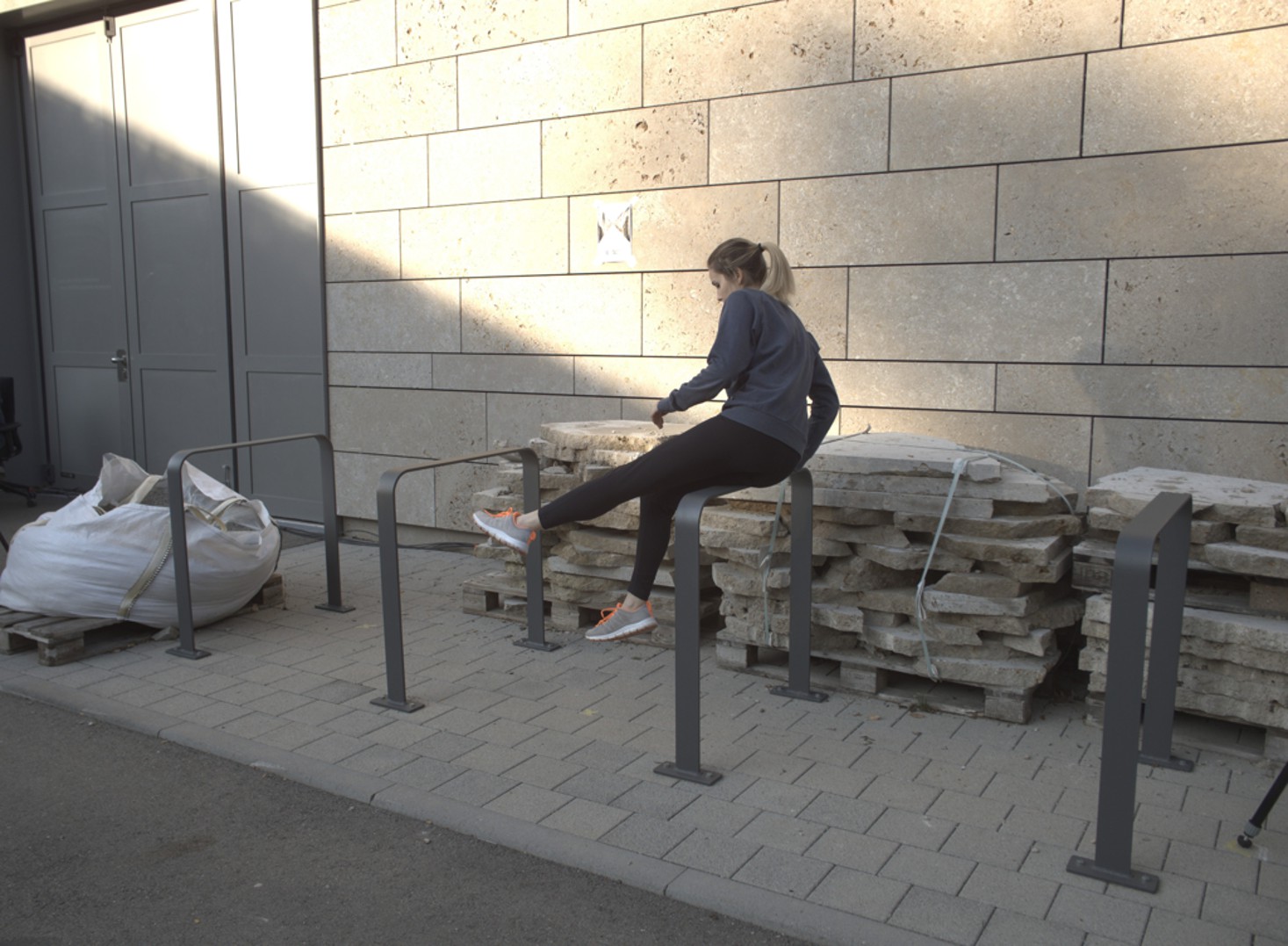}\\[-1pt]
            \end{minipage}\hspace{\appfigmargin}
            \begin{minipage}[t]{\appfigsevencolwidth}
                \centering
                \includegraphics[width=\linewidth]{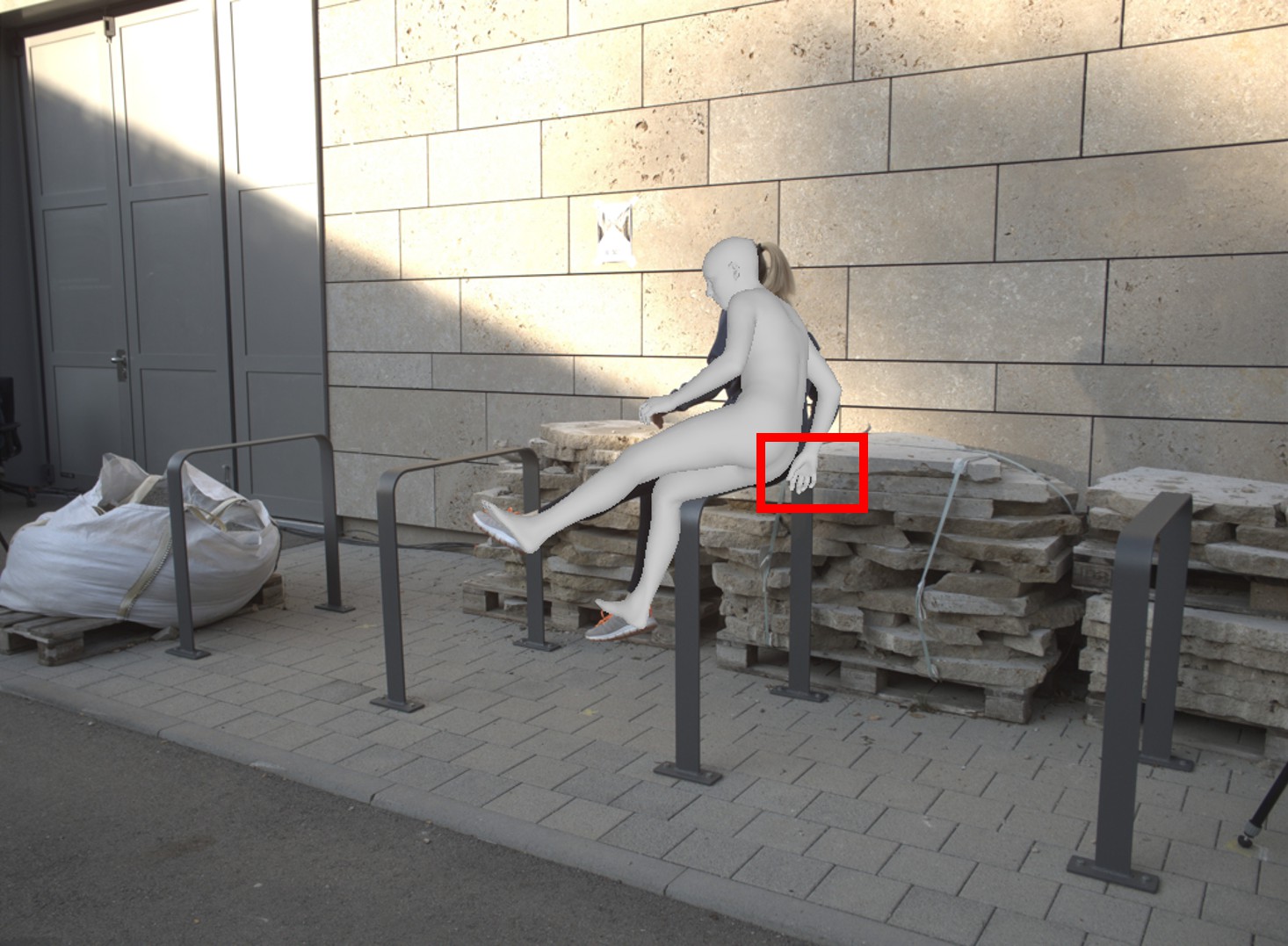}\\[-1pt]
            \end{minipage}\hspace{\appfigmargin}
            \begin{minipage}[t]{\appfigsevencolwidth}
                \centering
                \includegraphics[width=\linewidth]{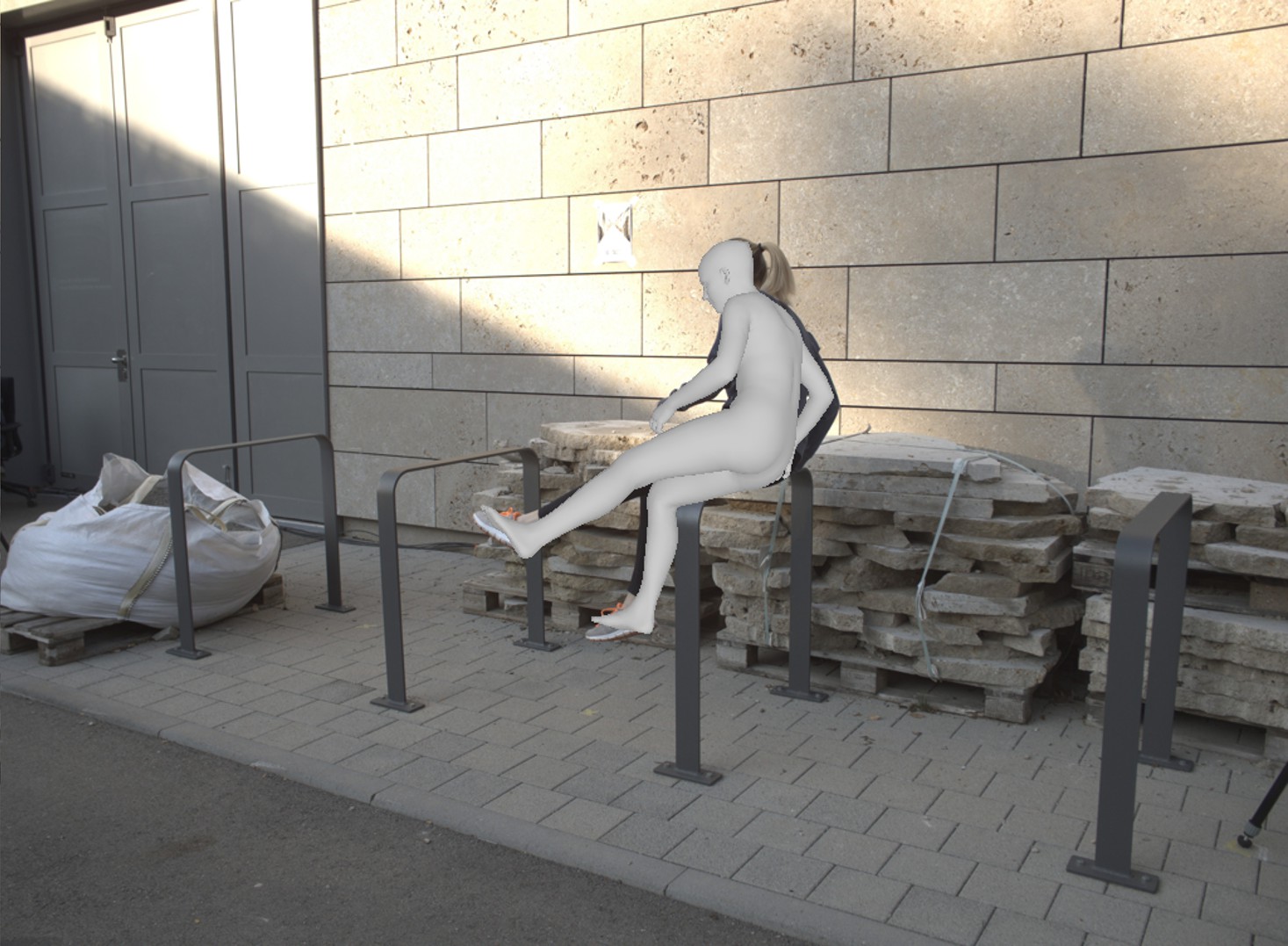}\\[-1pt]
            \end{minipage}
        \end{minipage}%
    }

    \vspace{0.5em}

    \makebox[\linewidth][c]{%
        \begin{minipage}[t]{\roww}
            \centering
            \begin{minipage}[t]{\appfigsevencolwidth}
                \centering
                \includegraphics[width=\linewidth]{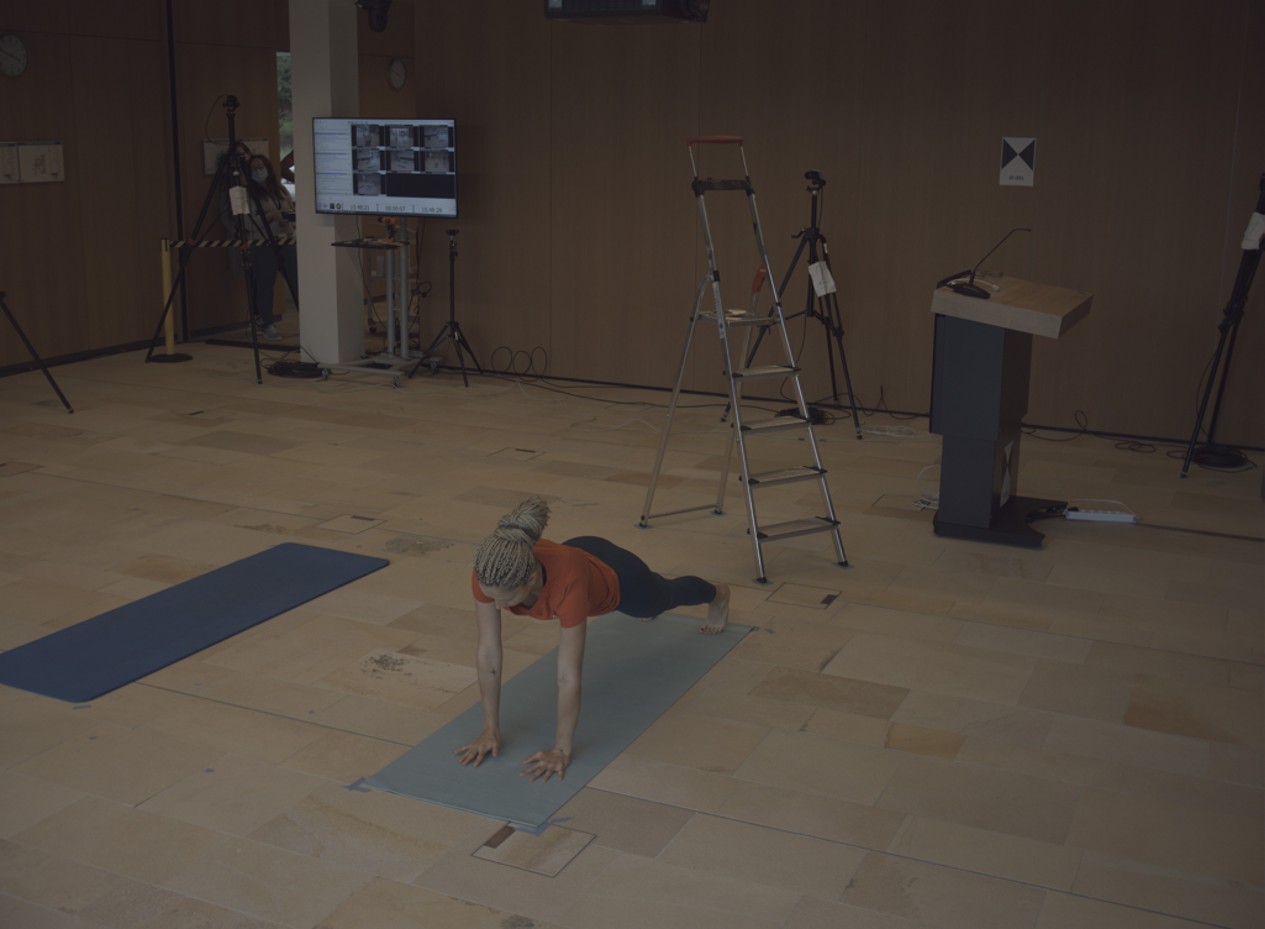}\\[-1pt]
            \end{minipage}\hspace{\appfigmargin}
            \begin{minipage}[t]{\appfigsevencolwidth}
                \centering
                \includegraphics[width=\linewidth]{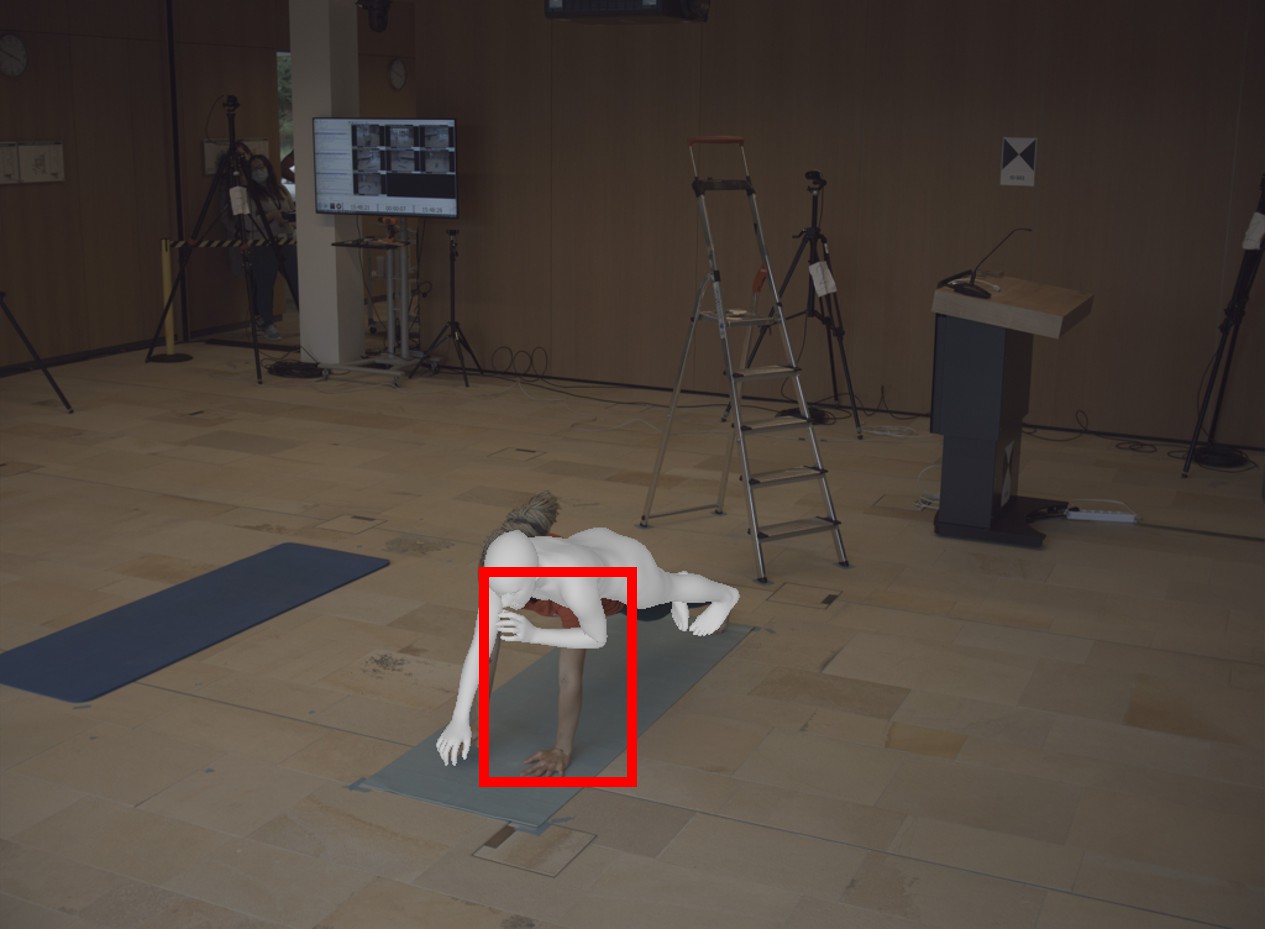}\\[-1pt]
            \end{minipage}\hspace{\appfigmargin}
            \begin{minipage}[t]{\appfigsevencolwidth}
                \centering
                \includegraphics[width=\linewidth]{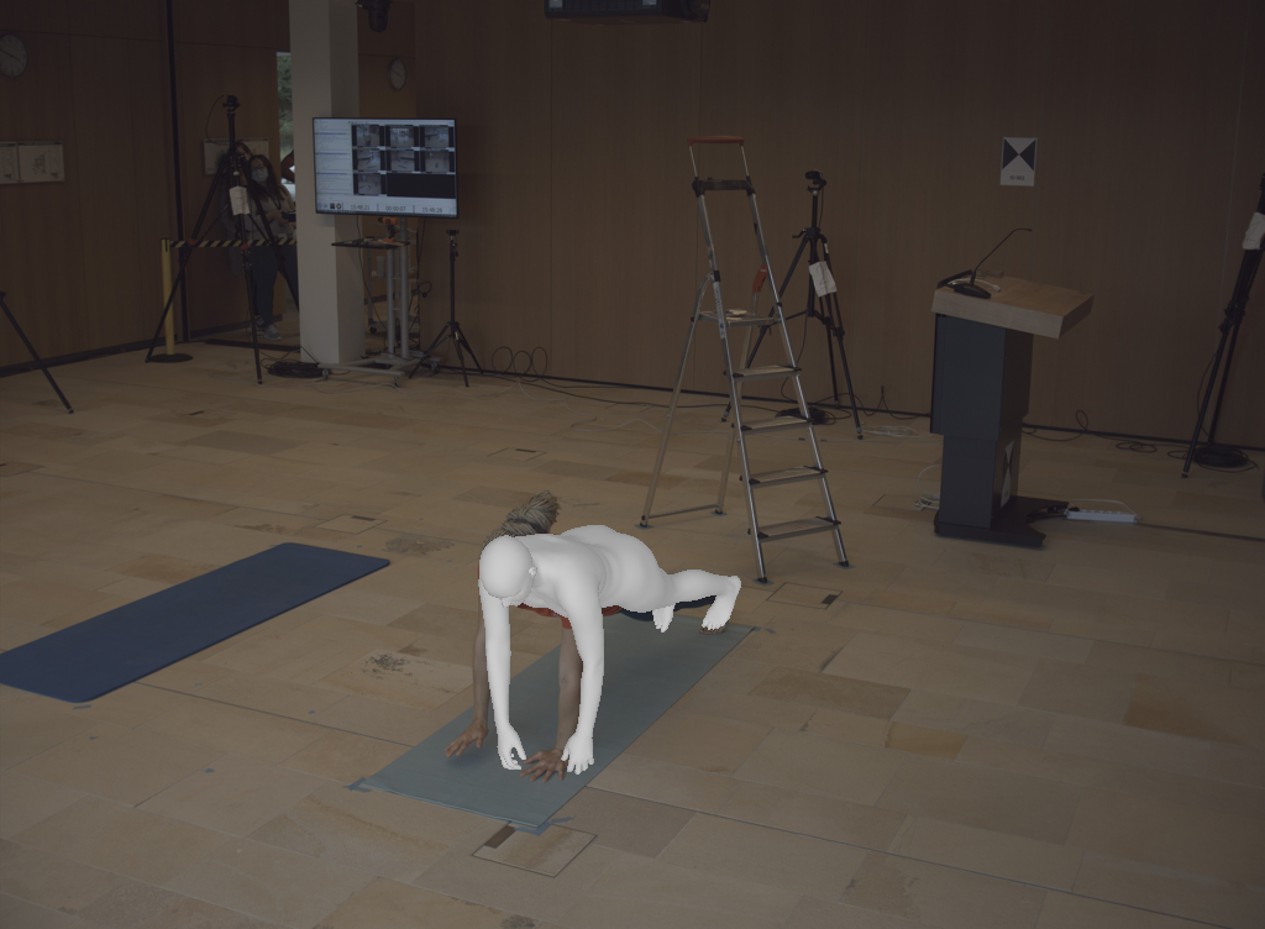}\\[-1pt]
            \end{minipage}
        \end{minipage}%
    }

    \vspace{0.5em}

    \makebox[\linewidth][c]{%
        \begin{minipage}[t]{\roww}
            \centering
            \begin{minipage}[t]{\appfigsevencolwidth}
                \centering
                \includegraphics[width=\linewidth]{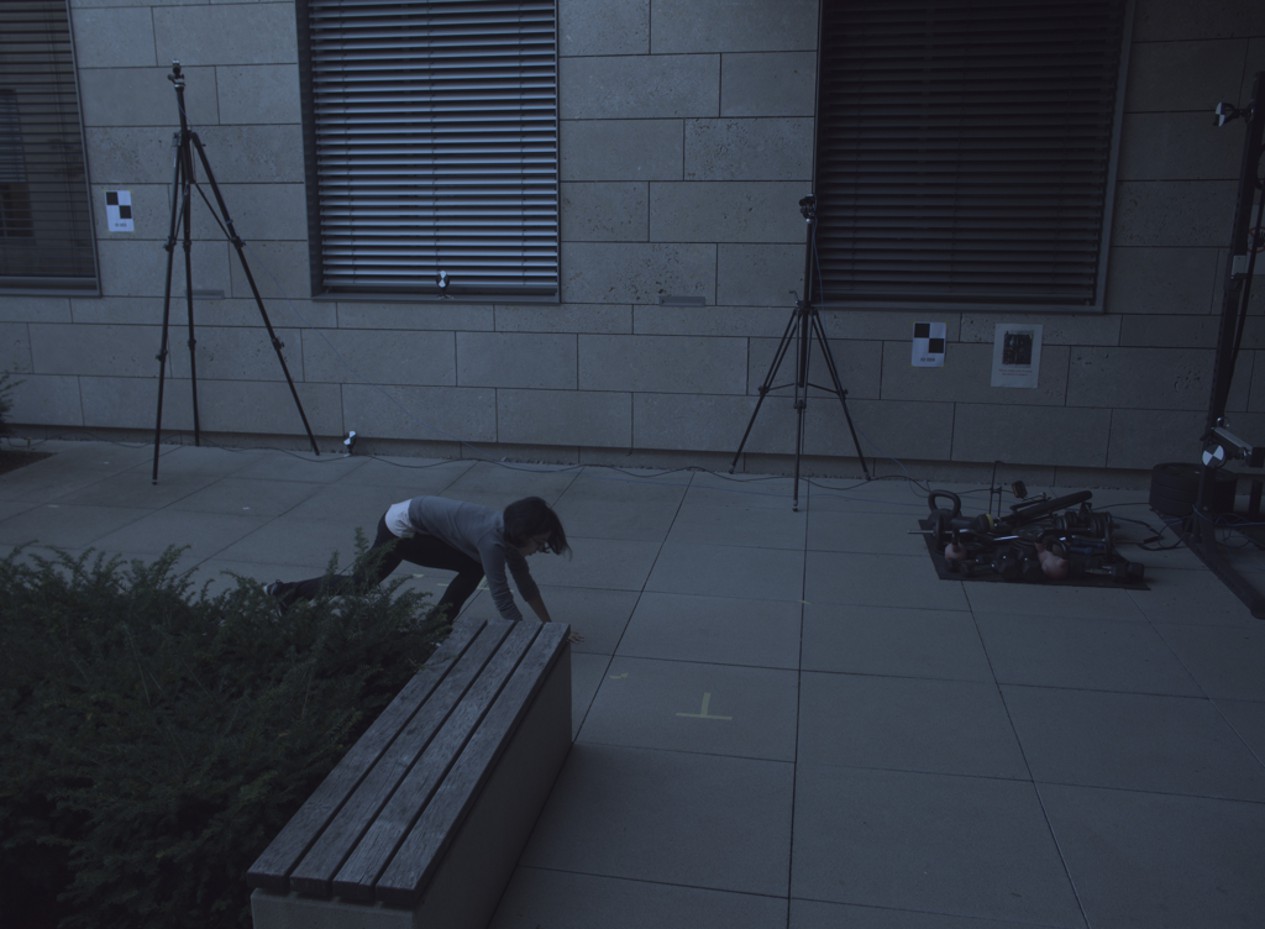}\\[-1pt]
            \end{minipage}\hspace{\appfigmargin}
            \begin{minipage}[t]{\appfigsevencolwidth}
                \centering
                \includegraphics[width=\linewidth]{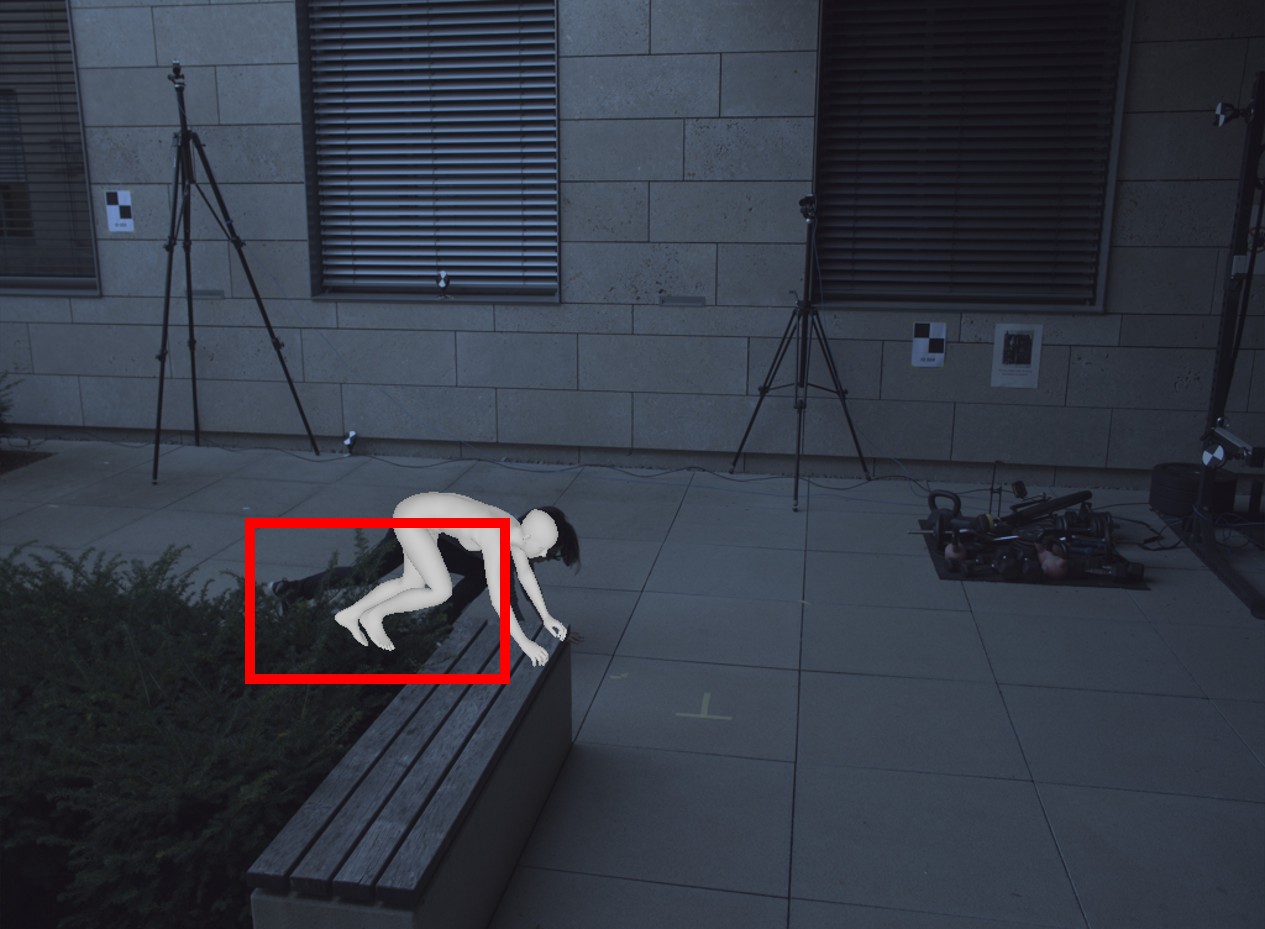}\\[-1pt]
            \end{minipage}\hspace{\appfigmargin}
            \begin{minipage}[t]{\appfigsevencolwidth}
                \centering
                \includegraphics[width=\linewidth]{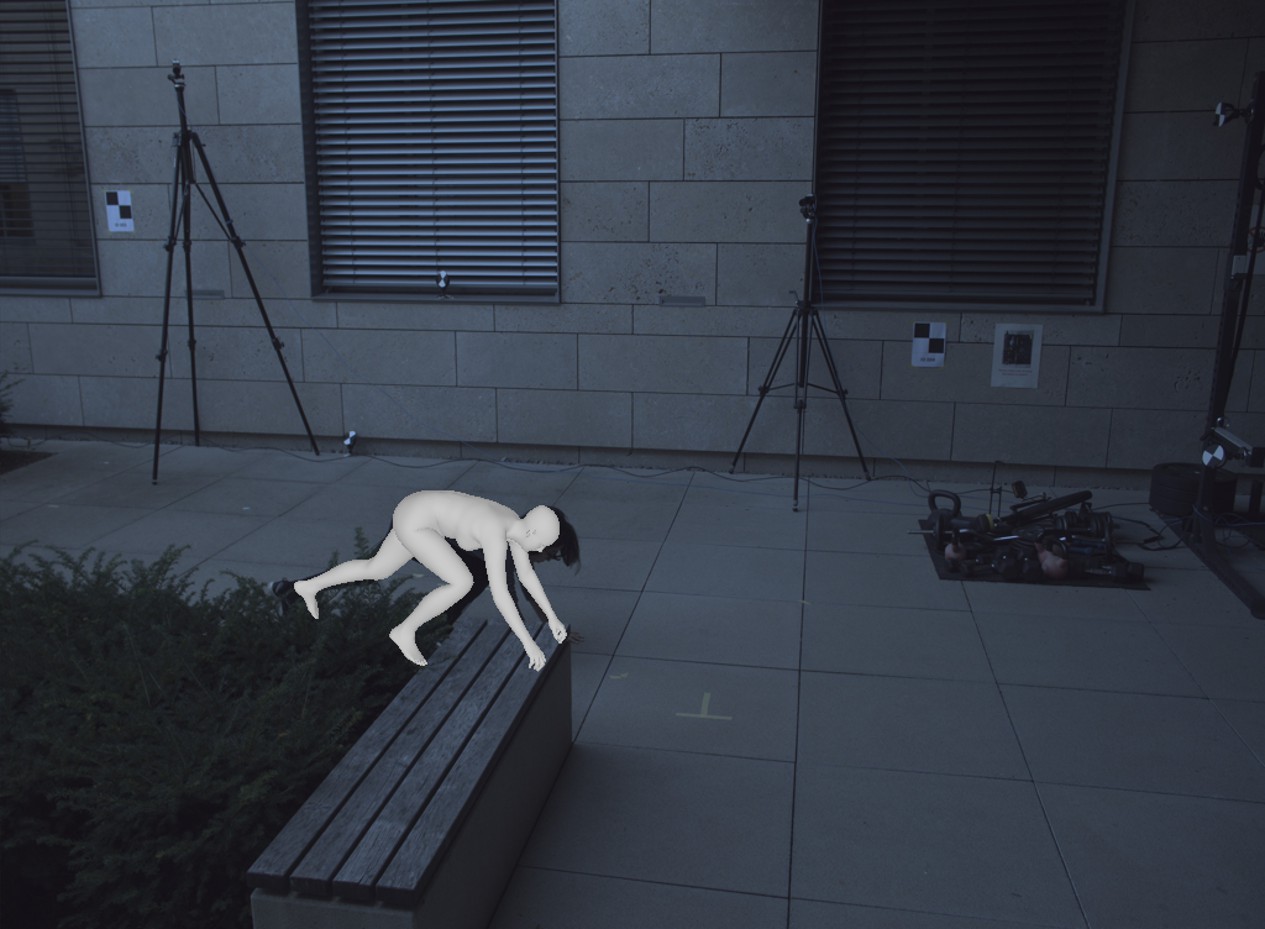}\\[-1pt]
            \end{minipage}
        \end{minipage}%
    }

    \caption{Additional body-pose recovery comparisons between our method and GVHMR \cite{shen2024gvhmr}}
    \label{fig:appendix_qualitative_single_row_2}
\end{figure}
\clearpage

\begin{figure}[p]
    \centering
    \newcommand{\roww}{1.2\linewidth}
    \def\appfigsevencolwidth{0.31\linewidth}
    \def\appfigmargin{0.01\linewidth}

    \makebox[\linewidth][c]{%
        \begin{minipage}[t]{\roww}
            \centering
            \begin{minipage}[t]{\appfigsevencolwidth}
                \parbox[t][8pt][c]{\linewidth}{\centering\scriptsize Input}\par
                \vspace{2pt}
                \includegraphics[width=\linewidth]{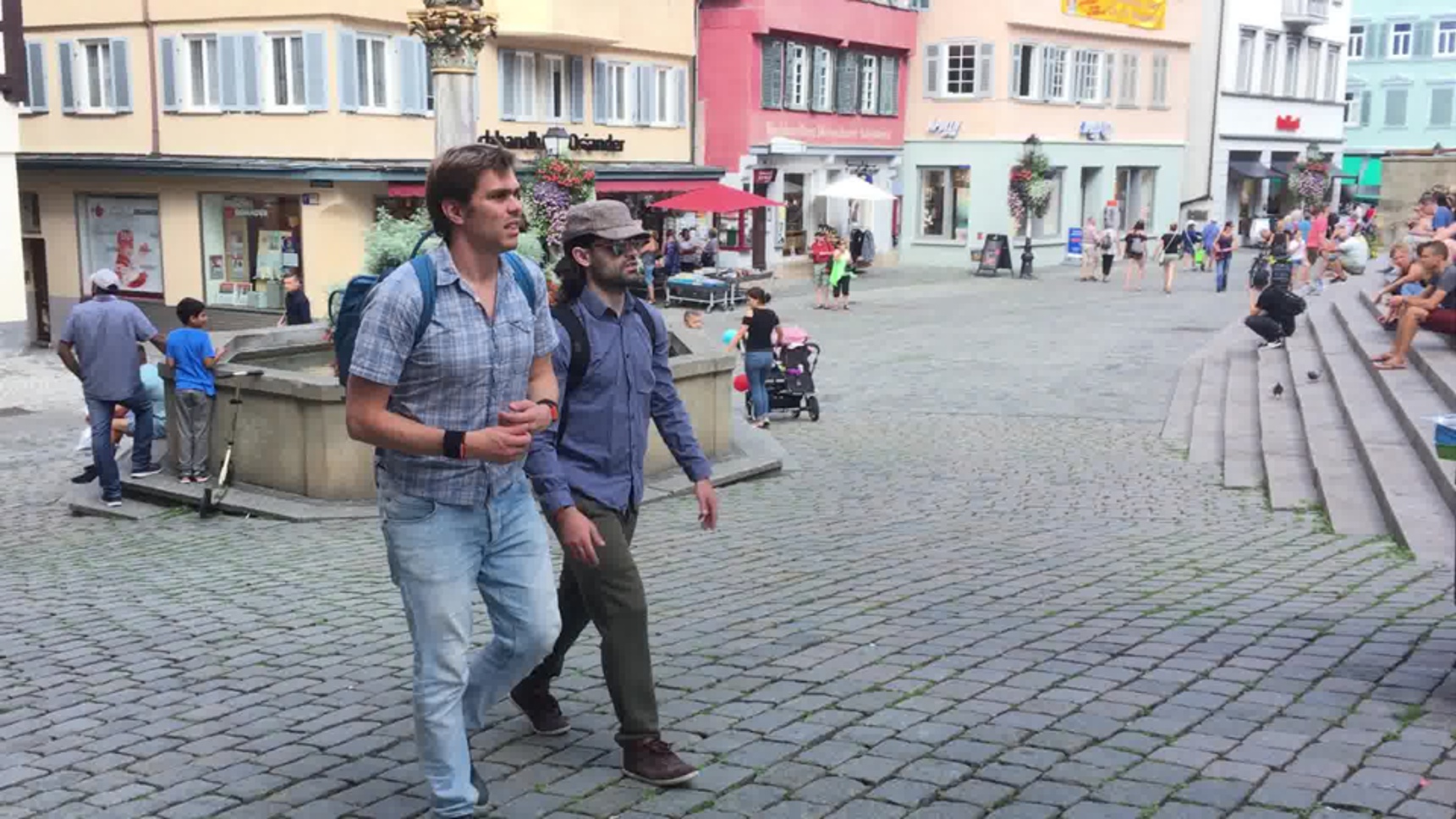}
            \end{minipage}\hspace{\appfigmargin}
            \begin{minipage}[t]{\appfigsevencolwidth}
                \parbox[t][8pt][c]{\linewidth}{\centering\scriptsize GENMO}\par
                \vspace{2pt}
                \includegraphics[width=\linewidth]{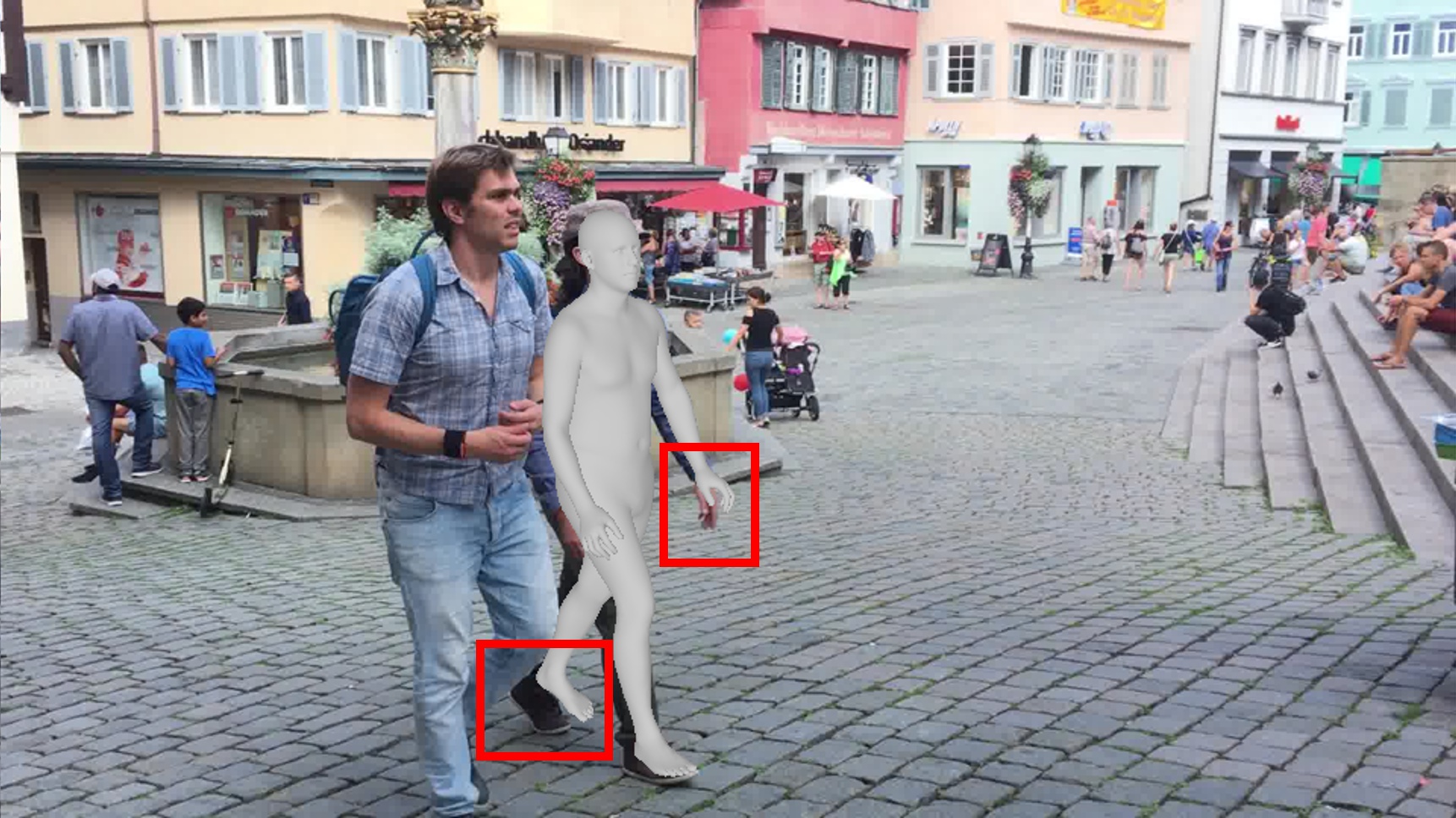}\\[-1pt]
            \end{minipage}\hspace{\appfigmargin}
            \begin{minipage}[t]{\appfigsevencolwidth}
                \parbox[t][8pt][c]{\linewidth}{\centering\scriptsize Ours}\par
                \vspace{2pt}
                \includegraphics[width=\linewidth]{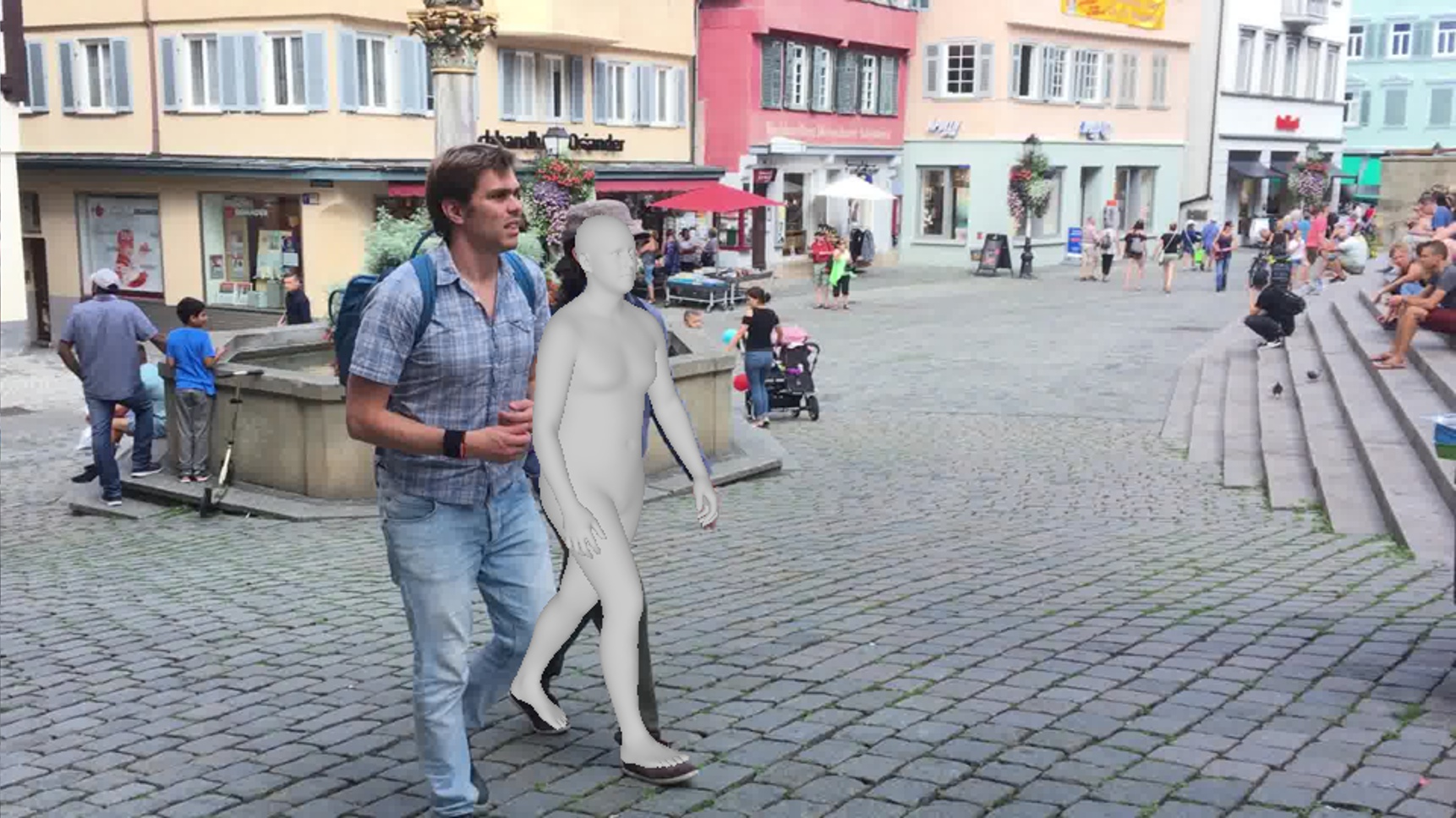}\\[-1pt]
            \end{minipage}
        \end{minipage}%
    }
    
    \vspace{-0.3em}

    \makebox[\linewidth][c]{%
        \begin{minipage}[t]{\roww}
            \centering
            \begin{minipage}[t]{\appfigsevencolwidth}
                \centering
                \includegraphics[width=\linewidth]{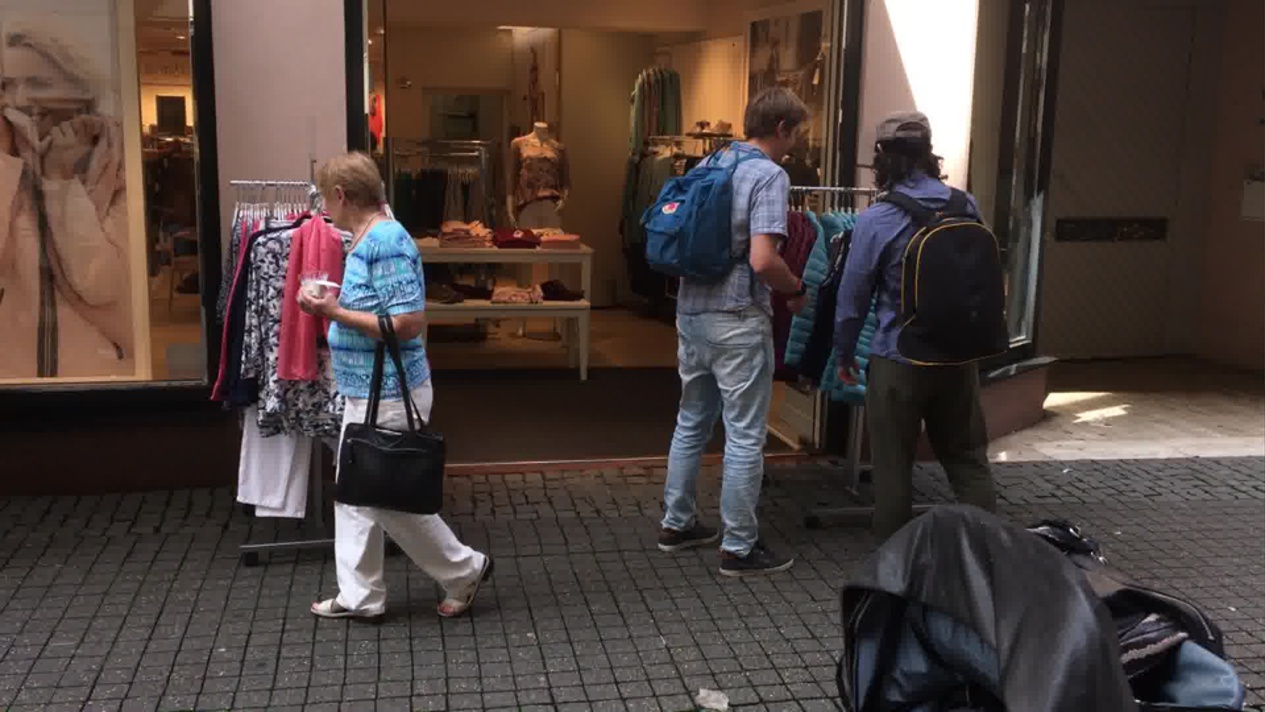}\\[-1pt]
            \end{minipage}\hspace{\appfigmargin}
            \begin{minipage}[t]{\appfigsevencolwidth}
                \centering
                \includegraphics[width=\linewidth]{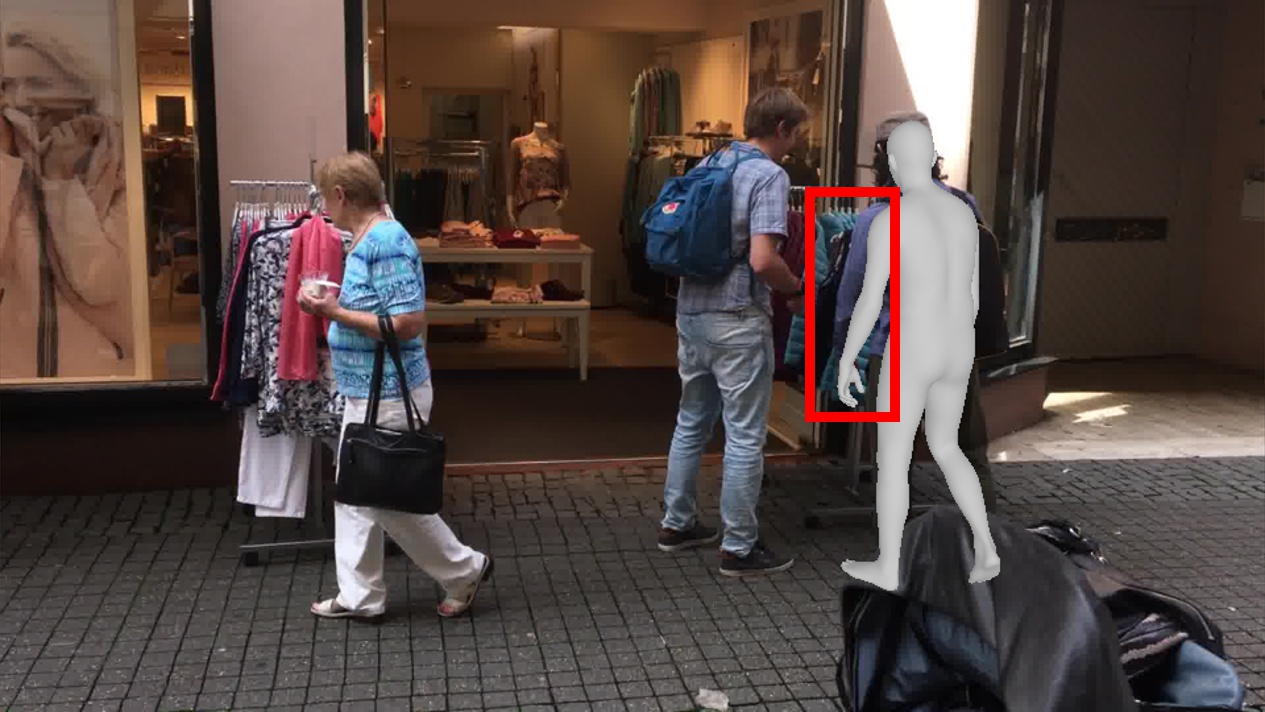}\\[-1pt]
            \end{minipage}\hspace{\appfigmargin}
            \begin{minipage}[t]{\appfigsevencolwidth}
                \centering
                \includegraphics[width=\linewidth]{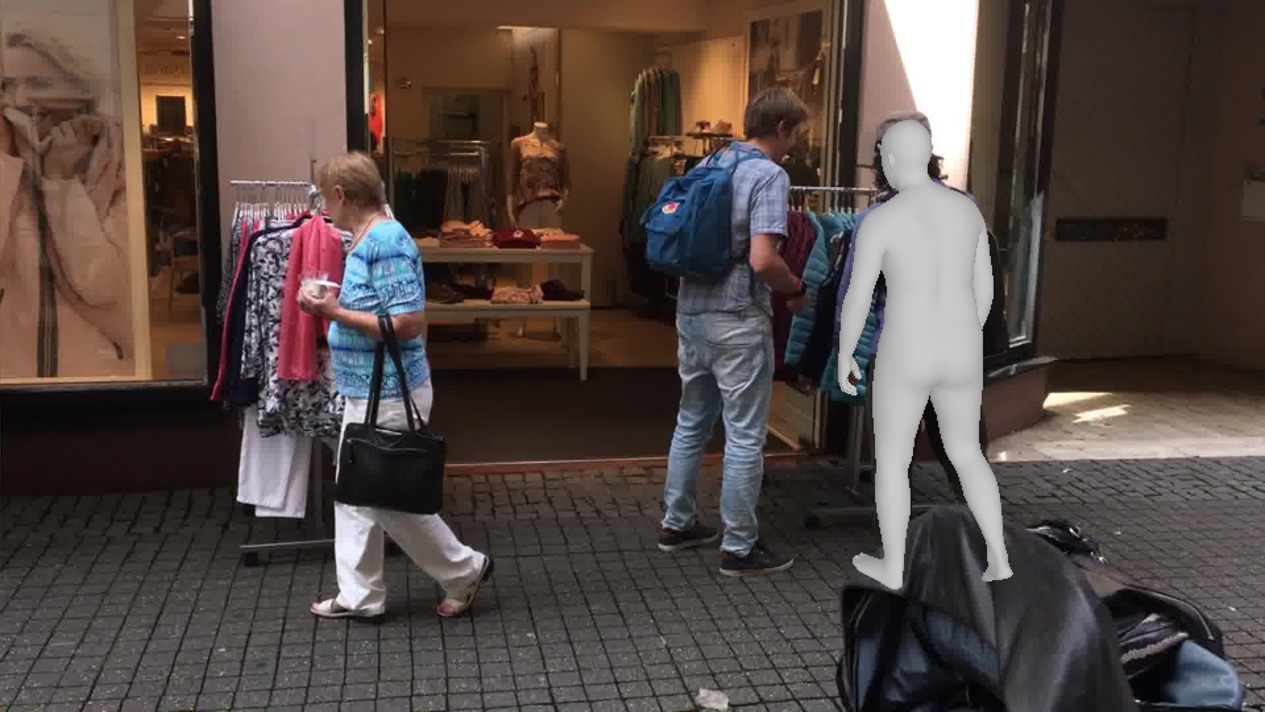}\\[-1pt]
            \end{minipage}
        \end{minipage}%
    }

    \vspace{0.5em}

    \makebox[\linewidth][c]{%
        \begin{minipage}[t]{\roww}
            \centering
            \begin{minipage}[t]{\appfigsevencolwidth}
                \centering
                \includegraphics[width=\linewidth]{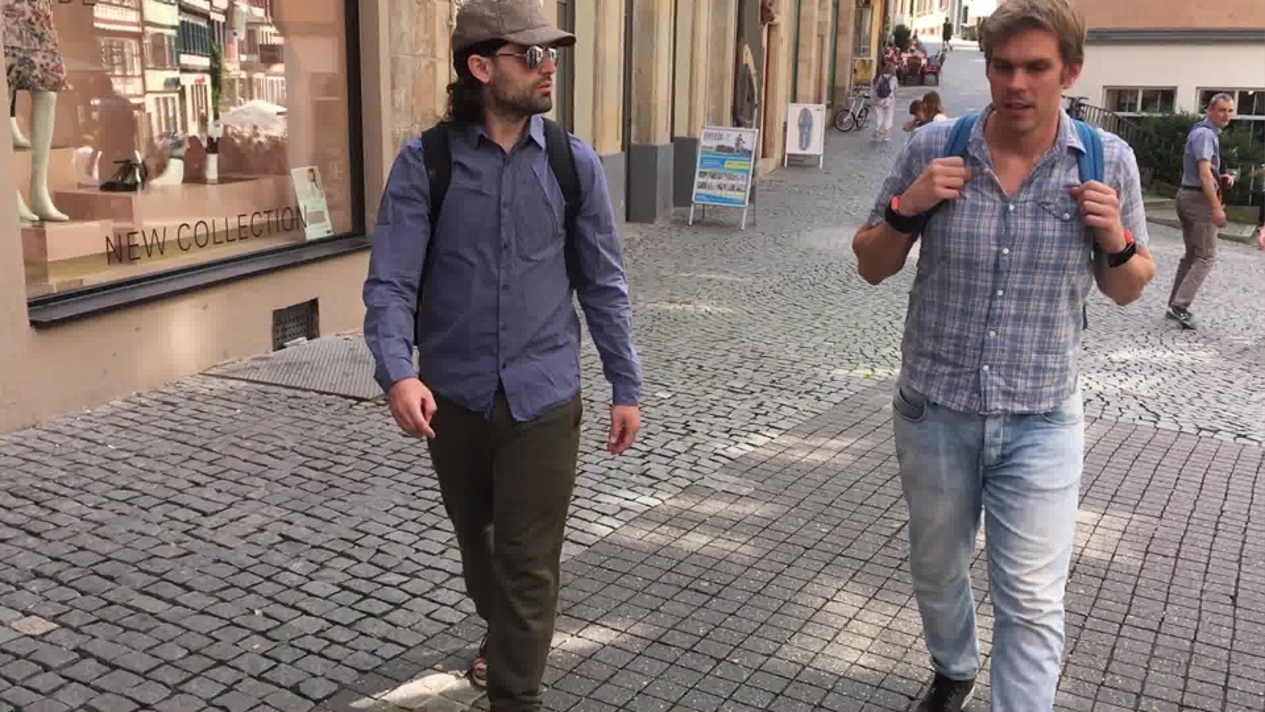}\\[-1pt]
            \end{minipage}\hspace{\appfigmargin}
            \begin{minipage}[t]{\appfigsevencolwidth}
                \centering
                \includegraphics[width=\linewidth]{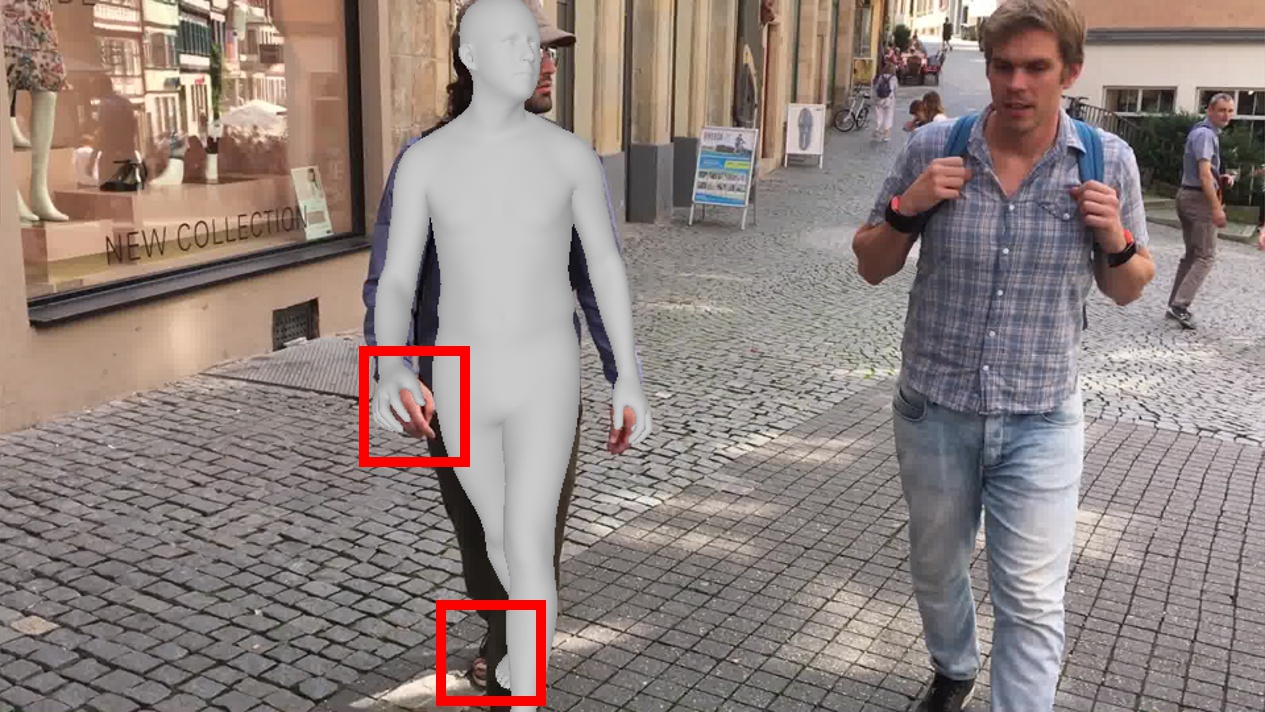}\\[-1pt]
            \end{minipage}\hspace{\appfigmargin}
            \begin{minipage}[t]{\appfigsevencolwidth}
                \centering
                \includegraphics[width=\linewidth]{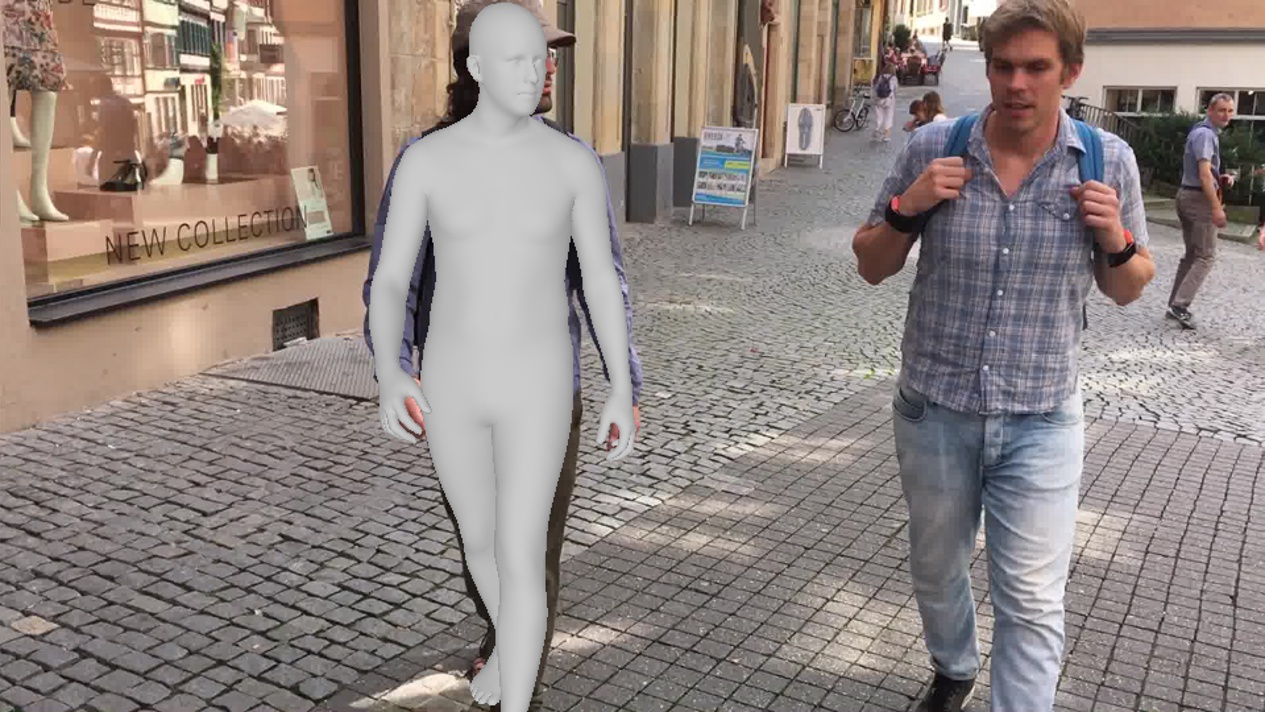}\\[-1pt]
            \end{minipage}
        \end{minipage}%
    }

    \vspace{0.5em}

    \makebox[\linewidth][c]{%
        \begin{minipage}[t]{\roww}
            \centering
            \begin{minipage}[t]{\appfigsevencolwidth}
                \centering
                \includegraphics[width=\linewidth]{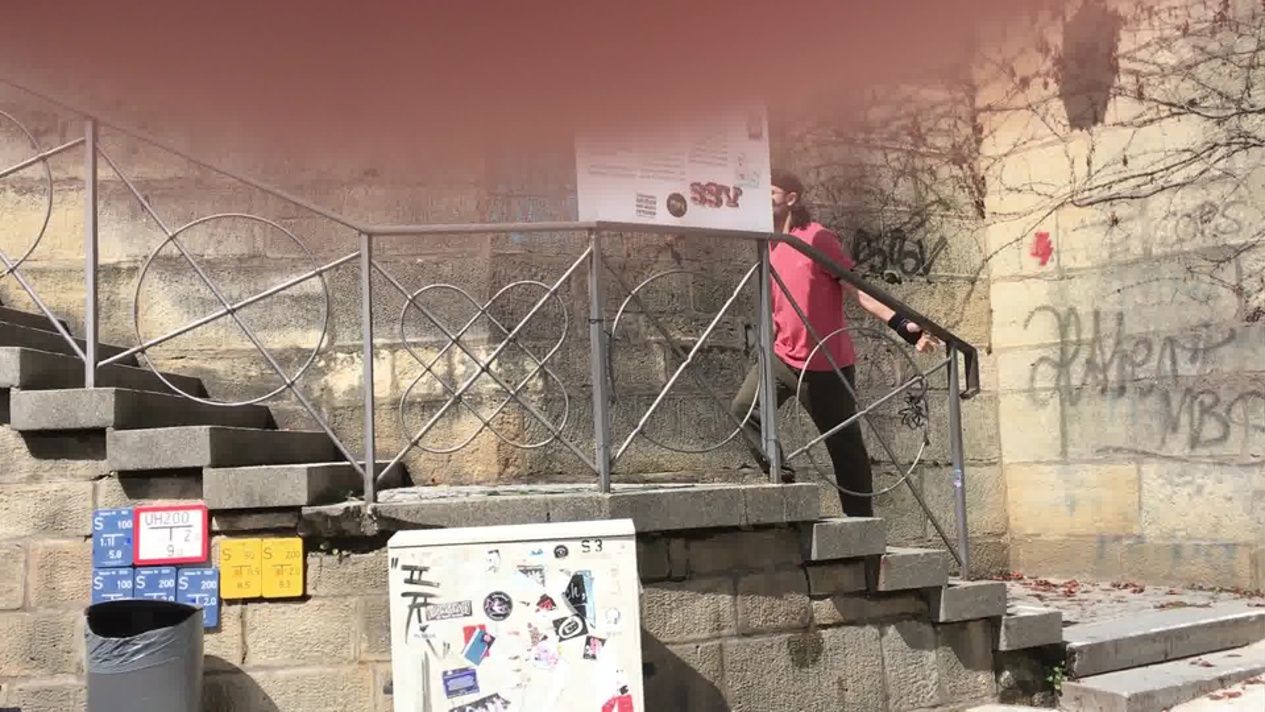}\\[-1pt]
            \end{minipage}\hspace{\appfigmargin}
            \begin{minipage}[t]{\appfigsevencolwidth}
                \centering
                \includegraphics[width=\linewidth]{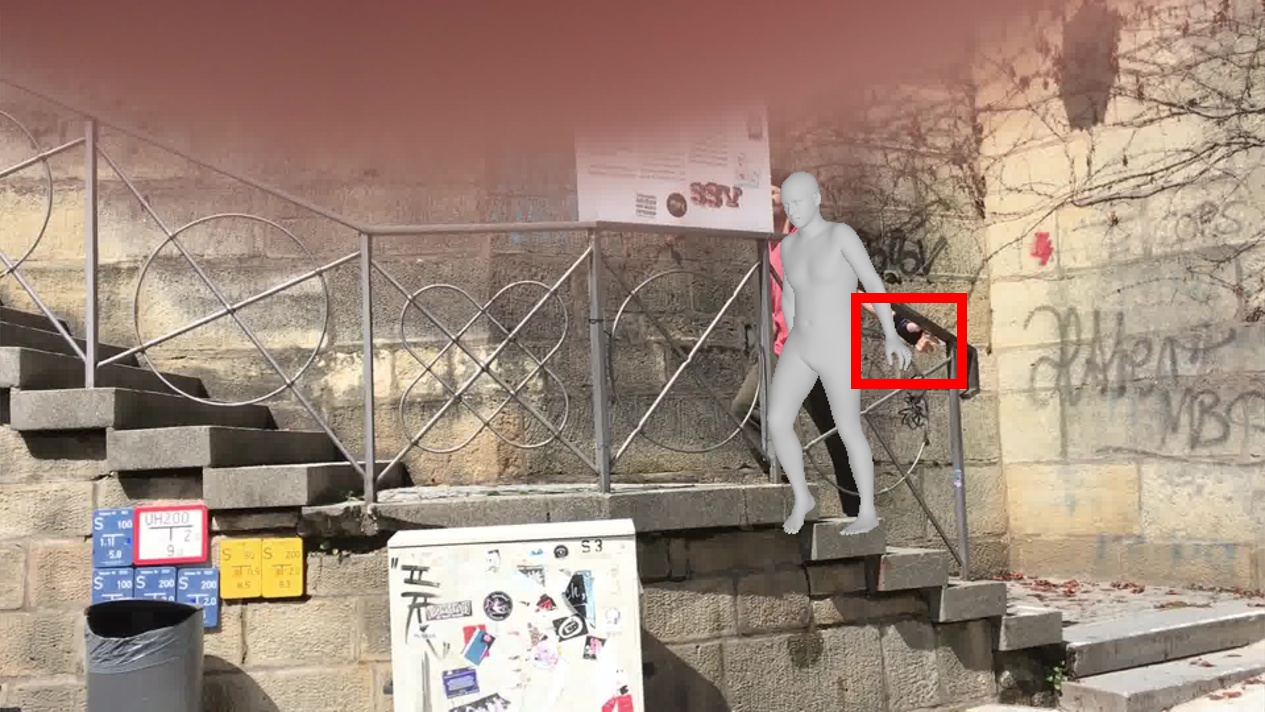}\\[-1pt]
            \end{minipage}\hspace{\appfigmargin}
            \begin{minipage}[t]{\appfigsevencolwidth}
                \centering
                \includegraphics[width=\linewidth]{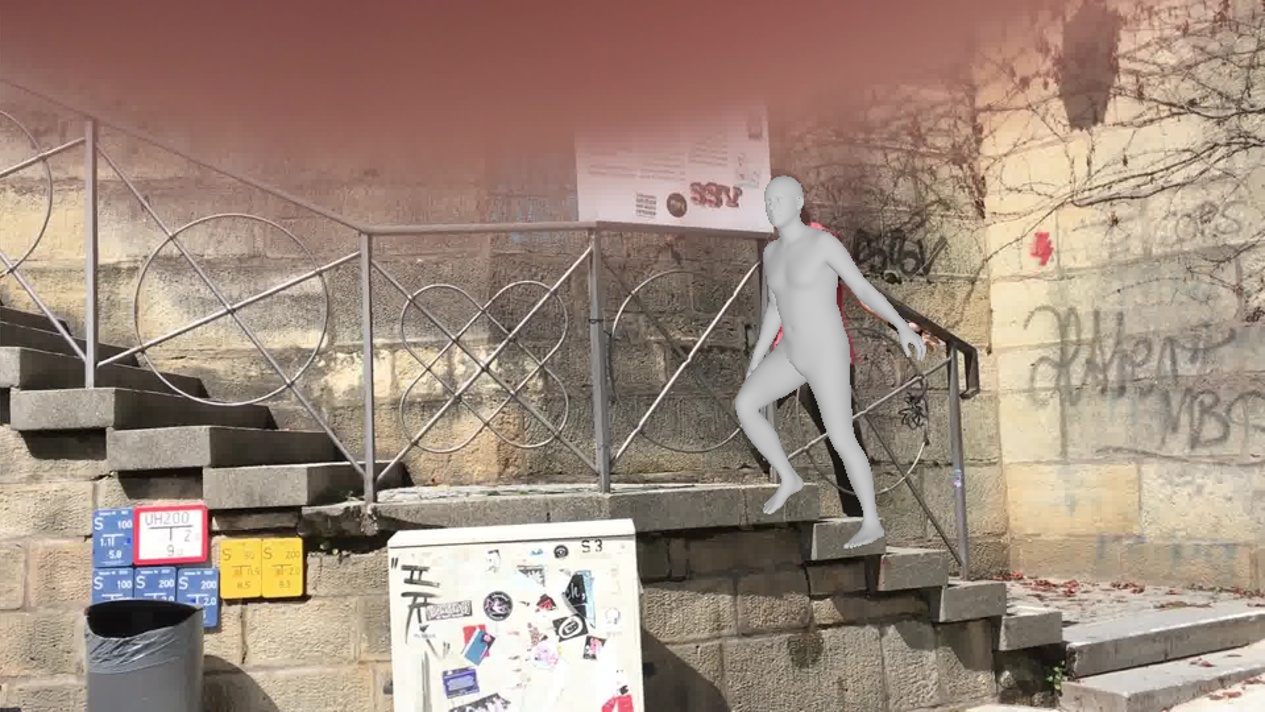}\\[-1pt]
            \end{minipage}
        \end{minipage}%
    }

    \vspace{0.5em}

    \makebox[\linewidth][c]{%
        \begin{minipage}[t]{\roww}
            \centering
            \begin{minipage}[t]{\appfigsevencolwidth}
                \centering
                \includegraphics[width=\linewidth]{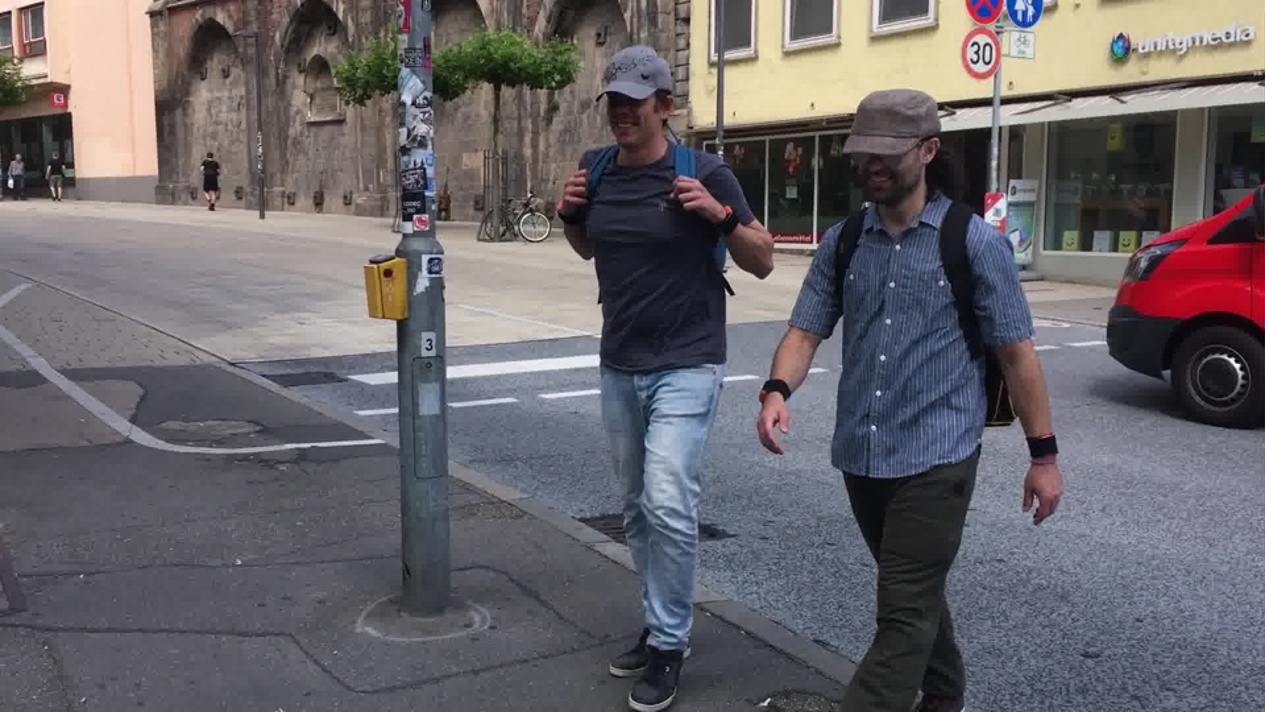}\\[-1pt]
            \end{minipage}\hspace{\appfigmargin}
            \begin{minipage}[t]{\appfigsevencolwidth}
                \centering
                \includegraphics[width=\linewidth]{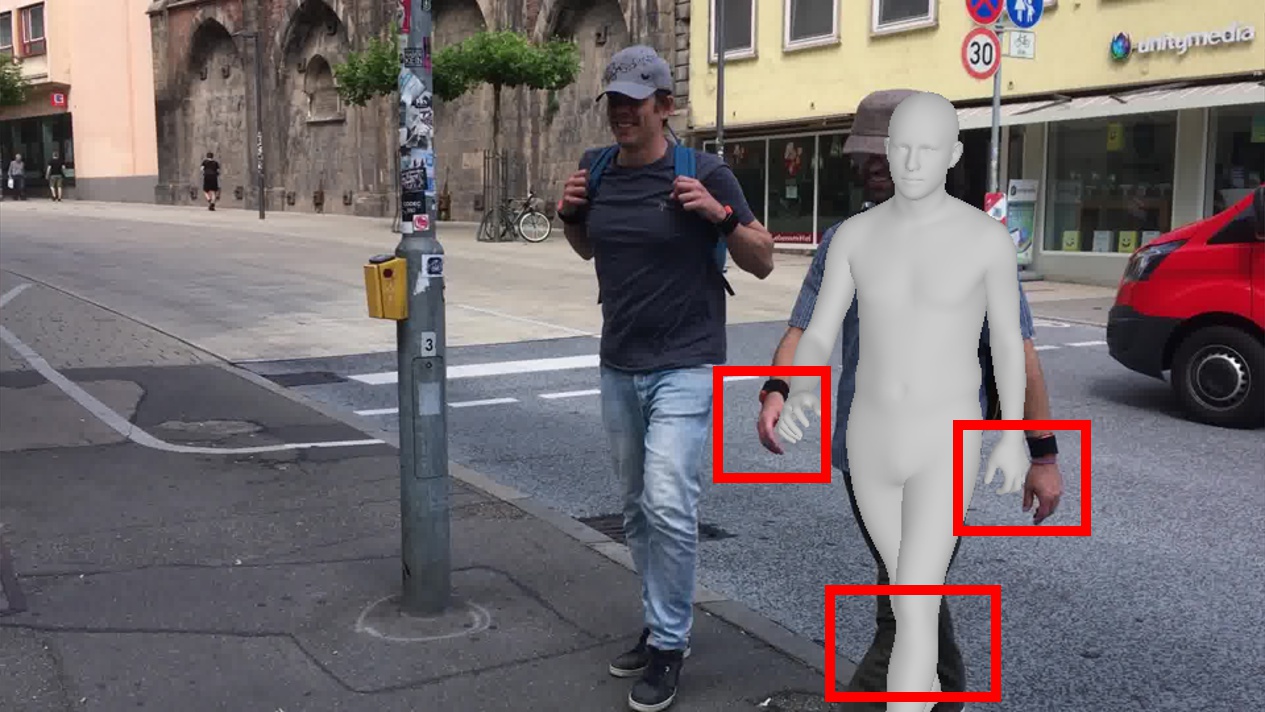}\\[-1pt]
            \end{minipage}\hspace{\appfigmargin}
            \begin{minipage}[t]{\appfigsevencolwidth}
                \centering
                \includegraphics[width=\linewidth]{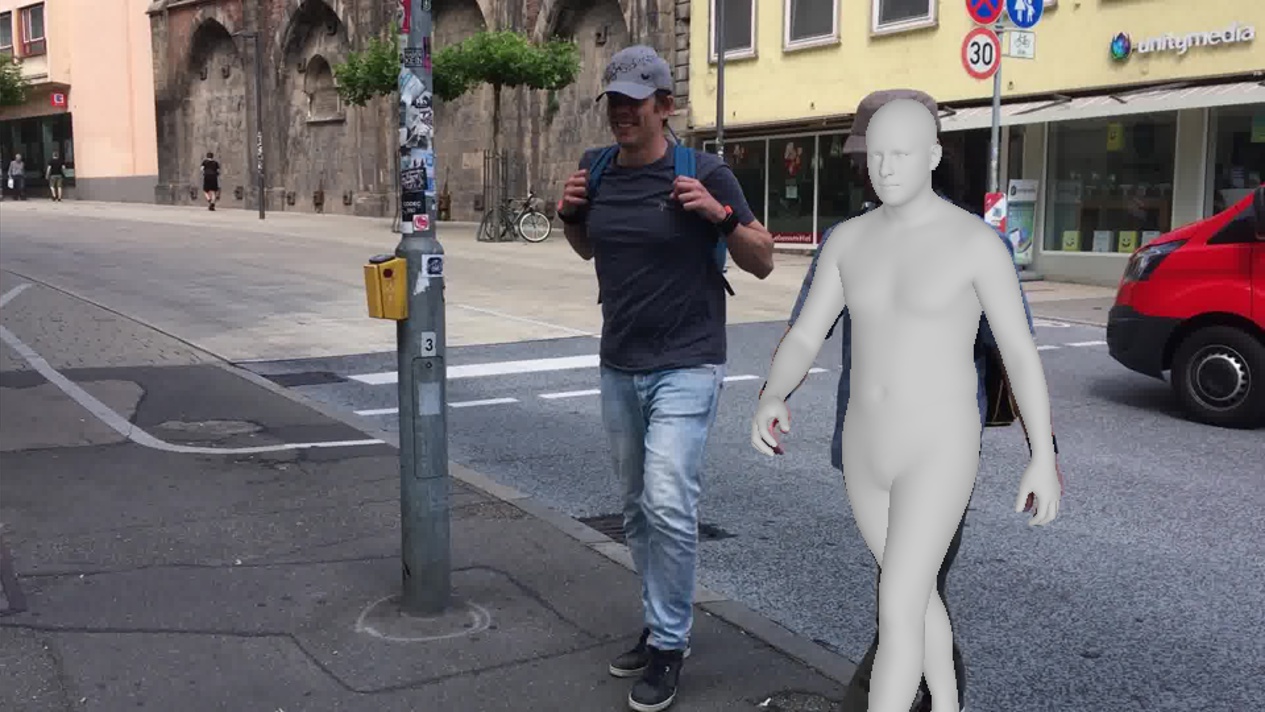}\\[-1pt]
            \end{minipage}
        \end{minipage}%
    }

    \vspace{0.5em}

    \makebox[\linewidth][c]{%
        \begin{minipage}[t]{\roww}
            \centering
            \begin{minipage}[t]{\appfigsevencolwidth}
                \centering
                \includegraphics[width=\linewidth]{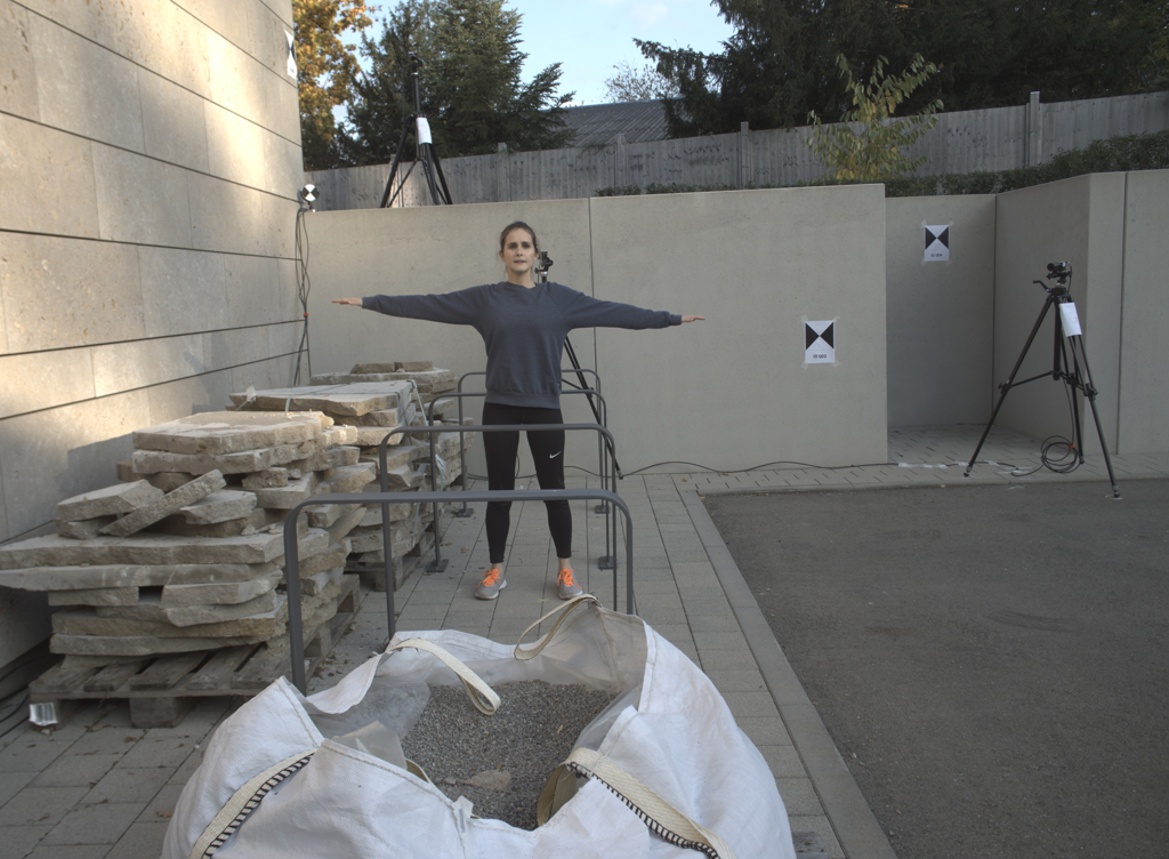}\\[-1pt]
            \end{minipage}\hspace{\appfigmargin}
            \begin{minipage}[t]{\appfigsevencolwidth}
                \centering
                \includegraphics[width=\linewidth]{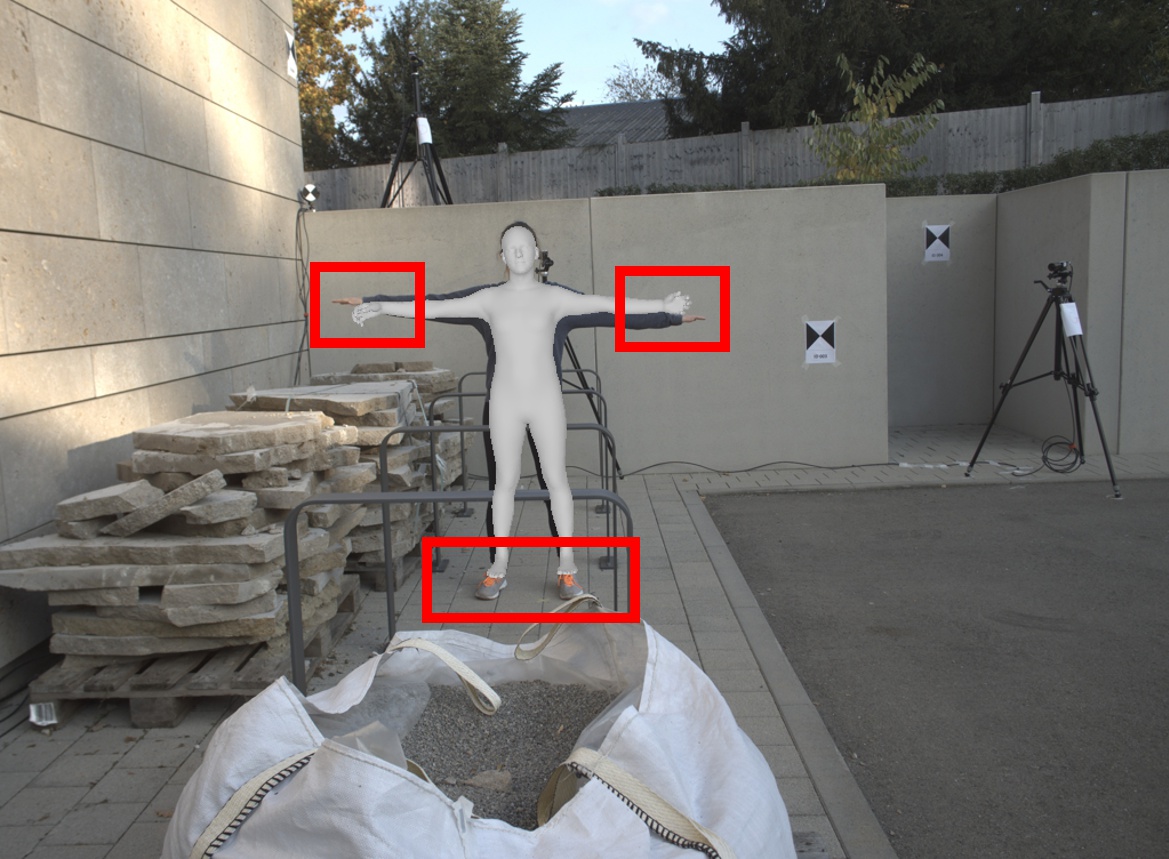}\\[-1pt]
            \end{minipage}\hspace{\appfigmargin}
            \begin{minipage}[t]{\appfigsevencolwidth}
                \centering
                \includegraphics[width=\linewidth]{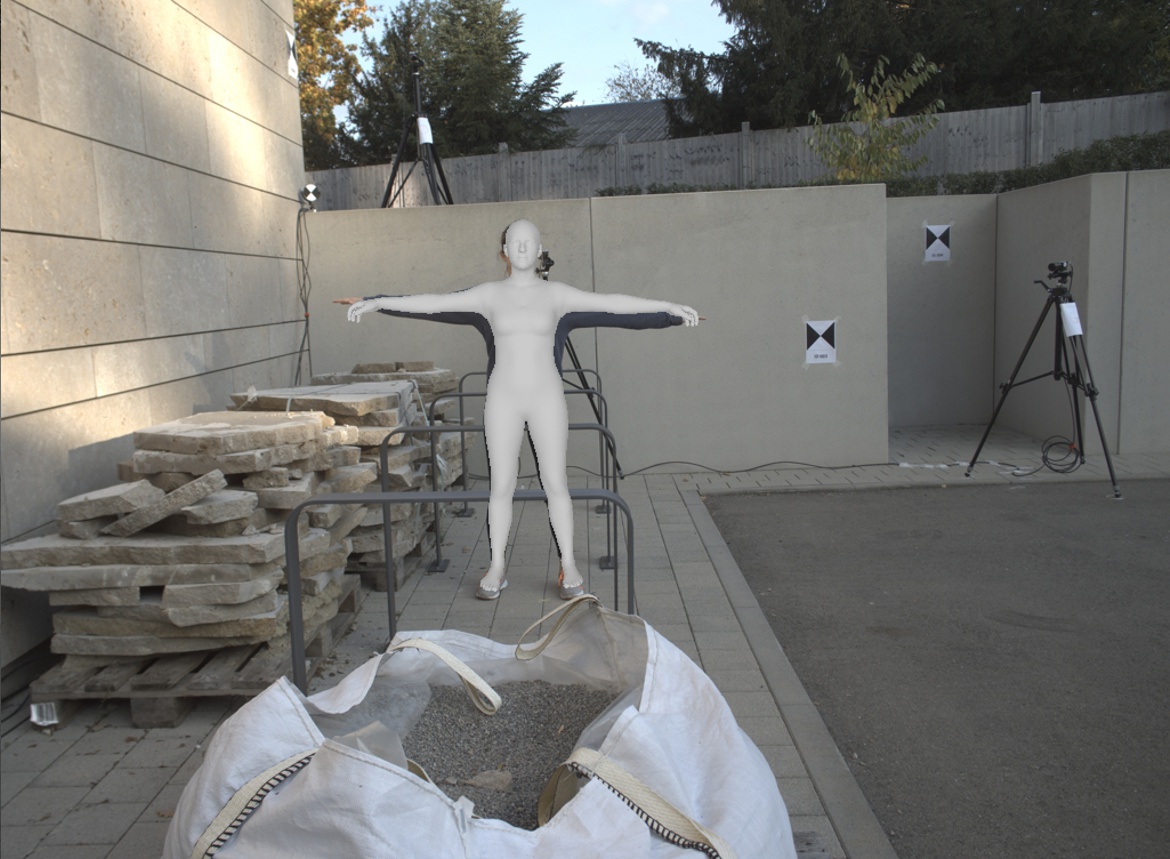}\\[-1pt]
            \end{minipage}
        \end{minipage}%
    }
    
    \caption{Additional body-pose recovery comparisons between our method and GENMO \cite{li2025genmo}.}
    \label{fig:appendix_qualitative_single_row_3}
\end{figure}
\clearpage

\end{document}